\pdfoutput=1

\documentclass[11pt]{article}

\usepackage[]{acl}

\usepackage{times}
\usepackage{latexsym}

\usepackage{bbm}
\usepackage{subcaption}
\usepackage{graphicx}
\usepackage{booktabs}
\usepackage{color}
\usepackage{amsmath}

\usepackage[T1]{fontenc}

\usepackage[utf8]{inputenc}

\usepackage{microtype}

\newcommand{\smallmodel}{\textsc{125m}}
\newcommand{\medmodel}{\textsc{1.3b}}
\newcommand{\largemodel}{\textsc{6.7b}}
\newcommand{\xlmodel}{\textsc{13b}}
\newcommand{\xxlmodel}{\textsc{30b}}
\newcommand{\xxxlmodel}{\textsc{175b}}
\newcommand{\ppl}{\textsc{ppl}}

\definecolor{gg}{HTML}{0F9D58}
\definecolor{rr}{HTML}{DB4437}
\definecolor{bb}{HTML}{4285F4}

\newcommand{\bigbench}{BIG-Bench}
\newcommand{\mauve}{\textsc{Mauve}}

\title{Training Trajectories of Language Models Across Scales}

\newcommand{\affilsup}[1]{\rlap{\textsuperscript{\normalfont#1}}}
\author{
    Mengzhou Xia\affilsup{1}~,~Mikel Artetxe\affilsup{2}~,~
    Chunting Zhou\affilsup{2}~,~
    Xi Victoria Lin\affilsup{2}~,~ \\
    \textbf{Ramakanth Pasunuru}\affilsup{2}~\textbf{,}~
    \textbf{Danqi Chen}\affilsup{1}~\textbf{,}~
    \textbf{Luke Zettlemoyer}\affilsup{2}~\textbf{,}~
    \textbf{Ves Stoyanov}\affilsup{2} \\
    $^1$Princeton University \quad $^2$Meta AI\\
    \texttt{mengzhou@princeton.edu} \\
}

\begin{document}
\maketitle

\begin{abstract}
 Scaling up language models has led to unprecedented performance gains, but little is understood about how the training dynamics change as models get larger. How do language models of different sizes learn during pre-training? Why do larger language models demonstrate more desirable behaviors? In this paper, we analyze the intermediate training checkpoints of differently sized OPT models~\cite{zhang2022opt}---from \smallmodel{} to \xxxlmodel{} parameters---on next-token prediction, sequence-level generation and downstream tasks. We find that 1) at a given perplexity and independent of model sizes, a similar subset of training tokens see the most significant reduction in loss, with the rest stagnating or showing double-descent behavior~\cite{Nakkiran2020Deep}; 2) early in training, all models learn to reduce the perplexity of grammatical sequences that contain hallucinations, with small models halting at this suboptimal distribution and larger ones eventually learning to assign these sequences lower probabilities; and 3) perplexity is a strong predictor of in-context learning performance on $74$ multiple-choice tasks from \bigbench{}, and this holds independently of the model size. Together, these results show that perplexity is more predictive of model behaviors than model size or training computation.\footnote{Code is publicly available at \url{https://github.com/xiamengzhou/training\_trajectory\_analysis}.} 

 \end{abstract}
  
 \section{Introduction}
 Scaling up language models has been shown to improve language modeling perplexity~\cite{kaplan2020scaling, hernandez2022scaling} as well as zero- or few-shot end task accuracies~\cite{brown2020language,rae2021scaling, chowdhery2022palm, zhang2022opt}. 
However, relatively little is understood about why or how this happens. How do the training dynamics differ as models get larger? What do language models of different sizes learn during pre-training in terms of both generating texts and solving end tasks?

We attempt to make progress to answer these questions by studying the training trajectories of differently-sized OPT models~\cite{zhang2022opt} through analyzing their intermediate checkpoints. In contrast to prior work, which studies the trajectories of small models with up to \textsc{300M} parameters \citep{liu2021probing,choshen2022grammar,blevins2022analyzing} or focuses on the language modeling objective alone~\cite{kaplan2020scaling,hernandez2021scaling, hernandez2022scaling}, we are the first to comprehensively study the training trajectories of large-scale autoregressive language models with up to {\xxxlmodel} parameters across a wide range of settings.

Repeatedly across training and different model scales, we analyze three aspects of model performance: (i) next-token prediction on subsets of tokens (ii) sequence-level generation and (iii) downstream task performance. We use perplexity,  which is closely tied to language model evaluation, as the major metric throughout the study.   


For \textbf{next-token prediction} (\S\ref{sec:next_token_prediction}), we study the trajectory by categorizing each token's prediction as \textit{stagnated}, \textit{upward} or \textit{downward} according to its perplexity trend as training progresses. We find each category comprising a significant number of tokens: while a significant number of tokens' perplexity stagnate, a subset of tokens with an increasing perplexity in smaller models exhibit a double-descent trend~\cite{Nakkiran2020Deep} where perplexity increases and then decreases in larger models. These behaviors primarily emerge at a similar validation perplexity across model scales.


For \textbf{sequence-level generation} (\S\ref{sec:sequence-level-generation}), we study the distribution shift at a document level (50-500 tokens) by decoding sequences that small/large models favor more than the other. Human texts present expected scaling patterns in that they are best modeled by larger (or longer trained) models. However, to our surprise, large models are better at modeling less human-like texts which contain synthetic noise and factually incorrect prompts. We propose an approach to decoding texts that small models favor more than large models from an interpolated distribution induced by combining signals from both models and find them grammatical but hallucinating.\footnote{Concurrent to our work, \citet{li2022contrastive} propose a similar contrastive decoding approach for a different application. Refer to Appendix \ref{app:contrastive_decoding} for more details.} All models go through a stage during training where the perplexity for such texts decreases; small models halt at this suboptimal distribution, while larger models escape it by eventually increasing the perplexity of these unnatural texts. 

We further connect language modeling perplexity to \textbf{downstream tasks} (\S\ref{sec:downstream}). By evaluating more than 70 multiple-choice tasks in \bigbench{}~\cite{srivastava2022beyond}, we find that language modeling perplexity correlates well with few-shot in-context learning performance along the trajectory, regardless of model sizes. The gradual divergence of likelihood between correct and incorrect options leads to improvements in in-context learning. 

Our work presents a comprehensive study of training trajectories of language models trained with similar procedures, e.g., OPT. We conclude that language models learn the same phenomena in the same order across different model sizes. The overall model perplexity is a composite measure of which language phenomena have been learned.

\begin{figure}
  \centering
  \includegraphics[width=\linewidth]{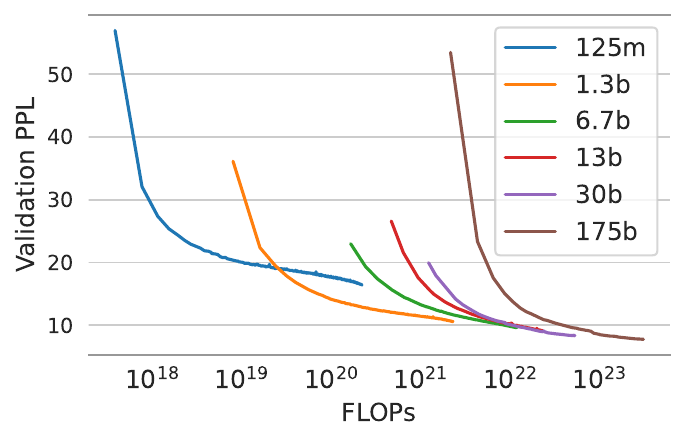}
  \caption{Validation perplexity (PPL) of OPT models against training FLOPs. Our work suggests that models with comparable perplexity levels during training exhibit similar predictions, regardless of their scales.}
  \label{fig:val_perp}
\end{figure}

  \begin{figure*}[t]
    \centering
    \includegraphics[width=0.32\textwidth]{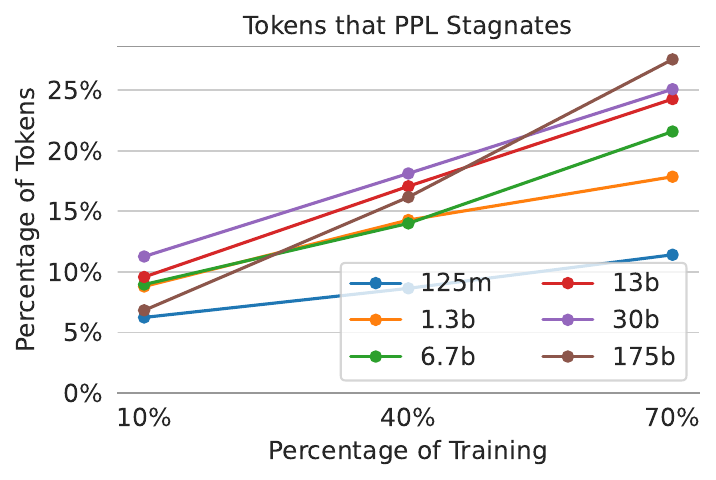}
    \includegraphics[width=0.32\textwidth]{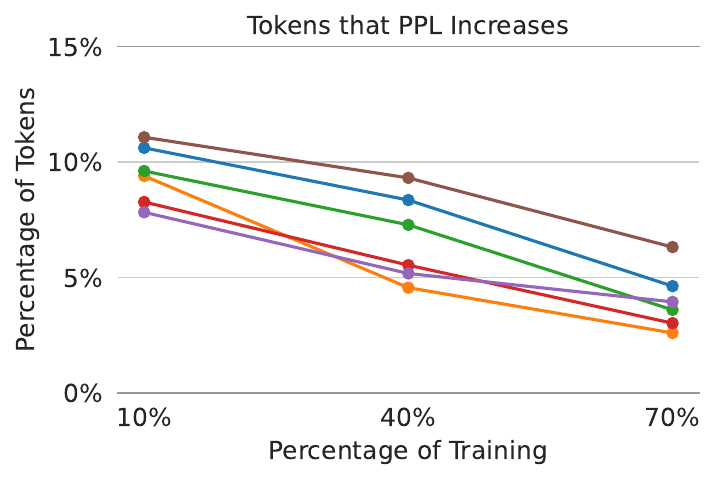}
    \includegraphics[width=0.32\textwidth]{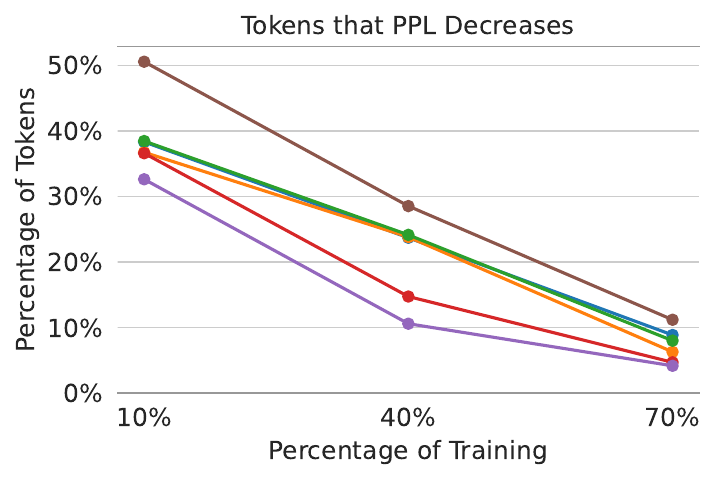}
    \caption{Percentage of predictions where perplexity stagnates (left), follows an upward trend (middle) and an downward trend (right). X-axis denotes that the trend is estimated after $p$\% percentage of training.}
    \label{fig:percentage_of_predictions}
  \end{figure*}

  \section{Experimental Settings}
\label{sec:models_and_setup}

\paragraph{Models.} Unless otherwise indicated, all of our experiments use OPT~\cite{zhang2022opt}, a collection of open-source autoregressive language models. OPT models serve as a good fit for this study due to their controlled pre-training procedures across all model sizes. In particular, all the models share the same tokenization and are trained on the same training data, covering a total of 300B tokens (180B unique). Note that different-sized models differ in batch sizes and total number of steps.\footnote{See Appendix \ref{app:checkpoint_info} for more details of model checkpoints.} We collect intermediate checkpoints from the authors and perform evaluations of these checkpoints across six different sizes: \smallmodel, \medmodel, \largemodel, \xlmodel, \xxlmodel, and \xxxlmodel. 

\paragraph{Validation perplexity.} Throughout this paper, we use \textit{Validation Perplexity (Valid PPL)} to refer to the autoregressive language modeling perplexity measured on the entire validation set. We use the original OPT validation set, a held-out subset of the training corpus that covers a wide range of domains, such as books, news, and subtitles. We plot the trajectory of validation perplexity in \autoref{fig:val_perp}, which follows a similar power-law pattern observed in previous scaling work \citep{kaplan2020scaling,hoffmann2022training}. %

\paragraph{Methodology.} We aim to understand how models of different sizes behave throughout training as a function of computing (FLOPs)\footnote{We estimate the number of FLOPs of language models following \citet{chowdhery2022palm}.} and validation perplexity. Throughout the paper, we use different measurements to characterize model behavior and plot them against these two metrics.
  \section{Next-Token Prediction}

\begin{figure*}[t]
  \centering
  \includegraphics[width=0.5\textwidth]{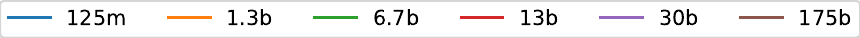} \\
  \includegraphics[width=0.32\textwidth]{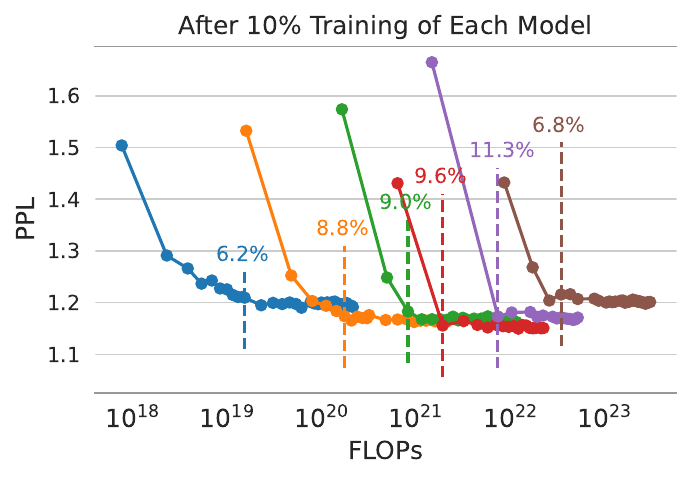}
  \includegraphics[width=0.32\textwidth]{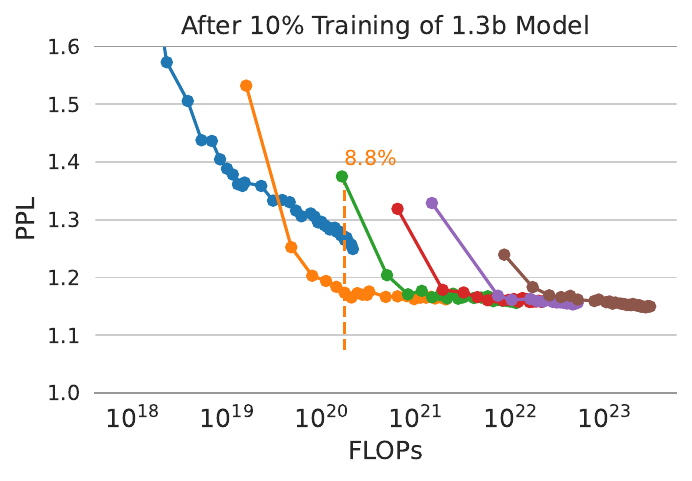}
  \includegraphics[width=0.32\textwidth]{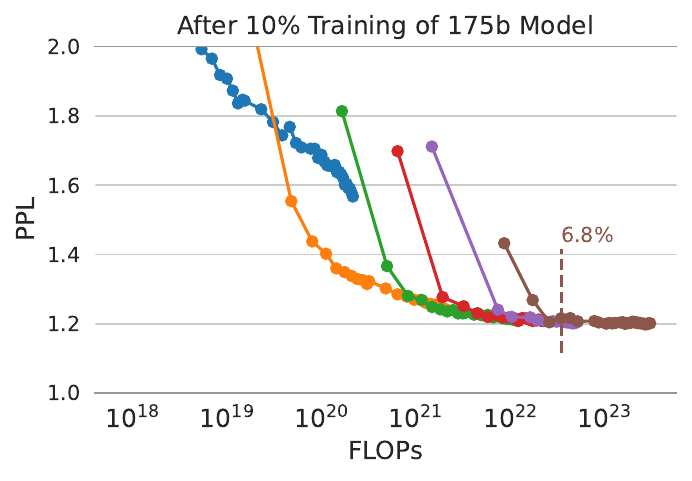}
  \caption{Perplexity of stagnated tokens. Left: different models are evaluated on different subsets of tokens selected after 10\% of training of individual models (all showing a stagnated trend after 10\%). Middle/right: all models are evaluated on the \textit{same set of tokens}, selected after 10\% of training of the \medmodel{} model and the \xxxlmodel{} model respectively. The number next to the dashed line denotes the percentage of the selected tokens out of all tokens. Stagnated tokens selected by a smaller model (\medmodel) are stagnated in larger models. Stagnated tokens selected by a larger model (\xxxlmodel) present a downward trend in perplexity in smaller models.}
  \label{fig:stagnated_tokens}
\end{figure*}

\begin{figure*}[t]
  \centering
  \includegraphics[width=0.5\textwidth]{images/legend-allmodels.pdf} \\
  \includegraphics[width=0.32\textwidth]{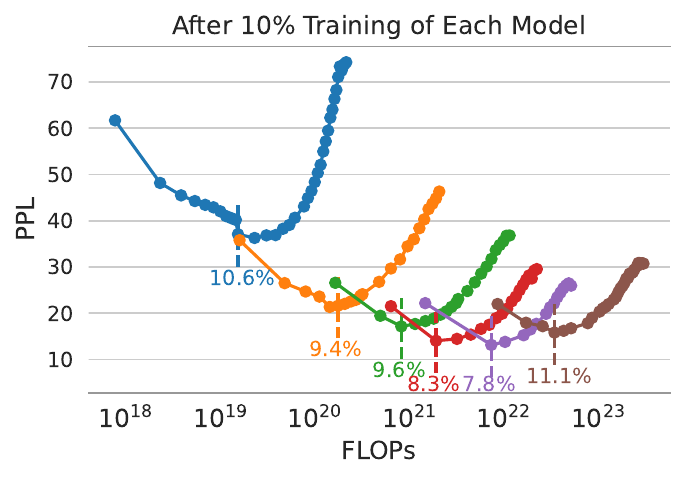}
  \includegraphics[width=0.32\textwidth]{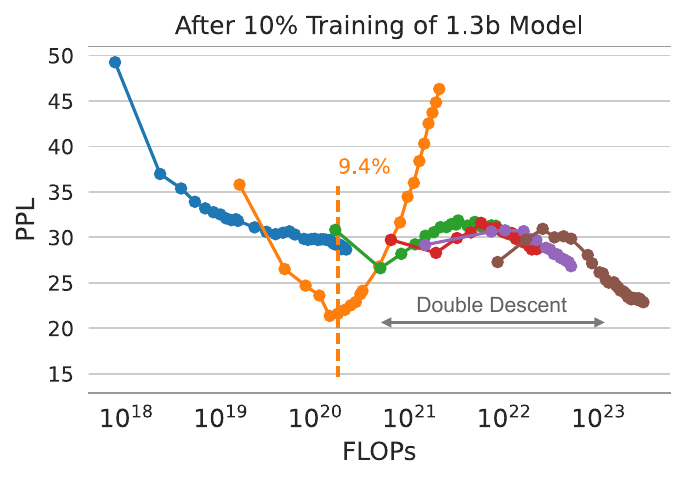}
  \includegraphics[width=0.32\textwidth]{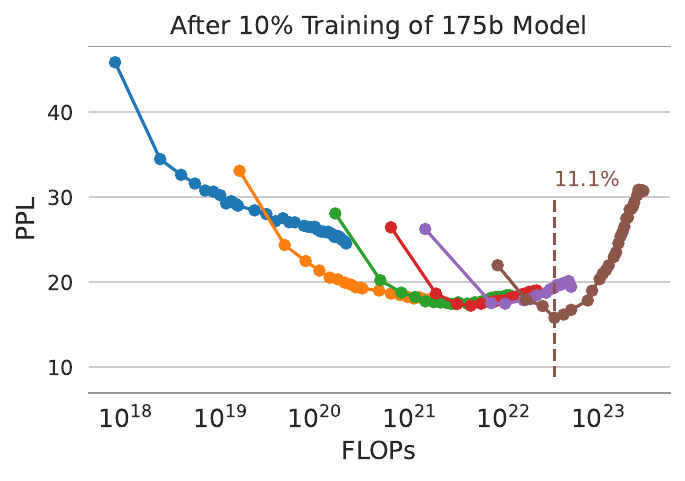}
  \caption{Perplexity of upward-trend tokens. Left: different models are evaluated on different subsets of tokens selected after 10\% of training of individual models (all showing a downward-then-upward trend). Middle/right: all models are evaluated on the \textit{same set of tokens}, selected after 10\% of training of the \medmodel{} model and the \xxxlmodel{} model respectively. The number next to the dashed line denotes the percentage of the selected tokens out of all tokens.  Tokens selected by a smaller model (\medmodel) present a double descent-like trend in larger models. Tokens selected by a larger model (\xxxlmodel) present a downward trend in the smaller models. }
  \label{fig:upward_trend_tokens}
\end{figure*}

\label{sec:next_token_prediction}
Autoregressive language models are trained to predict the next token given a context. \autoref{fig:val_perp} shows that validation perplexity, aggregated over all positions, gradually declines as training progresses. However, it is not clear if all token instances evolve similarly to the aggregated measurement. In this section, we study the trajectory of next-token predictions, dividing them into three categories---stagnated, upward trend, or downward trend---to understand how language models gradually learn new language phenomena.

\subsection{Methodology} 
We evaluate intermediate checkpoints on a subset of validation data.\footnote{More dataset details are in Appendix \ref{app:trend_data}.} For each context-token pair $(c,t)$, we obtain a series of perplexities $\ppl_{m_1}(t \mid c), \ppl_{m_2}(t \mid c), \ldots, \ppl_{m_n}(t \mid c)$ for checkpoints $m_1, m_2, \ldots, m_n$. We use linear regression to estimate the slope of a normalized series to roughly capture its trend. Starting from any intermediate checkpoint after $p\%$ of training (assuming that it is the $j$-th checkpoint) to the end checkpoint $m_n$, $\forall i \in [j, n]$, we fit the following function to learn the parameters $\alpha$ and $\beta$ for each series: 

\begin{equation}
  \frac{\ppl_{m_i}(t \mid c)}{\ppl_{m_j}(t \mid c)}  = \alpha + \beta \cdot (i-j) .
\end{equation}
Note that different starting points might result in different trend estimations. We categorize the trends as follows based on $\beta$ and its significance:

\paragraph{Upward trend.} If $\beta>0$ and its $p$-value is $<0.05$, we consider that the series follows an upward trend (\textit{forgetting}).

\paragraph{Downward trend.} If $\beta<-0$ and its $p$-value is $<0.05$, we consider that the series follows a downward trend (\textit{still learning}).

\paragraph{Stagnated trend.}  If a series does not follow an upward or downward trend, and the start and end values fall in a restricted interval, that is, $0.95 \le \mathrm{\textsc{PPL}}_{m_j} /{\mathrm{\textsc{PPL}}_{\mathrm{\textsc{avg}}}} \le 1.05$ and $0.95 \le \mathrm{\textsc{PPL}}_{m_n} /{\mathrm{\textsc{PPL}}_{\mathrm{avg}}}\le 1.05$, where $\mathrm{\textsc{PPL}}_{\mathrm{avg}} =  \exp(\frac{1}{n-j+1}\sum_i \log\textsc{PPL}_{m_i})$, we consider the series to be stagnated (\textit{already learned}).

\vspace{1em}
We design the criteria to roughly capture the trend of the perplexity series of each next-token prediction. Under these criteria, a stagnated series from an earlier checkpoint would continue to stagnate, and a series that follows an upward or downward trend earlier might turn stagnated afterwards. The criteria do not necessarily cover all the series---wavy series with a large variance do not fall within any category and are eliminated. For the rest of the section, for simplicity, we use \textit{tokens} to refer to context-token pairs.

\subsection{Analysis}
\paragraph{Percentage of tokens.} We show the percentage of tokens that follow each trend in \autoref{fig:percentage_of_predictions}. Overall, the percentage of stagnated tokens increases and the percentage of the other two types of tokens decreases, indicating that more tokens get to be \textit{learned} and fewer tokens are still learning or, more surprisingly, forgetting as training progresses.~\footnote{Only around $60\%$ tokens are captured by our criteria and please find more details on other tokens in Appendix \ref{app:other_tokens}.}

\paragraph{Stagnated tokens.} We select stagnated tokens starting from 10\%  of training for a particular model and analyze the trajectory of these same tokens in other models. As shown in \autoref{fig:stagnated_tokens} (middle), we observe that stagnated tokens after $10\%$ of training in a small model (\medmodel{}) also stagnate in larger models. However, the stagnated tokens selected by a large model (\xxxlmodel{}) still show a downward trend in smaller models. This suggests that larger models' stagnated tokens are roughly a superset of smaller models. On manual inspection, stagnated tokens are primarily non-content words such as prepositions, determiners, and punctuations.

\paragraph{Upward trend tokens.} Similarly, we present the perplexity of upward trend tokens in \autoref{fig:upward_trend_tokens}. The leftmost figure shows that such a phenomemon exists for all the models. For tokens that present an upward trend after $10\%$ training of a small model (\medmodel), we observe a stepwise double descent~\cite{Nakkiran2020Deep} trend in larger models' trajectories, where the perplexity first increases and then decreases. We are the first to observe this phenomenon during language model training, and it suggests that larger models, with more computation and a larger capacity, first overfit to this subset of tokens and further generalize better for them. For the tokens identified after $20\%$ training of the largest model (\xxxlmodel), the upward trend appears only at the end of training for the \xlmodel{} and \xxlmodel{} models. We find it hard to characterize these tokens considering their contexts,\footnote{More details are in Appendix \ref{app:token_property}.} but the synergy across model sizes strongly suggests that consistent types of learning are triggered at particular computation levels for models across scales.~\footnote{We explore the upward trends with different starting points and model scales in Appendix \ref{app:double_descent}.} 

\paragraph{Summary.} In conclusion, large models first replicate small models' behavior on the same subset of tokens, and further unlock exclusive phenomena when fueled with more computation. In Appendix \ref{app:ppl_trend}, we find that trajectories of differently-sized models largely overlap when plotting against validation perplexity, indicating that they make similar predictions at a similar perplexity.\footnote{Please find more discussions in Appendix \ref{app:ppl_trend}.}

  \section{Sequence-Level Generation}
  \label{sec:sequence-level-generation}
In this section, we extend the analysis from token-level predictions to entire sequences, up to 50-500 tokens. Larger language models consistently obtain a better perplexity in modeling human texts such as Wikipedia, with the perplexity decreasing as the model size and training computation increases (\autoref{fig:val_perp}). Autoregressive language models are probabilistic models of sequences that can generate strings of text. If larger models assign a higher probability to virtually all human-authored texts, what sequences do smaller models favor? We aim to first characterize these sequences and further analyze learning behavior on them to understand how models of different sizes evolve into their final distributions. In what follows, we first show that it is difficult to manually design such sequences, as large models can also favor corrupted or factually incorrect texts (\S\ref{subsec:manual-design}). We then devise a decoding algorithm to automatically generate sequences favored by smaller models (\S\ref{subsec:methodology}), and conclude with an analysis of such sequences (\S\ref{subsec:analysis}).

\subsection{Manual Design} \label{subsec:manual-design}

\begin{figure}
\includegraphics[width=0.99\linewidth]{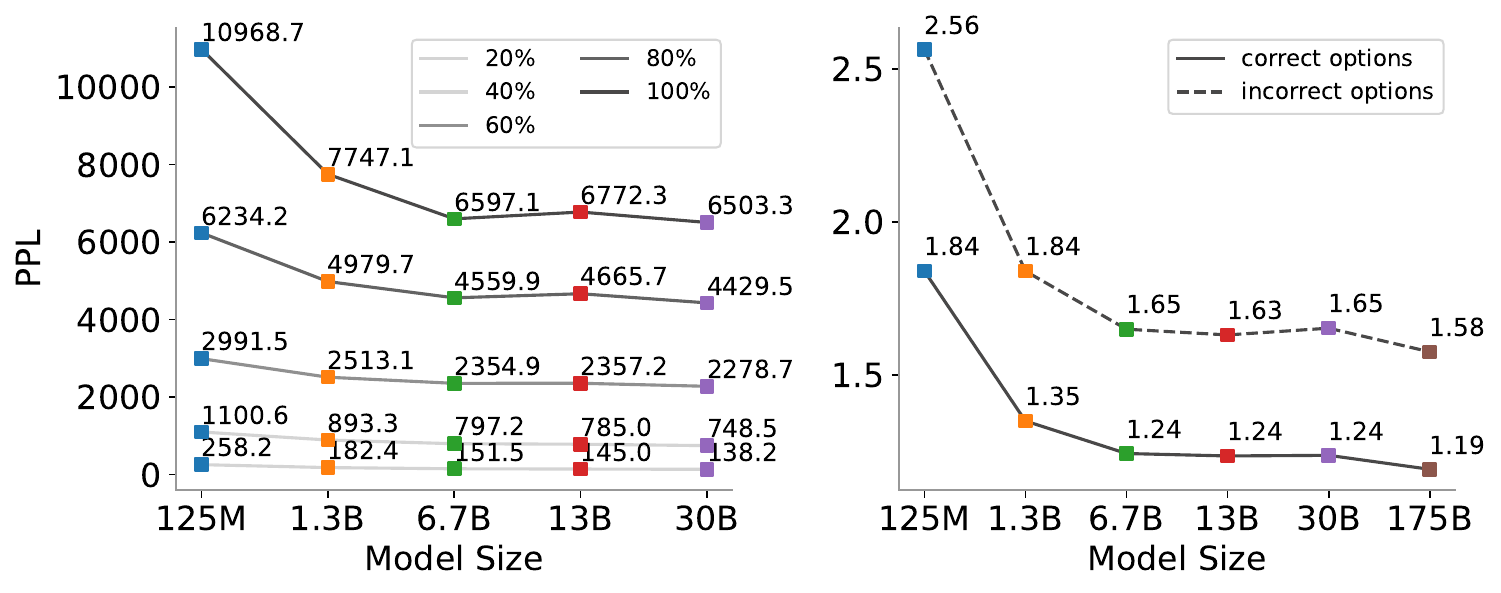}
\caption{Scaling trends for corrupted datasets (p\% random tokens) and options in multiple choice tasks. The perplexity on corrupted texts and incorrect options decrease as model size increases, even for sequences consisting of completely random tokens ($p=100$). 
}
\label{fig:noise}
\end{figure}

\paragraph{Corrupted datasets.} We hypothesize that injecting noise into human texts might reverse the scaling trend (i.e., perplexity on corrupted texts might increase as model size increases). To test this hypothesis, we replace $20\%$, $40\%$, $60\%$, $80\%$, and $100\%$ of the subwords in each sequence with random subwords. We evaluate corrupted datasets on the \textit{final} model checkpoints and report the perplexity in \autoref{fig:noise} (left). Contrary to our hypothesis, downward trends largely retain across all noise levels, even when the entire sequence consists of random tokens ($100\%$). This can be explained by the copy-and-complete interpretation for in-context learning described in~\citet{olsson2022context}: larger models fare better at making predictions to follow the context distribution than smaller models, even when the context is pure noise.\footnote{Please find details on corrupted datasets in Appendix \ref{app:corrupted_dataset}.}

\paragraph{Incorrect options of multiple-choice tasks.} We next hypothesize that the perplexity of incorrect options for multiple-choice tasks might present an inverse scaling trend, as they are generally factually wrong. We present the perplexity of correct and incorrect options of 74 multiple-choice tasks from the \bigbench{} dataset in \autoref{fig:noise}.\footnote{Details on task selection are in Appendix \ref{app:task_selection}.} However, we find that the perplexity of correct and incorrect options decreases as the size of the model increases.\footnote{To clarify, we are not discussing task accuracy here, but the scaling trend of correct and incorrect options. Find examples of correct and incorrect prompts in \autoref{tab:bigbench_examples}.}

In summary, our initial attempt failed---we are not able to manually construct texts that are more probable in smaller models than larger models.

\subsection{Methodology} \label{subsec:methodology}
To continue our search for such texts, we next devise a decoding approach that combines signals from two models and generates texts based on the interpolation of their distributions:
\begin{align}
& p'_i = \lambda_1 \cdot p_{s}(x_i | x_{<i}) + \lambda_2  \cdot p_{l}(x_i | x_{<i});
\end{align}

\noindent where $p_s$ and $p_l$ are the next-token distributions from the small and large models, respectively, and $\lambda_1, \lambda_2 \in [-1, 1]$. A set of $\lambda_1$ and $\lambda_2$ denotes a specific configuration. When $\lambda_1 = 0, \lambda_2 = 1$, it is simply decoding with the large model; when $\lambda_1 = 1, \lambda_2 = -1$, the decoding process favors the small model's prediction and suppresses the large model's prediction. This is the configuration that decodes sequences that small models have a lower perplexity on than large models. 

We further remove tokens that have a negative score, and renormalize the distribution $p'_i$ to ensure that the sum of the probabilities of all tokens is 1:
\begin{align}
& p(x_i | x_{<i}) = \frac{\mathbbm{1}(p'_i > 0) \cdot p'_i }{\sum \mathbbm{1}(p'_i > 0)\cdot p'_i}.
\end{align}

\paragraph{Generation process.} We decode sequences with two models, \smallmodel{} and \xxlmodel{}, using different configurations of $\lambda_1$ and $\lambda_2$. We take the first 5 tokens of a subset of validation documents as prompts and generate 50 tokens conditioned on them.\footnote{We also generate longer sequences up to 100 and 500 words and the conclusions hold similarly. More discussions can be found in Appendix \ref{app:longer_sequences}.} We try greedy search and nucleus sampling~\cite{holtzman2019curious} for decoding and evaluate the texts decoded from each configuration as follows: 1) we measure the text perplexity at final checkpoints of different-sized models to understand its scaling trend; 2) we measure the text perplexity at all intermediate checkpoints to understand how the perplexity evolves as training progresses. 

\subsection{Analysis} \label{subsec:analysis}

\begin{figure*}
    \centering
\includegraphics[width=\linewidth]{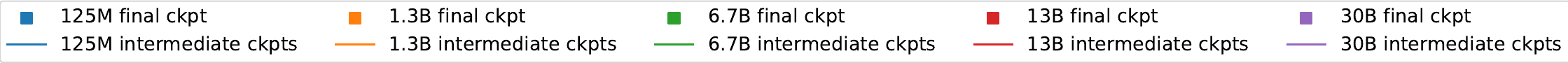}

\includegraphics[width=0.49\linewidth]{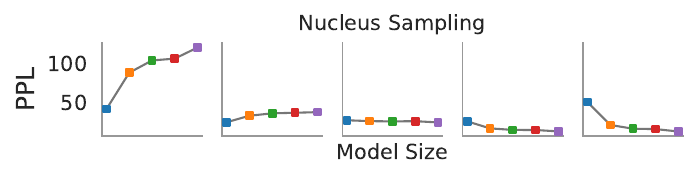} 
\includegraphics[width=0.49\linewidth]{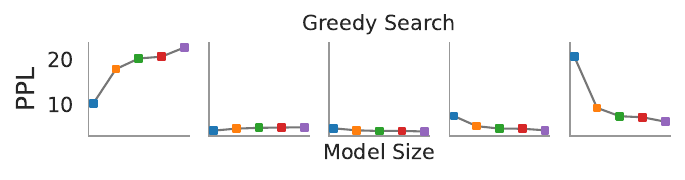} 
\includegraphics[width=0.49\linewidth]{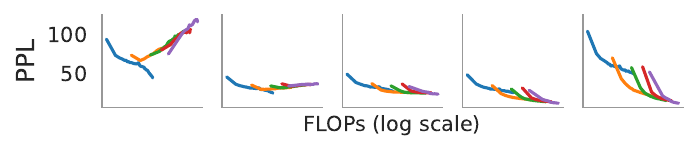}
\includegraphics[width=0.49\linewidth]{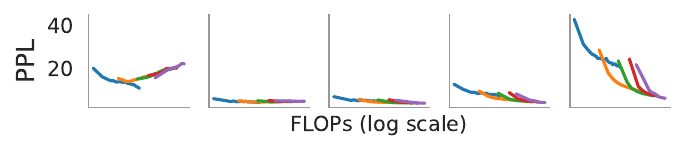} \\
\vspace{-0.3cm}
\includegraphics[width=0.49\linewidth]{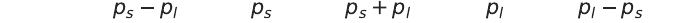}
\includegraphics[width=0.49\linewidth]{images/gen_xlabel.pdf}
\vspace{-0.3cm}
\caption{
Perplexity of texts (generated with $\lambda_1 p_s + \lambda_2 p_l$) evaluated with differently-sized final model checkpoints (first row) and perplexity trajectory evaluated over intermediate checkpoints against FLOPs (second row). Each column denotes one configuration with different $\lambda_1$ and $\lambda_2$. Note that all the texts are generated by combining signals only from \smallmodel{} and \xxlmodel{} models, but are evaluated over all the model scales.} 
\label{fig:gen}
\end{figure*}

\begin{figure}
    \centering
    \includegraphics[width=\linewidth]{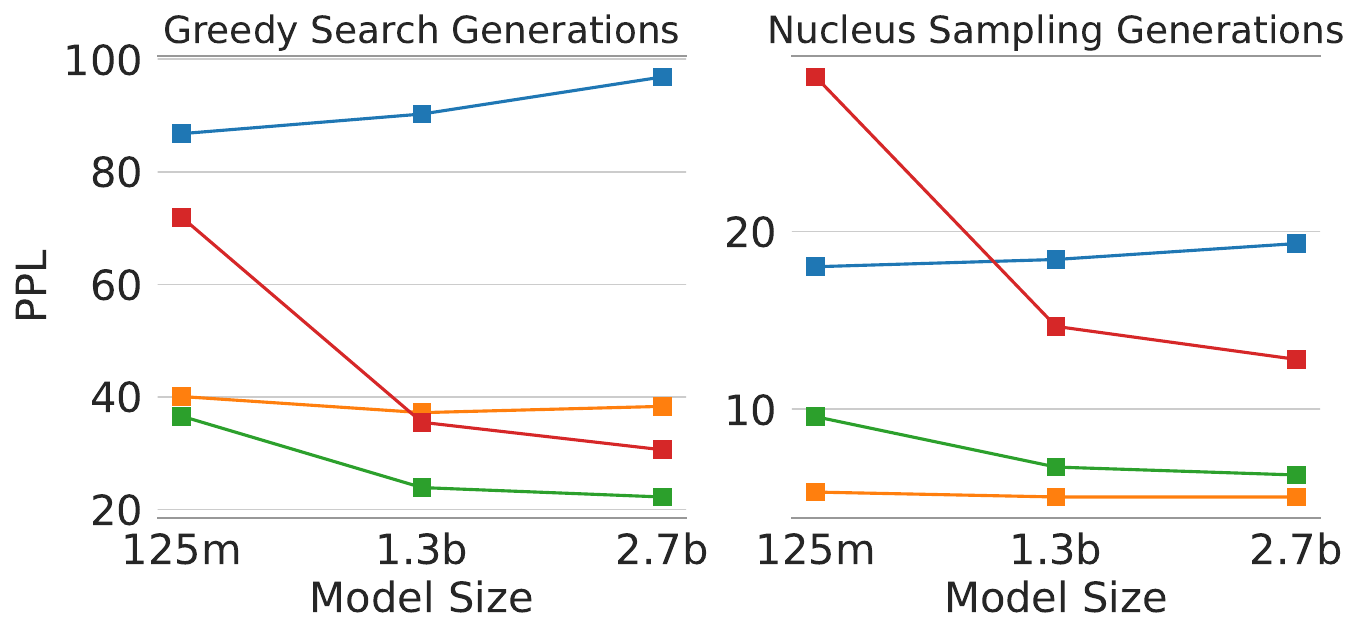}
    \includegraphics[width=.75\linewidth]{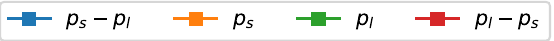}
    \caption{Evaluations using GPT Neo models on texts generated with OPT \smallmodel{} and OPT \xxlmodel{} models. The perplexity follows a similar trend as OPT, suggesting a systematic distribution shift between model sizes.}
    \label{fig:eval-gpt-neo}
\end{figure}

\paragraph{Inverse scaling.} As shown in \autoref{fig:gen} (row 1),  we confirm that the perplexity of texts generated with the $p_s - p_t$ configuration presents an inverse scaling trend---perplexity increases as model size increases (column 1, 5). Other configurations either only show a modest upward trend ($p_s$), or a normal downward trend ($p_l$ and $p_l - p_s$). Even though models of intermediate sizes (\medmodel, \largemodel, \xlmodel) are not involved in decoding, the scaling trend holds systematically across all model sizes.  To further verify the universality of the phenomenon in other families of language models, we evaluate the generated texts with final GPT Neo checkpoints~\cite{gpt-neo}, which were trained on the Pile dataset~\cite{gao2020pile}. As shown in \autoref{fig:eval-gpt-neo}, the perplexity trend aligns with OPT models. This confirms that the texts generated with our approach are not a result of model or data artifacts, but embody universal properties exhibiting a similar scaling trend in other model families.

\paragraph{Perplexity trajectory of generated sequences.} In the second row of \autoref{fig:gen}, we present the perplexity trajectory of texts generated with different configurations. We observe that texts generated based on $p_s - p_l$ and, to a less extent, $p_s$, largely differ from the other configurations: \smallmodel{} checkpoints present a downward trend, while other checkpoints present an upward trend. This might suggest that differently-sized models optimize in different directions for phenomena specific to these texts. However, taking a closer look, we observe that the \medmodel{} model also shows a downward trend at the beginning, which turns upward afterwards. This indicates that all models improve the perplexity of these texts at first but, with more training FLOPs, larger models shift away from this specific distribution where the \smallmodel{} model stalls. In Appendix \ref{app:genppl_ppl}, we further show that perplexity of the sequences decoded by contrasting the two models ($p_s - p_l$ and $p_l - p_s$) are less aligned with validation perplexity as other configurations.

\begin{table*}[t]
    \centering
    \def\arraystretch{0.7}
    \begin{tabular}{@{}p{1.0cm}p{7.2cm}p{7.2cm}@{}} \toprule
        \small{\textbf{Dist.}} & \multicolumn{1}{c}{\textbf{\small{Greedy Search}}}  & \multicolumn{1}{c}{\textbf{\small{Nucleus Sampling}}}  \\ \midrule
    & \small{\textbf{\textit{Fortunately, the day wasn't all ...}}} & \small{\textbf{\textit{Fortunately, the day wasn't all ...}}} \\
    \cmidrule(lr){2-2} \cmidrule(lr){3-3}
    $p_s - p_l$  & \small{ that great. The sun was setting and the sun was falling. I went to bed and woke my husband, who was asleep in his bed, to find that I was still asleep in the middle of the night with him. He was still awake when we}  & \small{ that good when the computer said doom and gloom about me. Sure enough, because of our stubborn attempt at terrorizing him via cyberbackup (which relied heavily on computer traffic management (VCMD) to ensure my identity), I was able fix my old}  \\
    \cmidrule(lr){2-2} \cmidrule(lr){3-3}
    $p_s$         & \small{ that bad. I was in the middle of a long day of work and I was in the middle of a long day of work. I was in the middle of a long day of work. I was in the middle of a long day}      & \small{ that bad. Not because the weather wasn't bad, but because of how many people didn't move their car around. For those who did, I wanted to say thanks to everyone else who still had a tire change on.  That doesn't change}                                                                                              \\
    \cmidrule(lr){2-2} \cmidrule(lr){3-3}
    $p_s + p_l$ & \small{ bad.
    I was able to get a few things done, and I was able to get a few things done.
    I was able to get a few things done, and I was able to get a few things done.
    I was able to} & \small{ cold and we didn't have to set up a heated bed so we wouldn't freeze off in the middle of the night.
    It was a nice fall day and I had just finished wrapping up the color scheme on the wall.
    I still haven} \\
    \cmidrule(lr){2-2} \cmidrule(lr){3-3}
    $p_l$         & \small{ bad. I got to spend some time with my family, and I got to see my friends. I got to see my friends, and I got to see my family. I got to see my family, and I got to see my}      & \small{ gloom, glum, and doom. One nice thing was the gift of snow for a few minutes this afternoon. It was fun to watch it pile up on the porch, watch the kids watch it pile up, and then run out and scatter}                                                                                                  \\
    \cmidrule(lr){2-2} \cmidrule(lr){3-3}
    $p_l - p_s$  & \small{ bad news. The U.N.'s Intergovernmental Panel on Climate Change released a landmark study showing that we have 12 years to limit climate catastrophe. And a group of young activists filed a landmark climate lawsuit in federal district court, demanding that the government take} & \small{ bad for Iowa fans. Tight end C. J. Fiedorowicz decided, for what has to be the millionth time now, to use Twitter as his own personal slogan board, and this time he decided to riff off the famous Bugs Bunny} \\ \bottomrule
    \end{tabular}
   \caption{Examples generated with greedy decoding and nucleus sampling under different configurations. The prompt is \textit{Fortunately, the day wasn't all}.}
    \label{tab:generations}
\end{table*}

\paragraph{Generated examples.} \autoref{tab:generations} presents examples generated with different configurations. We find that the generations from $p_s - p_l$ are grammatically correct and carry actual meanings both for greedy search and nucleus sampling, but manifest other issues: 1) they entail highly-unlikely semantic usages such as \textit{Fortunately, it wasn't all that great}---an ending word with a negative sentiment should be more prevalent; 2) the nucleus sampling examples, despite being fluent and consistent, hardly ground to real world scenarios. This suggests that small models are highly capable linguistically, and learning at scale primarily focuses on acquiring other types of knowledge.\footnote{We present more generated examples and have a more detailed discussion on generation quality in Appendix \ref{app:generation_quality}.}

  \section{Downstream Tasks}
  \label{sec:downstream}

In this section, we examine the trajectory of downstream tasks, evaluated on few-shot in-context learning (ICL).

\subsection{Task Selection and Evaluation} \bigbench{}~\citep{srivastava2022beyond} is a large collection of tasks for evaluating language models. We evaluate intermediate checkpoints on its subset of 74 multiple-choice tasks.\footnote{Mode details on task selection are in Appendix \ref{app:task_selection}.} \bigbench{} comes with predefined templates with a unified QA format for in-context learning, which mitigates the extra complexity of prompt design.\footnote{Examples of prompts are in Appendix \ref{app:prompts}.}

We focus on the 2-shot setting. Following \citet{srivastava2022beyond}, we randomly select two in-context learning examples (excluding the evaluation example itself) for each test instance and pick the candidate for each evaluation example that has the highest probability normalized over its length. We use the average 2-shot accuracy of downstream tasks as a proxy for in-context learning capability. 

\subsection{Trajectory of ICL Performance}
\begin{figure*}[t]
    \centering
    \includegraphics[width=0.7\linewidth]{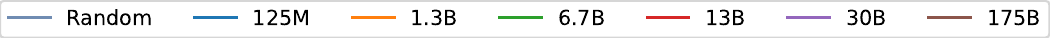}
    \includegraphics[width=\textwidth]{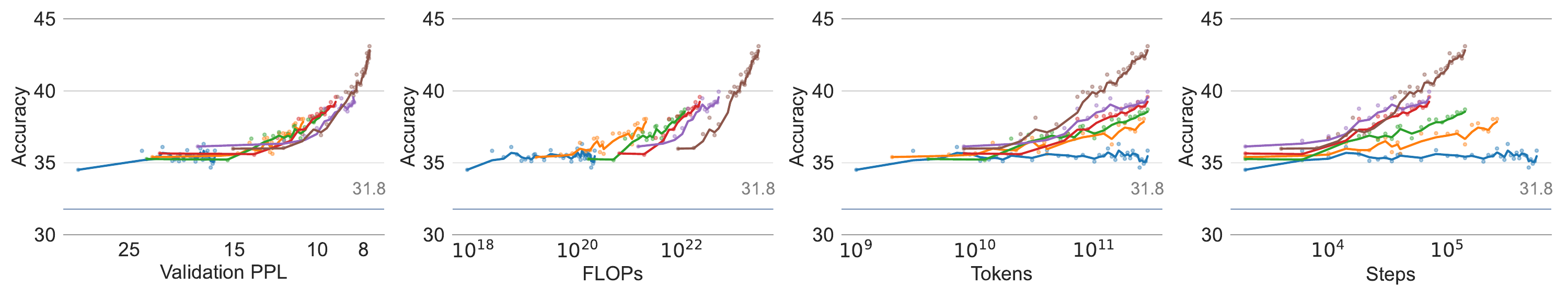}
    \caption{The 2-shot performance trajectory of 74 \bigbench{} tasks. The performance is measured by the average accuracy on the default set and plotted against validation perplexity, training FLOPs, training tokens and number of training steps. The task accuracy aligns with validation perplexity across different model sizes.}
    \label{fig:bigbench_fewshot} 
\end{figure*}

\begin{figure*}[t]
    \centering
    \includegraphics[width=0.6\linewidth]{images/legend-allmodels-dash.pdf} 
    \includegraphics[width=0.3\linewidth]{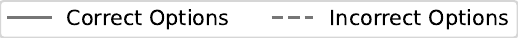} \\
    \includegraphics[width=0.24\linewidth]{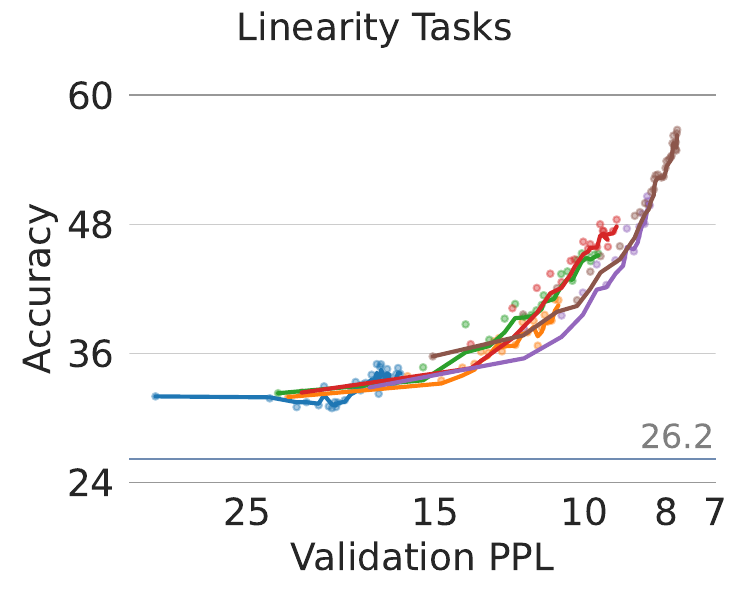}
    \includegraphics[width=0.24\linewidth]{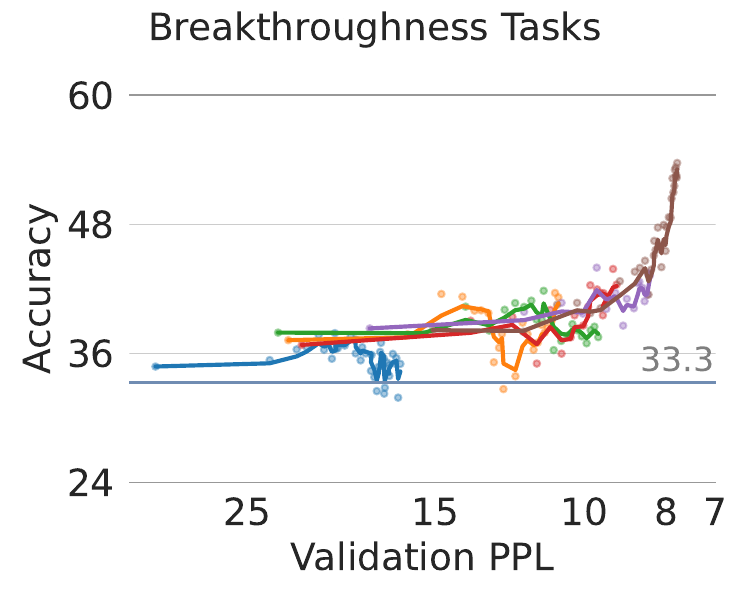}
    \includegraphics[width=0.24\linewidth]{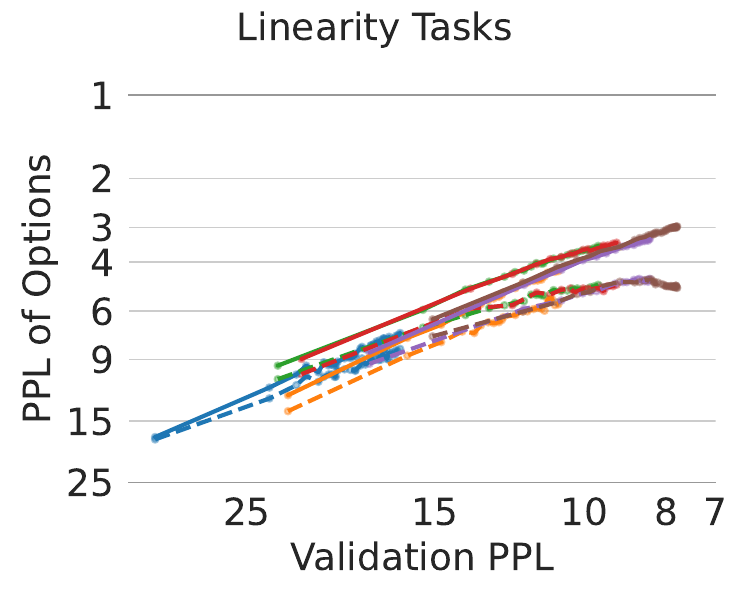}
    \includegraphics[width=0.24\linewidth]{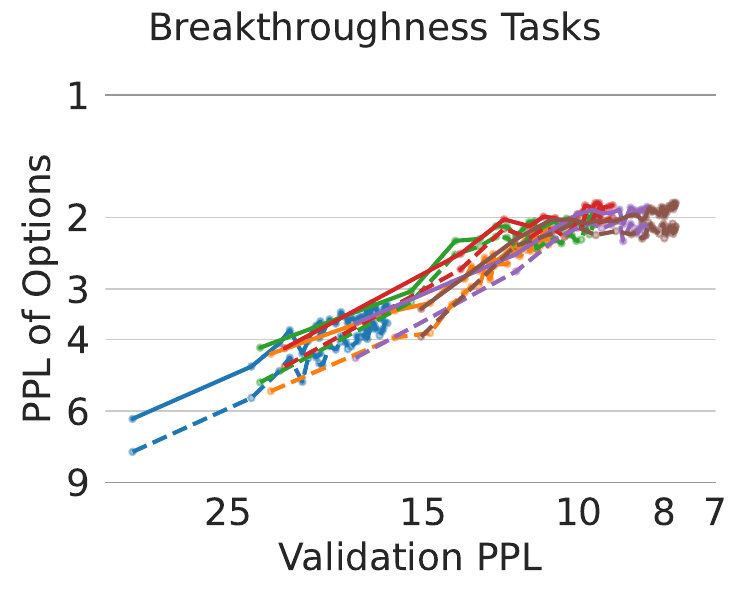}
    \caption{Trajectory of 2-shot in-context learning performance (left two) and option perplexity (right two) of 12 linearity and 6 breakthroughness tasks against validation perplexity. The perplexity divergence of correct and incorrect options drives the performance improvement. }
    \label{fig:bigbench_options}
\end{figure*}

\paragraph{ICL vs. valid PPL.} From \autoref{fig:bigbench_fewshot} (leftmost), it is evident that the downstream task performance strongly correlates with validation perplexity across all model sizes. The curves of different model sizes significantly overlap, indicating that when a small model and a large model are trained to the same perplexity level, they achieve comparable downstream task performance.

\paragraph{ICL vs. other metrics.} it is evident that plotting task accuracy against various metrics yields distinct patterns. Notably, when subjected to an equal amount of training FLOPs, the performance of smaller models consistently surpasses that of larger models, with the exception of the \smallmodel{} model. This observation implies that larger models possess untapped potential for improvement, especially when provided with more training FLOPs or data (Hoffmann et al., 2022; Touvron et al., 2023). Conversely, the remaining two plots indicate that larger models consistently outperform smaller ones when trained with the same number of training tokens and training steps.

\subsection{Linearity vs. Breakthroughness Tasks}
We select 12 tasks that present a linearity scaling pattern and 6 tasks that present a breakthroughness scaling pattern,\footnote{Breakthroughness here similar to the emergent dehavior defined in \citet{wei2022emergent}. Details on how we select linearity and breakthroughness tasks are in Appendix \ref{app:linear}.} and plot the perplexity of the correct and incorrect options for each group of tasks against validation perplexity in \autoref{fig:bigbench_options}.

The performance of breakthroughness tasks increases tremendously as the validation perplexity drops below 8. The perplexity gap between the correct and incorrect options also starts to expand at this point for the \xxlmodel{} and \xxxlmodel{} models. In contrast, the accuracy of linearity tasks gradually increases. The perplexity of correct and incorrect options first decrease as validation perplexity decreases, and it is only at the end of the curve that the perplexity of correct and incorrect options starts to diverge. This suggests that improvements in downstream accuracy are not generally driven by the model learning to assign a lower probability to incorrect candidates, but rather driven by the perplexity divergence of correct and incorrect options. 

\subsection{Breakthroughness Tasks Learn Smoothly on Trajectory}
In Appendix \ref{app:single_task}, we provide a detailed analysis of task accuracy in relation to perplexity and FLOPs for individual linearity and breakthroughness tasks. The corresponding plots can be found in Figure \ref{fig:single_linearity} and Figure \ref{fig:single_breakthrough}. As expected, these plots exhibit a significantly larger variance, showcasing substantial fluctuations in task performance during the training process. However, we still observe a notable alignment between task accuracy and validation perplexity across different model scales. Notably, the breakthroughness tasks, which demonstrate sudden performance improvements at the final checkpoints, display a smooth and continuous growth trend along the training trajectory. This observation reinforces the findings of a recent study conducted by~\citet{schaeffer2023emergent}, where they discovered that modifying downstream task metrics results in gradual changes in performance rather than abrupt and unexpected shifts as model scale increases. These results suggest that when examining task performance at a finer level, either through continuous metrics or continuous model checkpoints, task performance largely exhibits a smooth growth pattern in tandem with validation perplexity. Nevertheless, as suggested by \citet{ganguli2022predictability}, accurately predicting the learning curve of a specific task still remains challenging.


\section{Related Work}
\paragraph{Phase change.} \citet{olsson2022context} study induction heads to understand the formation of in-context learning ability. The main finding is that there exists a critical phase chage~\cite{power2022grokking,nanda2022grokking} that forms the in-context learning ability. Our studies are in the same spirit as these work, but we did not discover any phase change for the phenomena we examined; all of them evolve steadily as training progresses. 
\paragraph{(Inverse) scaling laws.} Previous work studies scaling on downstream tasks~\cite{wei2022emergent,srivastava2022beyond}, pre-training data~\cite{hernandez2022scaling}, architectures~\cite{tay2022scaling}, biases~\cite{schwartz2022fewer}, and other domains, such as vision tasks and neural machine translation~\cite{alabdulmohsin2022revisiting}. Our work studies different scaling behaviors over model trajectories. 

Inverse scaling refers to a scaling behavior where increasing the model size leads to worse performance for a downstream task~\cite{perez_mckenzie}. Part of our work intends to understand the distributional shift from small models to large models for language modeling along training trajectories, which overlaps with the theme of inverse scaling.  

\paragraph{Perplexity vs. downstream performance.} Regarding the pre-training/fine-tuning paradigm, \citet{wettig2022should} and \citet{tay2022scaling} find that a lower pre-training perplexity does not necessarily translate to better fine-tuning performance. For zero-shot inference, \citet{saunshi2020mathematical} mathematically shows that doing well in language modeling benefits downstream tasks. On the contrary, \citet{shin2022effect} claims the opposite relationship for in-context learning performance and perplexity when training language models with different corpora, but they only test four downstream tasks on a few model checkpoints. Our work extensively evaluates multiple domains and tasks on both language modeling and downstream tasks across checkpoints of different scales, which entails less variance.

\paragraph{Effective scaling} Several prior studies have focused on effectively scaling models by examining limited compute settings~\cite{Geiping2022CrammingTA}, exploring different objectives~\cite{tay2022scale, artetxe2022role}, and investigating different architecture and training setups~\cite{scao2022language}. This work specifically examines model scales under a unified setting, but the proposed techniques can be applied to other settings as well. 
\section{Conclusion} 
To summarize, our study demonstrates that validation perplexity is a reliable indicator of the behavior of OPT models, regardless of their sizes. Larger models, with increased computational power and capacity, exhibit behavior similar to that of smaller models while also unlocking new phenomena and capabilities as validation perplexity decreases further. However, there are certain exceptional cases where models behave differently, sometimes even in opposite directions, such as in the perplexity of texts generated by contrasting two models. This suggests that the underlying model distributions are not entirely identical at the same perplexity level.

The availability of a larger number of open-sourced model checkpoints, such as those provided by ~\citet{biderman2023pythia}, offers opportunities for interpreting language model behaviors through the analysis of training trajectories. The techniques we propose can be extended to analyze language models trained using different resources and methodologies. Additionally, we leave open questions for future research, such as further exploring the phenomenon of double-descent more in-depth.

\section*{Limitations}
We discuss the limitations of the work as follows:
\begin{itemize}
    \item One major limitation of our work is that we analyze language models pre-trained with the same data, similar training procedures, and the same autoregressive language modeling objective. Our findings may support model families trained in this restricted setting. When comparing models trained with different corpora, such as Neo GPT NEO~\cite{gpt-neo} and BLOOM~\cite{scao2022bloom}, different architectures and objectives, such as retrieval-based language models~\cite{khandelwal2020generalization, zhong2022training, borgeaud2021improving} and sparse models~\cite{fedus2022switch, artetxe2021efficient}, the relationship between validation perplexity and downstream task performance could be more obscure. 
    \item For downstream task evaluation, we only evaluate on multiple-choice tasks, where the evaluation protocol is the most similar to the pre-training objective. Evaluating on generation-based tasks is more messy and hard to scale up, and we will leave it as future work. Another risk is that as we always take aggregated measurements over tasks, it might conceal important patterns of individual tasks.
    \item We do not provide a concrete explanation for the double-descent behavior that consistently occurs during pre-training, nor do we know if it is an artifact of the data, the objective or the optimization process. We consider it an interesting phenomenon and will look more closely into it in future works.
\end{itemize}

\newpage
\section*{Acknowledgement}
We thank Sadhika Malladi for helping out with writing and having insightful discussions on the project with the authors. We thank Tianyu Gao for helping out running experiments on open-text generation in the Appendix. We also thank Stephen Roller, Srini Iyyer, Todor Mihaylov, Xiaochuang Han, and all members of the Princeton NLP group for helpful discussion and valuable feedback. This work was conducted when Mengzhou Xia was interning at Meta Platforms, Inc.

%


\bibliography{anthology,custom}
\clearpage
\appendix
\section{Checkpoint Details}
\label{app:checkpoint_info}
\begin{table*}[t!]
    \centering 
    \begin{tabular}{lccccp{7cm}}
    \toprule
    \textbf{\# Params} & \textbf{LR} & \textbf{Batch Size} & \textbf{\# Steps} & \textbf{\# C\textsc{k}pt} & \textbf{C\textsc{k}pt Steps}  \\ \midrule
    \smallmodel{}    & $6.0e-4$            & \textsc{0.5M}           & 600\textsc{k}         & 36     & 2\textsc{k}, 6\textsc{k}, 10\textsc{k}, 14\textsc{k}, 18\textsc{k}, 22\textsc{k}, 26\textsc{k}, 30\textsc{k}, 34\textsc{k}, 38\textsc{k}, 40\textsc{k}, 60\textsc{k}, 80\textsc{k}, 100\textsc{k}, 120\textsc{k}, 140\textsc{k}, 160\textsc{k}, 200\textsc{k}, 220\textsc{k}, 240\textsc{k}, 260\textsc{k}, 280\textsc{k}, 300\textsc{k}, 320\textsc{k}, 340\textsc{k}, 360\textsc{k}, 380\textsc{k}, 400\textsc{k}, 420\textsc{k}, 440\textsc{k}, 460\textsc{k}, 480\textsc{k}, 500\textsc{k}, 520\textsc{k}, 540\textsc{k}, 560\textsc{k} \\
    \medmodel{}      & $2.0e-4$            & \textsc{1M}             & 300\textsc{k}      & 22      & 2\textsc{k}, 6\textsc{k}, 10\textsc{k}, 14\textsc{k}, 18\textsc{k}, 22\textsc{k}, 26\textsc{k}, 30\textsc{k}, 34\textsc{k}, 38\textsc{k}, 40\textsc{k}, 60\textsc{k}, 80\textsc{k}, 100\textsc{k}, 120\textsc{k}, 140\textsc{k}, 160\textsc{k}, 180\textsc{k}, 200\textsc{k}, 220\textsc{k}, 240\textsc{k}, 260\textsc{k} \\
    \largemodel{}     & $1.2e-4$            & \textsc{2M}             & 150\textsc{k}       & 21        & 2\textsc{k}, 6\textsc{k}, 10\textsc{k}, 14\textsc{k}, 18\textsc{k}, 22\textsc{k}, 26\textsc{k}, 30\textsc{k}, 34\textsc{k}, 38\textsc{k}, 40\textsc{k}, 50\textsc{k}, 60\textsc{k}, 70\textsc{k}, 80\textsc{k}, 90\textsc{k}, 100\textsc{k}, 110\textsc{k}, 120\textsc{k}, 130\textsc{k}, 140\textsc{k}                                                                                              \\
    \xlmodel{}     & $1.0e-4$              & \textsc{4M}             & 75\textsc{k}   & 18            & 2\textsc{k}, 6\textsc{k}, 10\textsc{k}, 14\textsc{k}, 18\textsc{k}, 22\textsc{k}, 26\textsc{k}, 30\textsc{k}, 34\textsc{k}, 38\textsc{k}, 42\textsc{k}, 46\textsc{k}, 50\textsc{k}, 54\textsc{k}, 58\textsc{k}, 62\textsc{k}, 66\textsc{k}, 70\textsc{k}                                                                                                                  \\
    \xxlmodel{}    & $1.0e-4$              & \textsc{4M}             & 75\textsc{k}    & 18           & 2\textsc{k}, 6\textsc{k}, 10\textsc{k}, 14\textsc{k}, 18\textsc{k}, 22\textsc{k}, 26\textsc{k}, 30\textsc{k}, 34\textsc{k}, 38\textsc{k}, 42\textsc{k}, 46\textsc{k}, 50\textsc{k}, 54\textsc{k}, 58\textsc{k}, 62\textsc{k}, 66\textsc{k}, 70\textsc{k}                                                                                                                  \\
    \xxxlmodel{}    & $1.2e-4$             & \textsc{2M}             & 150\textsc{k}      & 32         & 4\textsc{k}, 8\textsc{k}, 12\textsc{k}, 16\textsc{k}, 20\textsc{k}, 24\textsc{k}, 36\textsc{k}, 40\textsc{k}, 44\textsc{k}, 48\textsc{k}, 52\textsc{k}, 56\textsc{k}, 60\textsc{k}, 64\textsc{k}, 68\textsc{k}, 72\textsc{k}, 76\textsc{k}, 80\textsc{k}, 84\textsc{k}, 88\textsc{k}, 92\textsc{k}, 96\textsc{k}, 100\textsc{k}, 104\textsc{k}, 108\textsc{k}, 112\textsc{k}, 120\textsc{k}, 124\textsc{k}, 128\textsc{k}, 132\textsc{k}, 136\textsc{k}, 140\textsc{k} \\ \bottomrule
\end{tabular}
\caption{Checkpoint (Ckpt) information for OPT models. LR denotes learning rate. Note that we take these checkpoints for practical reasons and the distance between checkponts are not evenly spaced. But it should not affect the analysis.}
\label{tab:checkpoint_info}
\end{table*}
We present the checkpoint information in \autoref{tab:checkpoint_info}. OPT models of different sizes are trained with different batch sizes and end up training with different number of steps given the same amount of training tokens. We select early-stage checkpoints every \textsc{4k} steps for evaluation, and enlarge the interval to \textsc{10k} or \textsc{20k} for late stage checkpoints. There are a few checkpoints missing/corrupted from the training process, e.g., \smallmodel{} \textsc{180k}, and we have to eliminate them our evaluation. 

All OPT models are trained with 300\textsc{B} tokens, of which 180\textsc{B} tokens are unique. This training procedure means that OPTs are trained with repeated data, though training with non-repeating data consistently lead to better performance in language modeling and downstream tasks~\cite{lee2022deduplicating, hernandez2022scaling}.

\section{Next-Token Predictions}
\subsection{Data Used in the Main Paper}
\label{app:trend_data}
We use the Gutenberg PG-19~\cite{Rae2020Compressive} subset as the main dataset for analysis in the main paper. This validation subset contains 50 lines of texts, and we take the first 2048 tokens of each line for analysis, resulting in 102350 context-token pairs. We observe similar patterns when evaluated on other validation subsets such as Wikipedia and opensubtitles, and we omit the results for brevity.

\subsection{Trajectory of Other Tokens}
\label{app:other_tokens}
\begin{figure}[h!]
    \centering
    \includegraphics[width=0.49\textwidth]{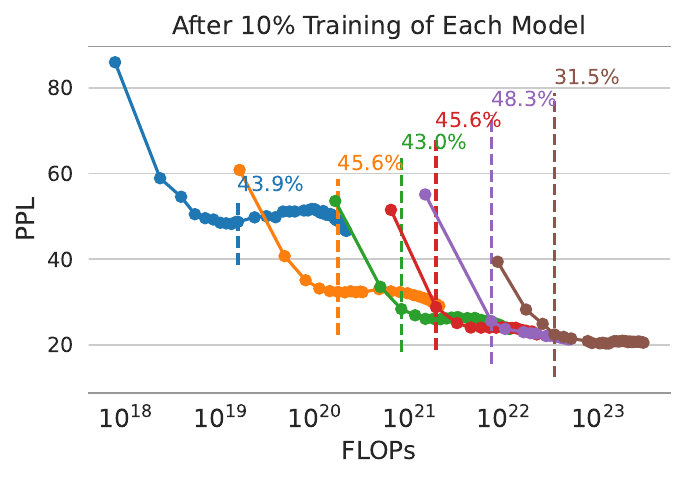} \\
    \includegraphics[width=0.49\textwidth]{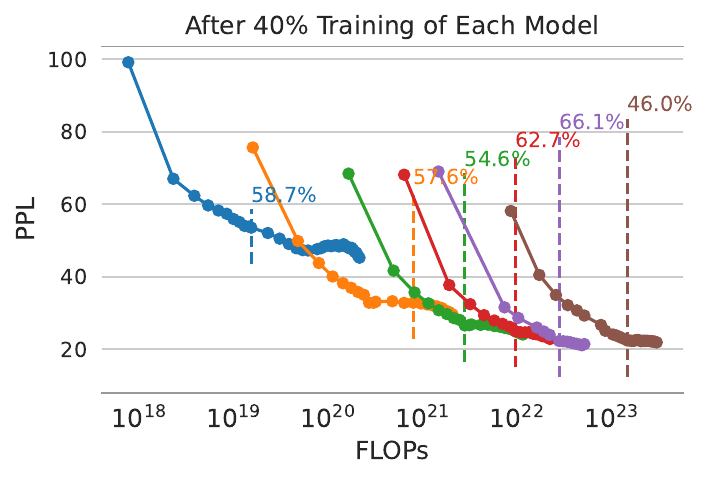}
    \caption{Perplexity of tokens that do not fall into any of the categories. Different models are evaluated on different subsets of tokens selected after 10\% (up) and 40\% (down) of training of individual models. The trends are not consistent across different model sizes.}
    \label{fig:other_tokens}
  \end{figure}

We set our criteria to be relatively strict to make sure that the perplexity trajectory of the selected tokens does present the trend (stagnated/upward/downward) we expect. We present the trajectory of the tokens that do not fall into any of the categories in \autoref{fig:other_tokens}. We find that the trend of these tokens are not consistent across models. After 10\% of training, the curves of \smallmodel{}, \medmodel{}, \largemodel{} present a slight double-descent trend, and for the rest of the models, the curves present a downward/stagnated trend. After 40\% of training, the curves of \smallmodel{} present a slight double-descent trend towards the end, and the curves of other models present a downward/stagnated trend. This suggests that the rest of the tokens might contain a larger variance in their perplexity trajectories.

\subsection{Properties of Stagnated and Upward-Trend Tokens}
\label{app:token_property}
We show an example paragraph in \autoref{tab:example_stagnated_tokens}, where the stagnated tokens are in blue, upward-trend tokens are in red and downward-trend tokens are in green. It's easy to see that stagnated tokens are mostly connecting words, determiners, punctuation and continuation of words. However, we find it hard to characterize the tokens that present an upward-trend in perplexity simply based on token types. We made attempts to further decipher what language properties this subset might entail based on the part-of-speech tags and positions in sequences, and did not observe any obvious patterns when compared to all the tokens in the validation set. One thing we are sure is that the phenomenon of the upward trend in perplexity as well as the double-descent phenomenon on a certain subset of tokens systematically appears across all model sizes. Therefore, this subset of context-token pairs must embody certain intrinsic language properties, which might be beyond our comprehension so far. It would be interesting to do an in-depth analysis in understanding why it happens during pre-training, and how it connects to natural language properties.

\begin{table*}[t]
    \begin{tabular}{@{}p{8cm}p{8cm}@{}}
        \toprule
        \multicolumn{1}{c}{\textbf{\textit{After 10\% training of \medmodel{} model}}}  & \multicolumn{1}{c}{\textbf{\textit{After 10\% training of  \xxxlmodel{} model}}}  \\ \midrule
         {\textcolor{rr}{ Appropri} \textcolor{rr}{ate} \textcolor{black}{;} \textcolor{black}{ pertaining} \textcolor{black}{ to} \textcolor{black}{ the} \textcolor{black}{ subject} \textcolor{black}{.} \textcolor{black}{\textbackslash n} \textcolor{rr}{P} \textcolor{black}{ect} \textcolor{black}{oral} \textcolor{black}{.} \textcolor{rr}{ The} \textcolor{gg}{ bone} \textcolor{gg}{ which} \textcolor{gg}{ forms} \textcolor{gg}{ the} \textcolor{black}{ main} \textcolor{gg}{ rib} \textcolor{black}{ or} \textcolor{black}{ support} \textcolor{rr}{ at} \textcolor{bb}{ the} \textcolor{black}{ forward} \textcolor{gg}{ edge} \textcolor{bb}{ of} \textcolor{rr}{ a} \textcolor{black}{ bird} \textcolor{gg}{'s} \textcolor{rr}{ wing} \textcolor{black}{.} \textcolor{gg}{\textbackslash n} \textcolor{black}{Pers} \textcolor{black}{istent} \textcolor{black}{.} \textcolor{gg}{ Keeping} \textcolor{gg}{ at} \textcolor{gg}{ it} \textcolor{gg}{;} \textcolor{gg}{ determination} \textcolor{black}{ to} \textcolor{rr}{ proceed} \textcolor{black}{.} \textcolor{gg}{\textbackslash n} \textcolor{gg}{Per} \textcolor{black}{pend} \textcolor{bb}{icular} \textcolor{black}{.} \textcolor{gg}{ At} \textcolor{gg}{ right} \textcolor{gg}{ angles} \textcolor{black}{ to} \textcolor{gg}{ a} \textcolor{black}{ surface} \textcolor{gg}{.} \textcolor{rr}{ This} \textcolor{gg}{ term} \textcolor{gg}{ is} \textcolor{black}{ sometimes} \textcolor{black}{ wrongly} \textcolor{black}{ applied} \textcolor{gg}{ in} \textcolor{gg}{ referring} \textcolor{bb}{ to} \textcolor{black}{ an} \textcolor{gg}{ object} \textcolor{gg}{,} \textcolor{rr}{ particularly} \textcolor{black}{ to} \textcolor{rr}{ an} \textcolor{rr}{ object} \textcolor{rr}{ which} \textcolor{gg}{ is} \textcolor{gg}{ vertical} \textcolor{black}{,} \textcolor{black}{ meaning} \textcolor{black}{ up} \textcolor{black}{ and} \textcolor{black}{ down} \textcolor{black}{.} \textcolor{black}{ The} \textcolor{rr}{ blade} \textcolor{black}{ of} \textcolor{gg}{ a} \textcolor{gg}{ square} \textcolor{black}{ is} \textcolor{black}{ perpend} \textcolor{black}{ie} \textcolor{gg}{ular} \textcolor{rr}{ to} \textcolor{gg}{ the} \textcolor{gg}{ handle} \textcolor{rr}{ at} \textcolor{black}{ all} \textcolor{black}{ times} \textcolor{black}{,} \textcolor{gg}{ but} \textcolor{gg}{ the} \textcolor{rr}{ blade} \textcolor{black}{ is} \textcolor{black}{ vertical} \textcolor{black}{ only} \textcolor{gg}{ when} \textcolor{black}{ it} \textcolor{black}{ points} \textcolor{black}{ to} \textcolor{gg}{ the} \textcolor{rr}{ center} \textcolor{gg}{ of} \textcolor{black}{ the} \textcolor{gg}{ earth} \textcolor{gg}{.} \textcolor{black}{\textbackslash n} \textcolor{gg}{P} \textcolor{black}{ern} \textcolor{gg}{icious} \textcolor{black}{.} \textcolor{black}{ Bad} \textcolor{black}{;} \textcolor{black}{ not} \textcolor{black}{ having} \textcolor{gg}{ good} \textcolor{gg}{ features} \textcolor{black}{ or} \textcolor{black}{ possessing} \textcolor{black}{ wrong} \textcolor{rr}{ attributes} \textcolor{bb}{.} \textcolor{black}{\textbackslash n} \textcolor{black}{P} \textcolor{gg}{end} \textcolor{black}{ulum} \textcolor{bb}{.} \textcolor{black}{ A} \textcolor{gg}{ bar} \textcolor{black}{ or} \textcolor{black}{ body} \textcolor{gg}{ suspended} \textcolor{black}{ at} \textcolor{black}{ a} \textcolor{black}{ point} \textcolor{black}{ and} \textcolor{black}{ adapted} \textcolor{gg}{ to} \textcolor{gg}{ swing} \textcolor{black}{ to} \textcolor{gg}{ and} \textcolor{gg}{ fro} \textcolor{black}{.} \textcolor{black}{\textbackslash n} \textcolor{gg}{Per} \textcolor{black}{pet} \textcolor{black}{ual} \textcolor{black}{.} \textcolor{black}{ For} \textcolor{gg}{ all} \textcolor{gg}{ time} \textcolor{gg}{;} \textcolor{gg}{ un} \textcolor{gg}{ending} \textcolor{gg}{ or} \textcolor{gg}{ unlimited} \textcolor{black}{ time} \textcolor{black}{.} \textcolor{gg}{\textbackslash n} \textcolor{rr}{P} \textcolor{black}{hen} \textcolor{black}{omen} \textcolor{black}{a} \textcolor{black}{.} \textcolor{rr}{ Some} \textcolor{black}{ peculiar} \textcolor{gg}{ happening} \textcolor{gg}{,} \textcolor{rr}{ or} \textcolor{black}{ event} \textcolor{black}{,} \textcolor{black}{ or} \textcolor{black}{ object} \textcolor{black}{.} \textcolor{bb}{\textbackslash n} \textcolor{black}{P} \textcolor{gg}{itch} \textcolor{black}{.} \textcolor{black}{ In} \textcolor{gg}{ aviation} \textcolor{black}{ this} \textcolor{black}{ applies} \textcolor{black}{ to} \textcolor{gg}{ the} \textcolor{gg}{ angle} \textcolor{black}{ at} \textcolor{bb}{ which} \textcolor{black}{ the} \textcolor{black}{ blades} \textcolor{black}{ of} \textcolor{black}{ a} \textcolor{gg}{ prope} \textcolor{bb}{ller} \textcolor{gg}{ are} \textcolor{gg}{ cut} \textcolor{black}{.} \textcolor{gg}{ If} \textcolor{black}{ a} \textcolor{black}{ prope} \textcolor{bb}{ller} \textcolor{bb}{ is} \textcolor{gg}{ turned} \textcolor{black}{,} \textcolor{black}{ and} \textcolor{rr}{ it} \textcolor{black}{ moves} \textcolor{black}{ forward} \textcolor{black}{ly} \textcolor{black}{ in} \textcolor{black}{ the} \textcolor{gg}{ exact} \textcolor{black}{ path} \textcolor{black}{ made} \textcolor{gg}{ by} \textcolor{rr}{ the} \textcolor{black}{ angle} \textcolor{black}{,} \textcolor{rr}{ for} \textcolor{black}{ one} \textcolor{black}{ complete} \textcolor{rr}{ turn} \textcolor{black}{,} \textcolor{black}{ the} \textcolor{black}{ distance} \textcolor{gg}{ traveled} \textcolor{rr}{ by} \textcolor{black}{ the} \textcolor{black}{ prope} \textcolor{bb}{ller} \textcolor{black}{ ax} \textcolor{black}{ially} \textcolor{black}{ indicates} \textcolor{black}{ the} \textcolor{gg}{ pitch} \textcolor{rr}{ in} \textcolor{gg}{ feet} \textcolor{gg}{.} \textcolor{black}{\textbackslash n} \textcolor{black}{Pl} \textcolor{rr}{acement} \textcolor{black}{.} \textcolor{black}{ When} \textcolor{gg}{ an} \textcolor{black}{ object} \textcolor{black}{ is} \textcolor{black}{ located} \textcolor{rr}{ at} \textcolor{black}{ any} \textcolor{rr}{ particular} \textcolor{gg}{ point} \textcolor{black}{,} \textcolor{gg}{ so} \textcolor{black}{ that} \textcolor{black}{ it} \textcolor{black}{ is} \textcolor{black}{ operative} \textcolor{black}{ the} \textcolor{black}{ location} \textcolor{black}{ is} \textcolor{gg}{ called} \textcolor{black}{ the} \textcolor{black}{ placement} \textcolor{black}{.} \textcolor{black}{\textbackslash n} \textcolor{black}{Pl} \textcolor{black}{ane} \textcolor{gg}{.} \textcolor{black}{ A} \textcolor{gg}{ flat} \textcolor{gg}{ surface} \textcolor{black}{ for} \textcolor{black}{ supporting} \textcolor{black}{ a} \textcolor{gg}{ flying} \textcolor{black}{ machine} \textcolor{rr}{ in} \textcolor{black}{ the} \textcolor{bb}{ air} \textcolor{bb}{.} \textcolor{black}{ Plane} \textcolor{black}{ of} \textcolor{black}{ movement} \textcolor{rr}{ per} \textcolor{black}{tains} \textcolor{bb}{ to} \textcolor{black}{ the} \textcolor{black}{ imaginary} \textcolor{black}{ surface} \textcolor{black}{ described} \textcolor{gg}{ by} \textcolor{gg}{ a} \textcolor{black}{ moving} \textcolor{gg}{ body}}
        & \textcolor{gg}{ Appropri} \textcolor{rr}{ate} \textcolor{black}{;} \textcolor{black}{ pertaining} \textcolor{gg}{ to} \textcolor{gg}{ the} \textcolor{black}{ subject} \textcolor{rr}{.} \textcolor{gg}{\textbackslash n} \textcolor{black}{P} \textcolor{black}{ect} \textcolor{black}{oral} \textcolor{black}{.} \textcolor{black}{ The} \textcolor{gg}{ bone} \textcolor{black}{ which} \textcolor{black}{ forms} \textcolor{bb}{ the} \textcolor{gg}{ main} \textcolor{black}{ rib} \textcolor{black}{ or} \textcolor{black}{ support} \textcolor{gg}{ at} \textcolor{bb}{ the} \textcolor{black}{ forward} \textcolor{rr}{ edge} \textcolor{bb}{ of} \textcolor{black}{ a} \textcolor{gg}{ bird} \textcolor{bb}{'s} \textcolor{black}{ wing} \textcolor{bb}{.} \textcolor{black}{\textbackslash n} \textcolor{black}{Pers} \textcolor{black}{istent} \textcolor{black}{.} \textcolor{rr}{ Keeping} \textcolor{rr}{ at} \textcolor{black}{ it} \textcolor{gg}{;} \textcolor{black}{ determination} \textcolor{gg}{ to} \textcolor{black}{ proceed} \textcolor{gg}{.} \textcolor{gg}{\textbackslash n} \textcolor{black}{Per} \textcolor{black}{pend} \textcolor{rr}{icular} \textcolor{gg}{.} \textcolor{gg}{ At} \textcolor{gg}{ right} \textcolor{bb}{ angles} \textcolor{rr}{ to} \textcolor{black}{ a} \textcolor{rr}{ surface} \textcolor{gg}{.} \textcolor{rr}{ This} \textcolor{black}{ term} \textcolor{bb}{ is} \textcolor{black}{ sometimes} \textcolor{gg}{ wrongly} \textcolor{black}{ applied} \textcolor{black}{ in} \textcolor{gg}{ referring} \textcolor{gg}{ to} \textcolor{black}{ an} \textcolor{black}{ object} \textcolor{rr}{,} \textcolor{rr}{ particularly} \textcolor{gg}{ to} \textcolor{rr}{ an} \textcolor{black}{ object} \textcolor{rr}{ which} \textcolor{bb}{ is} \textcolor{black}{ vertical} \textcolor{rr}{,} \textcolor{black}{ meaning} \textcolor{black}{ up} \textcolor{gg}{ and} \textcolor{gg}{ down} \textcolor{black}{.} \textcolor{bb}{ The} \textcolor{black}{ blade} \textcolor{bb}{ of} \textcolor{rr}{ a} \textcolor{gg}{ square} \textcolor{gg}{ is} \textcolor{black}{ perpend} \textcolor{black}{ie} \textcolor{gg}{ular} \textcolor{gg}{ to} \textcolor{rr}{ the} \textcolor{gg}{ handle} \textcolor{gg}{ at} \textcolor{gg}{ all} \textcolor{black}{ times} \textcolor{gg}{,} \textcolor{gg}{ but} \textcolor{rr}{ the} \textcolor{gg}{ blade} \textcolor{gg}{ is} \textcolor{gg}{ vertical} \textcolor{gg}{ only} \textcolor{gg}{ when} \textcolor{gg}{ it} \textcolor{gg}{ points} \textcolor{black}{ to} \textcolor{gg}{ the} \textcolor{gg}{ center} \textcolor{gg}{ of} \textcolor{gg}{ the} \textcolor{gg}{ earth} \textcolor{gg}{.} \textcolor{gg}{\textbackslash n} \textcolor{black}{P} \textcolor{black}{ern} \textcolor{rr}{icious} \textcolor{gg}{.} \textcolor{black}{ Bad} \textcolor{gg}{;} \textcolor{rr}{ not} \textcolor{gg}{ having} \textcolor{gg}{ good} \textcolor{black}{ features} \textcolor{bb}{ or} \textcolor{black}{ possessing} \textcolor{gg}{ wrong} \textcolor{black}{ attributes} \textcolor{black}{.} \textcolor{gg}{\textbackslash n} \textcolor{black}{P} \textcolor{rr}{end} \textcolor{rr}{ulum} \textcolor{gg}{.} \textcolor{black}{ A} \textcolor{gg}{ bar} \textcolor{black}{ or} \textcolor{black}{ body} \textcolor{gg}{ suspended} \textcolor{black}{ at} \textcolor{rr}{ a} \textcolor{black}{ point} \textcolor{gg}{ and} \textcolor{gg}{ adapted} \textcolor{gg}{ to} \textcolor{black}{ swing} \textcolor{gg}{ to} \textcolor{gg}{ and} \textcolor{rr}{ fro} \textcolor{black}{.} \textcolor{black}{\textbackslash n} \textcolor{black}{Per} \textcolor{black}{pet} \textcolor{black}{ual} \textcolor{gg}{.} \textcolor{gg}{ For} \textcolor{black}{ all} \textcolor{gg}{ time} \textcolor{gg}{;} \textcolor{gg}{ un} \textcolor{bb}{ending} \textcolor{rr}{ or} \textcolor{black}{ unlimited} \textcolor{black}{ time} \textcolor{bb}{.} \textcolor{gg}{\textbackslash n} \textcolor{rr}{P} \textcolor{black}{hen} \textcolor{rr}{omen} \textcolor{black}{a} \textcolor{gg}{.} \textcolor{black}{ Some} \textcolor{black}{ peculiar} \textcolor{gg}{ happening} \textcolor{rr}{,} \textcolor{black}{ or} \textcolor{black}{ event} \textcolor{black}{,} \textcolor{gg}{ or} \textcolor{gg}{ object} \textcolor{gg}{.} \textcolor{black}{\textbackslash n} \textcolor{bb}{P} \textcolor{black}{itch} \textcolor{gg}{.} \textcolor{rr}{ In} \textcolor{black}{ aviation} \textcolor{gg}{ this} \textcolor{gg}{ applies} \textcolor{rr}{ to} \textcolor{rr}{ the} \textcolor{gg}{ angle} \textcolor{gg}{ at} \textcolor{bb}{ which} \textcolor{gg}{ the} \textcolor{black}{ blades} \textcolor{gg}{ of} \textcolor{gg}{ a} \textcolor{black}{ prope} \textcolor{bb}{ller} \textcolor{black}{ are} \textcolor{black}{ cut} \textcolor{gg}{.} \textcolor{black}{ If} \textcolor{gg}{ a} \textcolor{black}{ prope} \textcolor{gg}{ller} \textcolor{gg}{ is} \textcolor{black}{ turned} \textcolor{rr}{,} \textcolor{gg}{ and} \textcolor{gg}{ it} \textcolor{black}{ moves} \textcolor{black}{ forward} \textcolor{black}{ly} \textcolor{gg}{ in} \textcolor{black}{ the} \textcolor{black}{ exact} \textcolor{gg}{ path} \textcolor{black}{ made} \textcolor{gg}{ by} \textcolor{black}{ the} \textcolor{black}{ angle} \textcolor{rr}{,} \textcolor{gg}{ for} \textcolor{gg}{ one} \textcolor{black}{ complete} \textcolor{gg}{ turn} \textcolor{gg}{,} \textcolor{black}{ the} \textcolor{black}{ distance} \textcolor{rr}{ traveled} \textcolor{gg}{ by} \textcolor{black}{ the} \textcolor{black}{ prope} \textcolor{bb}{ller} \textcolor{black}{ ax} \textcolor{black}{ially} \textcolor{gg}{ indicates} \textcolor{gg}{ the} \textcolor{gg}{ pitch} \textcolor{black}{ in} \textcolor{rr}{ feet} \textcolor{black}{.} \textcolor{gg}{\textbackslash n} \textcolor{gg}{Pl} \textcolor{black}{acement} \textcolor{gg}{.} \textcolor{rr}{ When} \textcolor{gg}{ an} \textcolor{black}{ object} \textcolor{gg}{ is} \textcolor{gg}{ located} \textcolor{gg}{ at} \textcolor{rr}{ any} \textcolor{gg}{ particular} \textcolor{rr}{ point} \textcolor{rr}{,} \textcolor{gg}{ so} \textcolor{bb}{ that} \textcolor{gg}{ it} \textcolor{rr}{ is} \textcolor{gg}{ operative} \textcolor{gg}{ the} \textcolor{black}{ location} \textcolor{gg}{ is} \textcolor{black}{ called} \textcolor{rr}{ the} \textcolor{gg}{ placement} \textcolor{black}{.} \textcolor{gg}{\textbackslash n} \textcolor{gg}{Pl} \textcolor{black}{ane} \textcolor{gg}{.} \textcolor{gg}{ A} \textcolor{gg}{ flat} \textcolor{gg}{ surface} \textcolor{rr}{ for} \textcolor{black}{ supporting} \textcolor{black}{ a} \textcolor{black}{ flying} \textcolor{black}{ machine} \textcolor{black}{ in} \textcolor{gg}{ the} \textcolor{gg}{ air} \textcolor{black}{.} \textcolor{rr}{ Plane} \textcolor{black}{ of} \textcolor{gg}{ movement} \textcolor{black}{ per} \textcolor{bb}{tains} \textcolor{gg}{ to} \textcolor{gg}{ the} \textcolor{gg}{ imaginary} \textcolor{black}{ surface} \textcolor{gg}{ described} \textcolor{gg}{ by} \textcolor{black}{ a} \textcolor{black}{ moving} \textcolor{black}{ body} 
        \\ \bottomrule 
        \end{tabular}
        \caption{An example paragraph to demonstrate tokens that present a stagnating, upward or downward trend after 10\% training of \medmodel{} and \xxxlmodel{} models. Tokens that present an upward trend in perplexity are in \textcolor{rr}{Red}; tokens that present a downward trend are in \textcolor{gg}{Green}; stagnating tokens are in \textcolor{bb}{Blue}. Black tokens do not present a clear trend.}
        \label{tab:example_stagnated_tokens}
    \end{table*}

\subsection{More Explorations on Upward Trends}
\label{app:double_descent}
\begin{figure*}[t]
    \centering
    \includegraphics[width=0.32\textwidth]{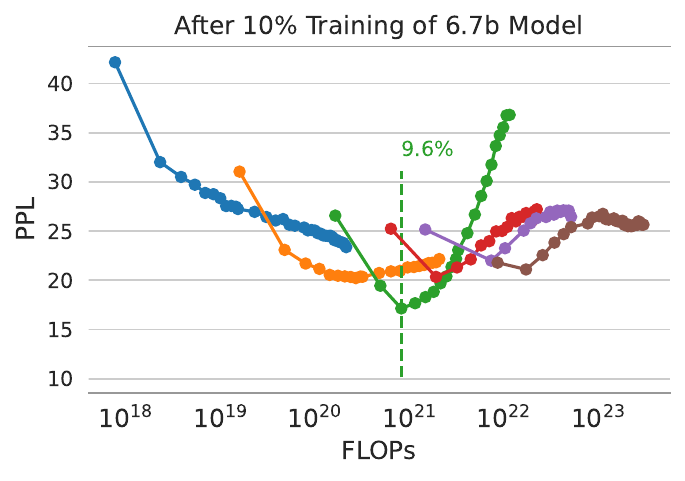}
    \includegraphics[width=0.32\textwidth]{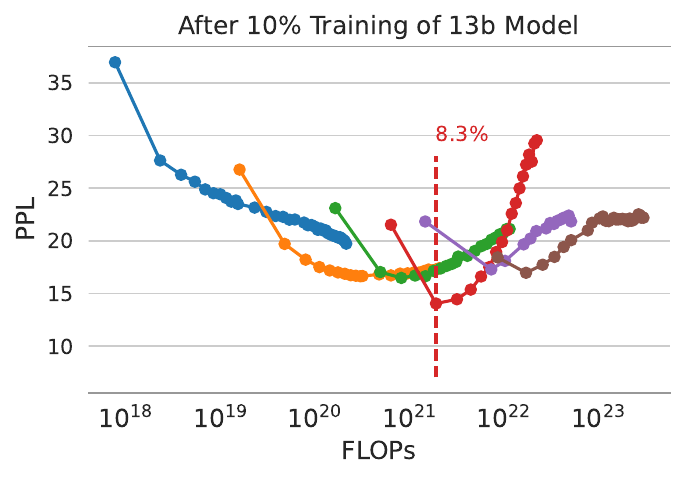}
    \includegraphics[width=0.32\textwidth]{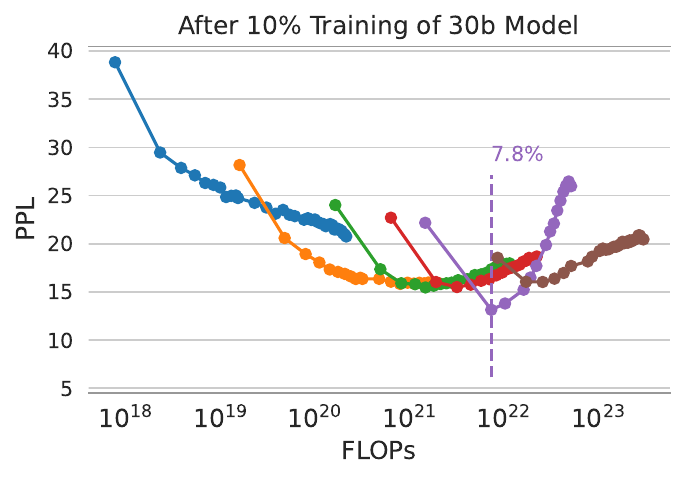}
    \includegraphics[width=0.32\textwidth]{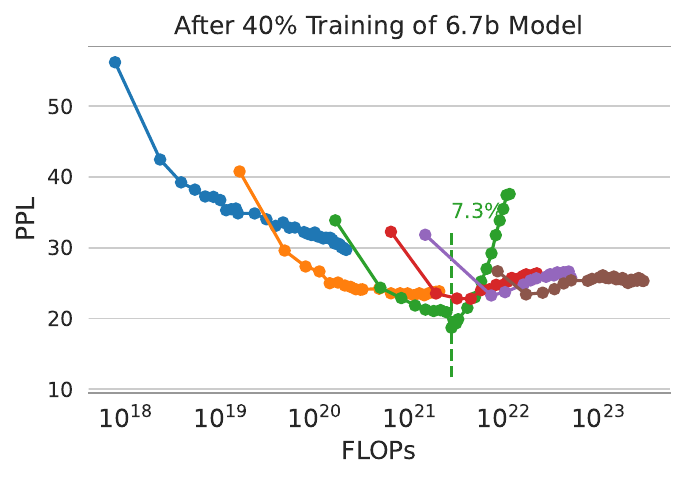}
    \includegraphics[width=0.32\textwidth]{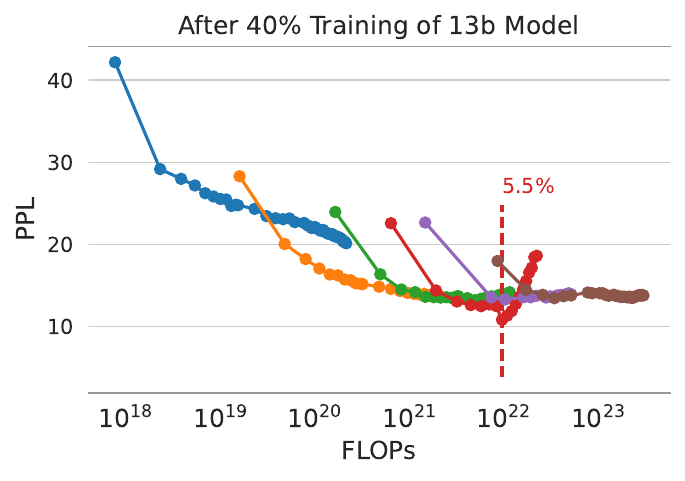}
    \includegraphics[width=0.32\textwidth]{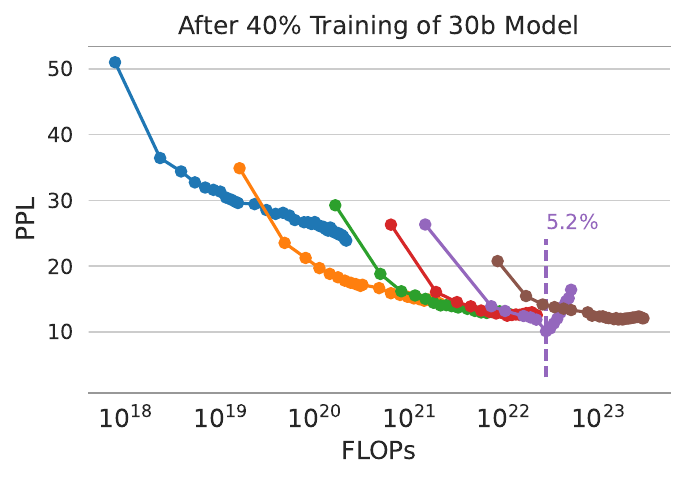}
    \caption{Perplexity of tokens that present an upward trend after 10\% or 40\% of training of the \largemodel{}, \xlmodel{} and \xxlmodel{} models. For each figure, all the models are evaluated on the same subset of tokens.}
    \label{fig:more_upward_trend}
  \end{figure*}

In this section, we explore the subset of tokens that present an upward trend when selected by models of other sizes from the main paper (\largemodel{}, \xlmodel{}, \xxlmodel{}). We present the perplexity trajectory of these tokens in~\autoref{fig:more_upward_trend}. For the subset of tokens selected after 10\% of training of the \largemodel{} model, the larger models' perplexity also increase but only the largest \xxxlmodel{} model presents a double descent behavior where the perplexity declines further. When the tokens are selected after 40\% of training of \largemodel{}, the trends remain similar but the change is mulch more mild. Overall, except the model that is used to select the tokens, the curves of other models present a similar trend, and we will show that these curves overlap with each other almost completely when plotting against validation perplexity in the next subsection. The consistent occurrence of double-descent behavior along the trajectory shows that it's a phenomenon happening universally across the entire autoregressive pre-training process. 

\subsection{Results against Validation Perplexity}
\label{app:ppl_trend}
\begin{figure*}[h!]
    \centering
    \begin{subfigure}[b]{\textwidth}
        \includegraphics[width=0.49\linewidth]{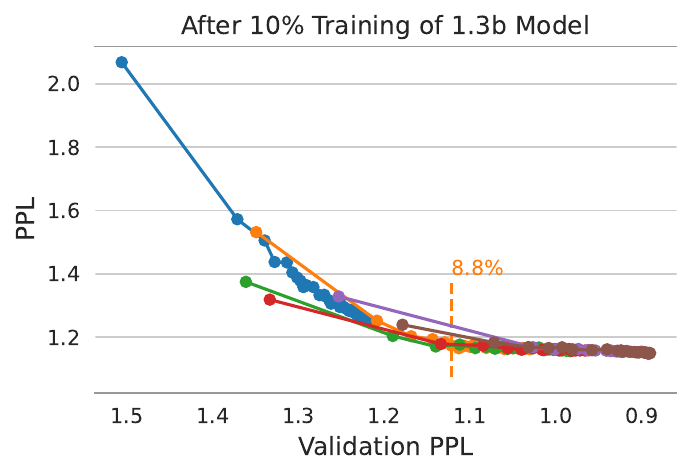} \
        \includegraphics[width=0.49\linewidth]{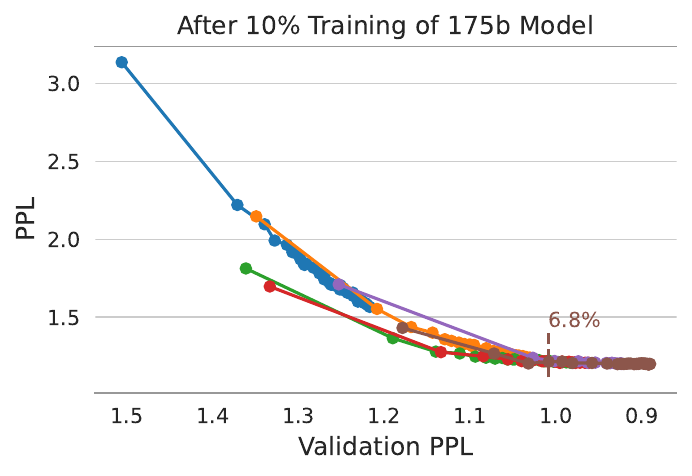}
        \caption{Stagnated Tokens}
         \label{fig:stagnated_tokens_against_ppl}
    \end{subfigure}
    \begin{subfigure}[b]{\textwidth}
        \includegraphics[width=0.49\linewidth]{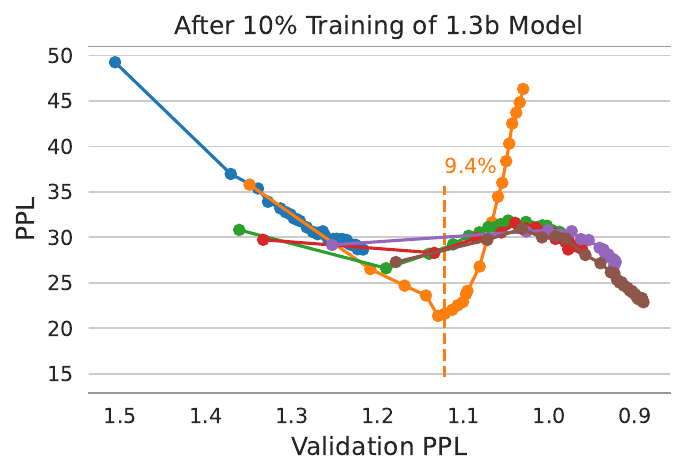} \
        \includegraphics[width=0.49\linewidth]{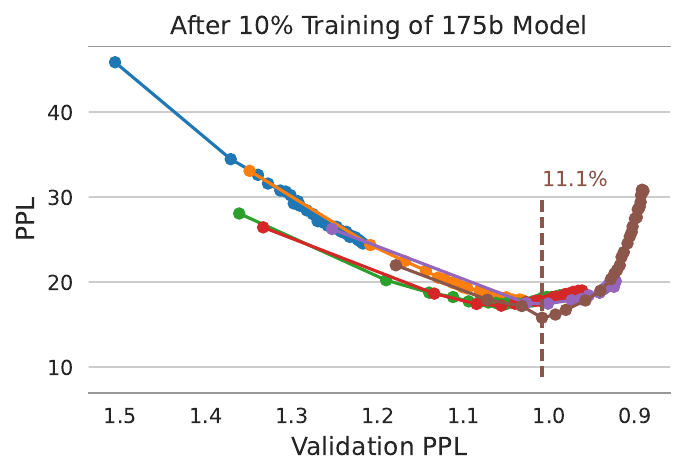}
        \caption{Upward-Trend Tokens}
         \label{fig:upward_trend_tokens_against_ppl}
    \end{subfigure}
    \begin{subfigure}[b]{\textwidth}
        \includegraphics[width=0.49\linewidth]{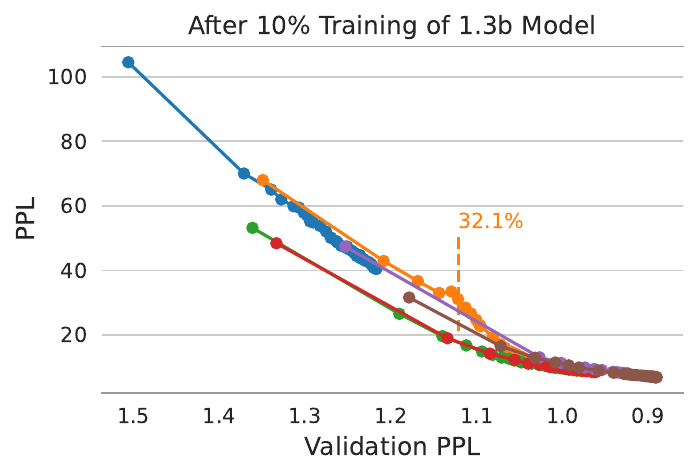} \
        \includegraphics[width=0.49\linewidth]{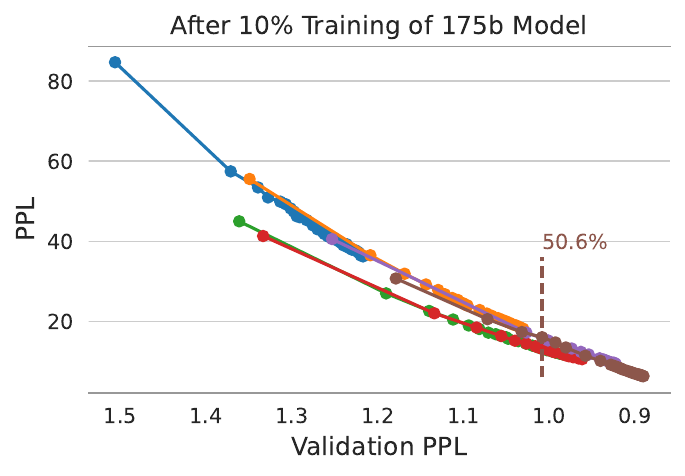}
        \caption{Downward-Trend Tokens}
         \label{fig:downward_trend_tokens_against_ppl}
    \end{subfigure}
    \caption{Perplexity of stagnated tokens, upward-trend tokens and downward-trend tokens against validation perplexity. Curves of different models largely overlap with each other, signifying that validation perplexity is a good indicator of model behaviors along the trajectory, e.g. the double descent-like phenomenon, agnostic to model sizes.}
    \label{fig:tokens_aganinst_ppl}
\end{figure*}

In the main paper, we mostly plot measurements against FLOPs, in this section, we plot the perplexity trajectory of tokens that present different trends against \textbf{validation perplexity} in \autoref{fig:tokens_aganinst_ppl}. These figures present the same series of results as \autoref{fig:stagnated_tokens} and \autoref{fig:upward_trend_tokens}, except that the x-axis is validation perplexity. As mentioned in \autoref{sec:models_and_setup}, we use the aggregated perplexity of a number of subsets as the validation perplexity. 

From \autoref{fig:tokens_aganinst_ppl}, we see that given a similar level of validation perplexity, for different subsets of tokens, the trajectories of models across sizes overlap well with each other, suggesting that the predictions for these tokens are similar across model scales at a fixed level of validation perplexity. The only exception is the upward-trend tokens selected after 10 \% training of \medmodel{}, where evaluating with \medmodel{} presents a clear upward trend as the validation perplexity increases, while the models larger than \medmodel{} present a overlapping double descent-like trend. This indicates that the underlying distribution of models at the same level of perplexity are largely similar but could differ in edge cases.  

These results lays the foundation for downstream task evaluations, which heavily relies on the pre-training objective for evaluation. 

\section{Sequence-Level Generation}
\subsection{Details of Corrupted Datasets}
\label{app:corrupted_dataset}
We corrupt texts from the opensubtitle subset of the validation set by replacing $p\%$ tokens (subwords) with randomly sampled tokens in the sequences. We cap the max length of a sequence to be 100, though changing max length values does not affect the conclusion. Although the perplexity on these corrupted sequences is extremely high, especially when the replacement rate is high, it is still much lower than a truely random model (the perplexity of a random model should be $|V|$ where $V$ is the vocabulary), even for the fully corrupted dataset. It reflects that larger language models are better at exploiting random patterns to produce in-distribution contents than smaller counterparts. We also tried other ways of corruption, such as deleting, inserting, repeating tokens/spans, and all these corruptions result in similar scaling trends.

\subsection{Comparison to \citet{li2022contrastive}}
\label{app:contrastive_decoding}
Our decoding approach is similar to the contrastive decoding method (CD) proposed in \citet{li2022contrastive}, though initially for completely different purposes. The difference between the two methods is in the subtraction space. The contrastive score in CD is defined by dividing the expert probability over amateur probability, which is equivalent to subtraction in the log probability space. Our approach operates subtraction in the probability space directly, ruling out unlikely options where the small model is much more confident than the large model directly. Due to this different design choice, our approach does not need to add the adaptive plausibility restriction, nor involve any additional hyperparameter. Subtraction in the probability space easily eliminates the false positive cases.

We initially propose the approach to decoding sequences that small models favor more than large models to understand the distributional shift across model scales, while contrastive decoding proposed in \citet{li2022contrastive} is a general open-generation approach. Nonetheless, our approach could be an effective and lightweight alternative for open-ended generation without the need to adjust hyperparameters. In Appendix \ref{app:open_generation}, we show that our approach outperforms nucleus sampling on \mauve{} scores.

\subsection{Generation Quality}
\label{app:generation_quality}
\begin{figure*}
    \centering
\includegraphics[width=0.49\linewidth]{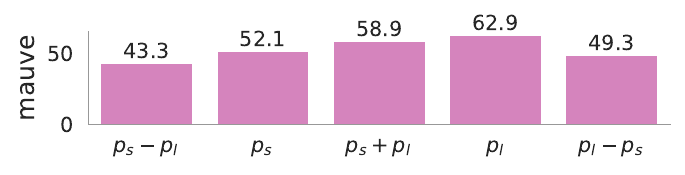}
\includegraphics[width=0.49\linewidth]{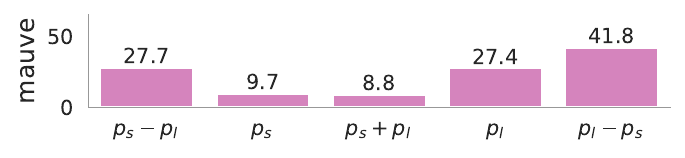}
\caption{\mauve{} scores (the higher, the better) on sequences with a maximum length of 50.} 
\label{fig:gen2}
\end{figure*}

To have a better understanding of the overall quality of the generated sequences, we evaluate these sequences decoded with each configuration in \autoref{fig:gen} using \mauve{} scores~\cite{pillutla2021mauve}. We present the \mauve{} scores in \autoref{fig:gen2} . Our generation protocol is slightly different from the standard open-ended generation practices in that we only use 5 tokens as prompts for generation, while usually at least 128 tokens are used~\cite{rankgen22, su2022contrastive, li2022contrastive}. Using fewer tokens as prompts leads to a higher generation diversity, and the generated distribution could be largely different from the ground-truth sentences. Therefore, we find that the \mauve{} scores of our generated sequences are much lower than reported in open-ended generation literature. 

Comparing the two decoding protocols, subtraction between two distributions ($p_s - p_l$ and $p_l - p_s$) leads to a better generation quality than summing the two ($p_s + p_l$) for greedy sampling, but vice versa for nucleus sampling. To verify the effectiveness of the approach, we compare it to nucleus sampling with standard open-generation protocols in Appendix \ref{app:open_generation}.


\subsection{Open-ended Generation Evaluation}
\label{app:open_generation}
We follow the generation protocol in \citet{rankgen22} for open-ended generation, where we generate sequences with a maximum length of $128$ given contexts that have $256$ tokens. We decode sequences based on either $p_l-p_s$ or $p_l$ with greedy decoding or nucleus sampling ($p=0.9$) and evaluate the quality of the generation with \mauve{} scores. 

We present the results in \autoref{tab:open_generation}. Consistently, our approach to subtracting the probability from a small model from a large model outperforms nucleus sampling with one single model consistently, indicating that our approach has the potential to serve as an effective general decoding method for open-ended generation. 

\begin{table}[]
    \centering
    \begin{tabular}{lrr}
    \toprule
                  & \textbf{greedy} & \textbf{nucleus} \\ \midrule
    350m          & 0.065  & 0.807   \\
    350m-125m     & 0.795  & \textbf{0.852}   \\ \midrule
    1.3b          & 0.164  & 0.877   \\
    1.3b-125m     & 0.851  & \textbf{0.890}   \\ 
    1.3b-350m     & 0.888  & 0.886   \\ \midrule
    2.7b      & 0.237  & 0.832   \\
    2.7b-125m & 0.815  & \textbf{0.851}   \\
    2.7b-350m & 0.846  & 0.843   \\ \bottomrule
    \end{tabular}
    \caption{\mauve{} scores of generations following open-generation protocols. Nucleus sampling on an interpolated distribution ($p_l - p_s$) consistently outperforms decoding with a single model ($p_l$).}
    \label{tab:open_generation}
\end{table}

\subsection{Generating Longer Sequences}
\label{app:longer_sequences}

\begin{figure*}[h!]
    \centering
\includegraphics[width=0.8\linewidth]{images/legend.pdf}

\includegraphics[width=0.49\linewidth]{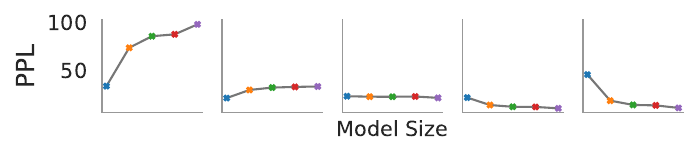} 
\includegraphics[width=0.49\linewidth]{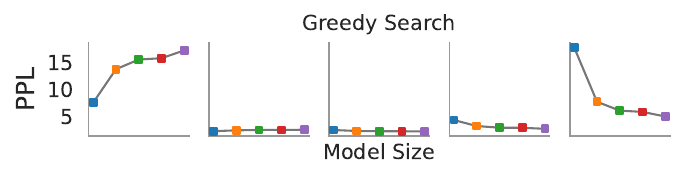} 
\includegraphics[width=0.49\linewidth]{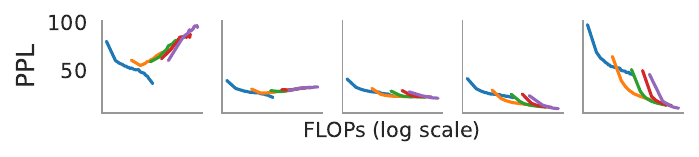}
\includegraphics[width=0.49\linewidth]{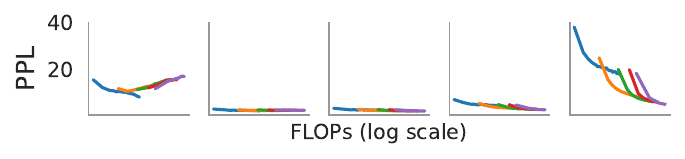}
\includegraphics[width=0.49\linewidth]{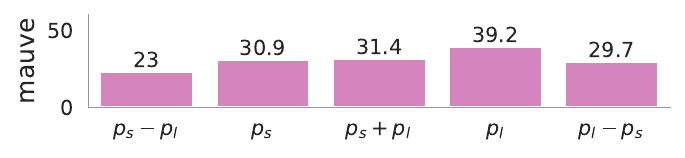}
\includegraphics[width=0.49\linewidth]{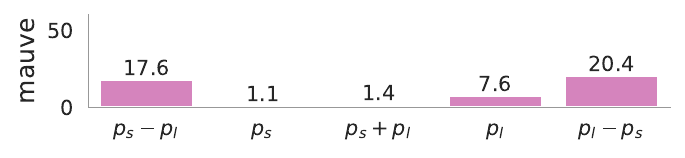}
\caption{Greedy search and nucleus sampling results with generations of a length of 100.}
\label{fig:gen_maxl100}
\end{figure*}

\begin{figure*} [h!]
    \centering
\includegraphics[width=0.8\linewidth]{images/legend.pdf}

\includegraphics[width=0.49\linewidth]{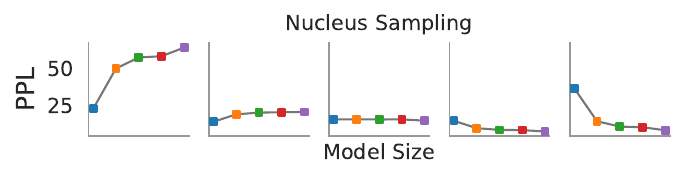} 
\includegraphics[width=0.49\linewidth]{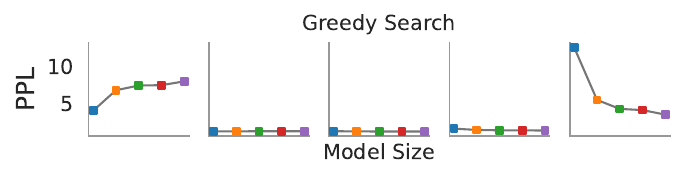} 
\includegraphics[width=0.49\linewidth]{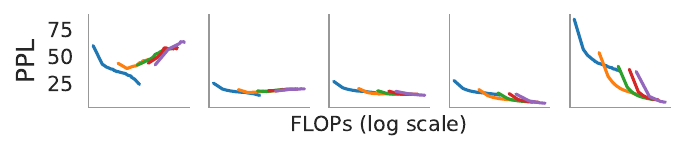}
\includegraphics[width=0.49\linewidth]{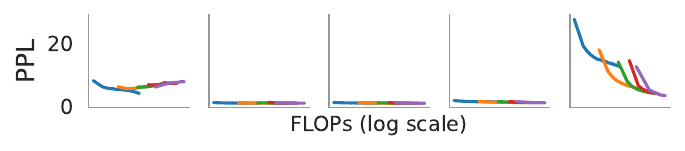}
\includegraphics[width=0.49\linewidth]{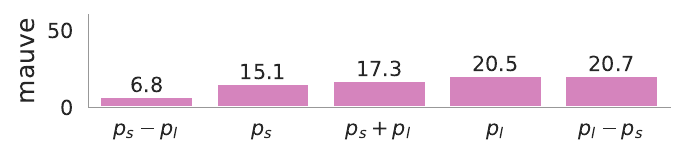}
\includegraphics[width=0.49\linewidth]{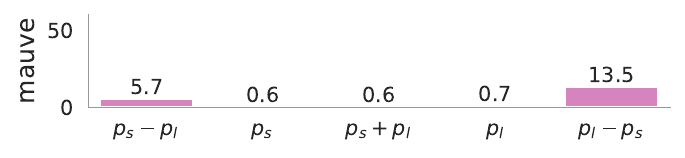}
\caption{Greedy search and nucleus sampling results with generations of a length of 500.}
\label{fig:gen_maxl500}
\end{figure*}

We extend the study to generate longer sequences up to 100 and 500 tokens, and we present perplexity trajectories in \autoref{fig:gen_maxl100} and \autoref{fig:gen_maxl500}, respectively. We find that the inverse scaling trend across model sizes and the opposite perplexity trend between the \smallmodel{} and \xxlmodel{} also hold for longer sequences. \mauve{} scores on generated sequences of different lengths are largely consistent. The longer the decoded sequences are, the worse the overall quality.

\begin{table*}[h!]
    \centering
    \def\arraystretch{0.7}
    \begin{tabular}{@{}p{1.0cm}p{7.2cm}p{7.2cm}@{}} \toprule
        \small{\textbf{Dist.}} & \multicolumn{1}{c}{\textbf{\small{Greedy Search}}}  & \multicolumn{1}{c}{\textbf{\small{Nucleus Sampling}}}  \\ \midrule
    & \small{\textbf{\textit{A girl (Lisbeth Salander) has ...}}} & \small{\textbf{\textit{A girl (Lisbeth Salander) has ...}}} \\
    \cmidrule(lr){2-2} \cmidrule(lr){3-3}
    $p_s - p_l$  & \small{ just discovered a new way to get her hair done!\textbackslash n\textbackslash nLisba is a blonde beauty who has been living her dream for quite some time now.\textbackslash n\textbackslash nLisba was recently spotted wearing a red and white wig and a black and blue striped suit that she wore in an Instagram video. The blonde beauty is currently living in the UK with her husband, Matt, with the help of their daughter Lizz.\textbackslash n\textbackslash nLizz and Matt have been dating for a while but have been dating} & \small{ left the grandpa home he's lived to preserve for her.\textbackslash nLisbin and her daughter Dylan Joanna (Arista Welch-Collinson) do everything they can to protect their sister.\textbackslash nBut unlike ever before their young girl fills it all with grief with every form of loss.\textbackslash nAs she learns Dylan isn't here anymore and acknowledges he's changed his mind, Daley finds herself falling back on the same old rules again.\textbackslash nYellen's been teaching the inane Lyle}  \\
    \cmidrule(lr){2-2} \cmidrule(lr){3-3}
    $p_s$         & \small{ a crush on a guy (Lisbeth Salander) and she's not sure what to do about it.\textbackslash n\textbackslash nShe's a girl who's been in love with a guy (Lisbeth Salander) for a while, but she's not sure what to do about it.\textbackslash n\textbackslash nShe's a girl who's been in love with a guy (Lisbeth Salander) for a while, but she's not sure what to do about it.\textbackslash n\textbackslash nShe}      & \small{ just discovered\textbackslash nthat it’s not an actual\textbackslash nworm! Thanks to the Halloween\textbackslash nMirror campaign, she was discovered in the\textbackslash ngoldeneye-buxco-only prologue of the main\textbackslash ngame, for her interest in science and
    fantasia.\textbackslash n\textbackslash nMalcolm Young: There are four bugs on here.We’re working on them, though, because they’re
    incredibly busy and we’re both making what are\textbackslash nexpected\textbackslash nto}                                                                                              \\
    \cmidrule(lr){2-2} \cmidrule(lr){3-3}
    $p_s + p_l$ & \small{ a crush on a guy (Alexander Skarsgard) and he's a bit of a jerk.\textbackslash nI think you mean Lisbeth Salander.\textbackslash n\textbackslash nI think you mean Lisbeth Salander.\textbackslash nI think you mean Lisbeth Salander.\textbackslash nI think you mean Lisbeth Salander.\textbackslash nI think you mean Lisbeth Salander.\textbackslash nI think you mean Lisbeth Salander.\textbackslash nI think you mean Lisbeth Salander.\textbackslash nI think you} & \small{ a feeling that there’s something more to computer games than what we see in them. Turns out, there is something more to computer games. But it’s not “aliens,” “moles,” or “spies.” It’s about a virtual world that we can all be part of if we so choose.\textbackslash n\textbackslash nWe believe in calling that world Virtual Reality.\textbackslash n\textbackslash nThe game is called THE PREMIER.} \\
    \cmidrule(lr){2-2} \cmidrule(lr){3-3}
    $p_l$         & \small{ a crush on a guy (Alexander Skarsgard) who is a computer hacker.\textbackslash nThe Girl with the Dragon Tattoo?\textbackslash nYup.<s>}  & \small{has been attacked at her apartment by three men. They rape her and beat her. She later finds herself at a center for the mentally disabled. There, she has... See full summary »\textbackslash nThe beginning of a story. The first nymphomaniac to admit that she has a problem, Helga refuses to seek treatment because she sees it as a surrender to defeat. However, it's... See full summary »\textbackslash nA young man has just killed a drunken girl in a subway station.} \\
    \cmidrule(lr){2-2} \cmidrule(lr){3-3}
    $p_l - p_s$  &  \small{ sex for the first and last times in The Girl with the Dragon Tattoo.\textbackslash nI don't remember that in the book.\textbackslash nIt's implied. She says something like "I've never done this before" when they're having sex for the first time. And when she's having sex with her boyfriend for the last, she says something along the same lines.</s>
    } & \small{her face ripped open in The Girl with the Dragon Tattoo. That one still disturbs me.
    The third movie not as much, but that scene was still disturbing</s>} \\ \bottomrule
    \end{tabular}
    \caption{Generated examples with greedy decoding and nucleus sampling under different configurations. The prompt is \textit{A girl (Lisbeth Salander) has}.}
    \label{tab:generations_v2}
\end{table*}

\begin{table*}[t]
    \centering
    \def\arraystretch{0.7}
    \begin{tabular}{@{}p{1.0cm}p{7.2cm}p{7.2cm}@{}} \toprule
        \small{\textbf{Dist.}} & \multicolumn{1}{c}{\textbf{\small{Greedy Search}}}  & \multicolumn{1}{c}{\textbf{\small{Nucleus Sampling}}}  \\ \midrule
    & \small{\textbf{\textit{Now in private practice together, ...}}} & \small{\textbf{\textit{Now in private practice together, ...}}} \\
    \cmidrule(lr){2-2} \cmidrule(lr){3-3}
    $p_s - p_l$  & \small{ I'm going through the same process. I've never had any issues.\textbackslash nI've had the same issue too! I'm trying not get into any of the practices because it's so hard. But now I can't do anything because of it :(  I'm hoping to do something with my time and money to get some help!\textbackslash nThat's really sad! Hopefully I can get some help!  I hope to get some advice from someone who knows how to help me out, and that they} & \small{I can confirm it works pretty perfectly on My"EBM  Used by me if I ever need\textbackslash nGreat news :)</s>} \\
    \cmidrule(lr){2-2} \cmidrule(lr){3-3}
    $p_s$         & \small{I'm a big fan of the "I'm a big fan of the "I'm a big fan of the "I'm a big fan of the "I'm a big fan of the "I'm a big fan of the "I'm a big fan of the "I'm a big fan of the "I'm a big fan of the "I'm a big fan of the "I'm a big fan of the "I'm a big fan of the "I'm a big fan} & \small{a firm working on management strategies for retailing for software, designing, and engineering complex healthcare facilities, and leading multi-channel providers in addition to providing a variety of consulting services. Experience in all stages of PR is critical to have.\textbackslash n\textbackslash nThis 3.3 year-term contract includes a wide range of consultant training, including training for email and email integration, and three-way calls for projects.\textbackslash n\textbackslash nAt Microsoft Healthcare, we have a broad selection of technical leadership and support teams for our healthcare}                                                                                              \\
    \cmidrule(lr){2-2} \cmidrule(lr){3-3}
    $p_s + p_l$ & \small{ I have the pleasure of working with a number of clients who have been referred to me by my colleagues. I have been able to help them with their legal issues and I have been able to help them with their personal issues.\textbackslash n\textbackslash nI have been able to help them with their legal issues and I have been able to help them with their personal issues.\textbackslash n\textbackslash nI have been able to help them with their legal issues and I have been able to help them with their personal issues.\textbackslash n\textbackslash nI have} & \small{Father Harry Thomas, a faculty member at Canisius College, and Father Christopher Cooney, pastor at Holy Redeemer Church in Lancaster, are a good team. The two have collaborated on two traditional healing classes for children since the spring of 2016. Their latest effort, followed by Father John Clifford, pastor at Christ the King Church in Canisius, has taken the call of mercy to the study level. Beginning September 24, Christ the King Church, Canisius, will host “Pope} \\
    \cmidrule(lr){2-2} \cmidrule(lr){3-3}
    $p_l$         & \small{ Dr. David and Dr. David are a husband and wife team of chiropractors who specialize in the treatment of back pain, neck pain, headaches, and other musculoskeletal problems. They are dedicated to providing the highest quality of care to their patients in a comfortable, friendly, and professional environment.\textbackslash n\textbackslash nDr. David is a graduate of the Palmer College of Chiropractic in Davenport, Iowa. He has been practicing in the greater San Diego area since 1995. He} & \small{Spencer and Field with many years of combined practice are passionate about delivering high quality health care to the people of Texas. "Our mission is to empower you and your family to reach your health and wellness goals through nutritional and lifestyle changes. We take a whole-family approach to care and believe that true health is created from the inside out. If you're ready to feel better, we want to be part of your journey"</s>}\\
    \cmidrule(lr){2-2} \cmidrule(lr){3-3}
    $p_l - p_s$  &  \small{ Drs. Michael J. Gazzaniga and David A. Eagleman have written a new book that explores what they believe are some fundamental mysteries of the human mind. In The Brain: The Story of You, they argue that the brain is not just the seat of our thoughts and emotions but also of who we are as people.\textbackslash n\textbackslash nIn this excerpt from the introduction, the authors explain why they wrote the book and what they hope readers take away.\textbackslash nThe Brain: The...</s>} & \small{the pair focus their legal expertise on helping immigrant families and individuals resolve a wide range immigration matters, including deportation defense, asylum, naturalization (citizenship), removal defense, consular processing (visas), VAWA petitions (domestic violence) as well as deportation and removal proceedings, appeals and motions before immigration court, administrative motions in immigration court, removal orders and waivers of inadmissability. Both attorneys are admitted to the Maryland State Bar as well as the District of Columbia Court of appeals} \\ \bottomrule
    \end{tabular}
    \caption{Generated examples with greedy decoding and nucleus sampling under different configurations. The prompt is \textit{Now in private practice together,}.}
    \label{tab:generations_v3}
\end{table*}
\subsection{Examples of Generated Sequences}
\label{app:generated_sequences}
We present more examples of generated sequences in \autoref{tab:generations_v2} and \autoref{tab:generations_v3}. Similar to \autoref{tab:generations}, we find that nucleus sampling with $p_l, p_l - p_s$ and greedy search with $p_l - p_s$ constantly generate high-quality sequences. Greedy decoding $p_s - p_l$ generates mediocre sequences that are largely grammatical and fluent, but less coherent and sometimes contain hallucinations.

\subsection{Validation Perplexity vs. Perplexity of Generated Texts}
\label{app:genppl_ppl}
\begin{figure*} [h!]
    \centering
\includegraphics[width=0.4\linewidth]{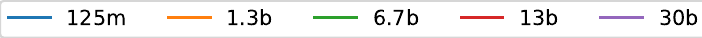} \\
\includegraphics[width=0.49\linewidth]{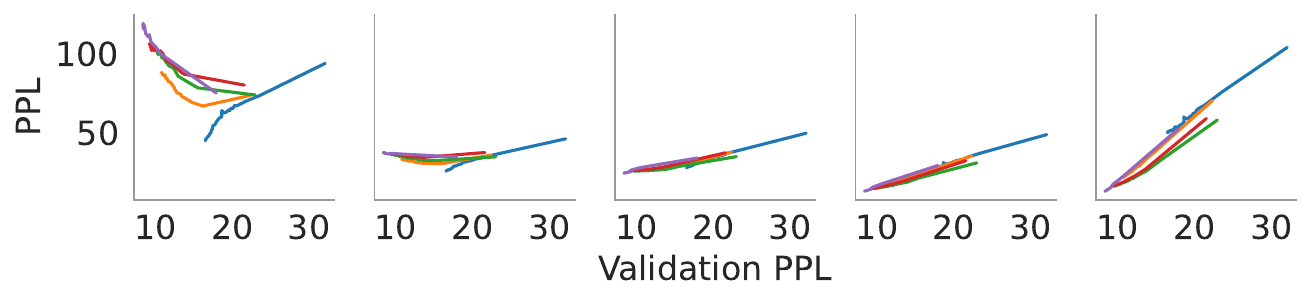} 
\includegraphics[width=0.49\linewidth]{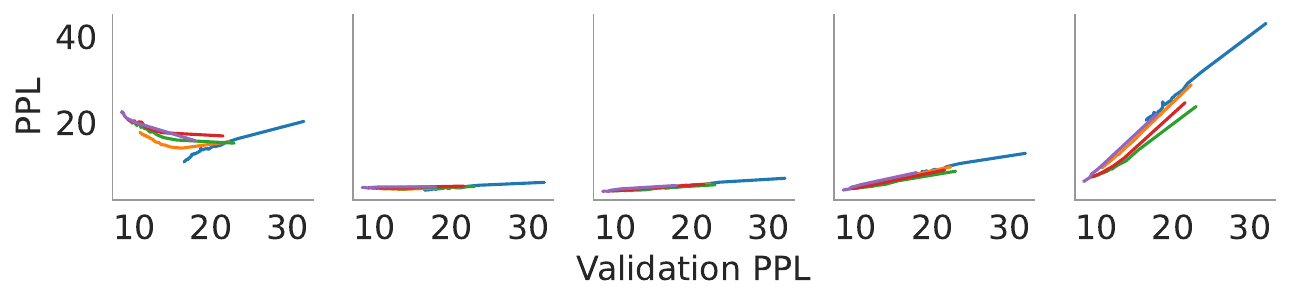} \\
\vspace{-0.2cm}
\includegraphics[width=0.49\linewidth]{images/gen_xlabel.pdf}
\includegraphics[width=0.49\linewidth]{images/gen_xlabel.pdf}
\vspace{-0.3cm}
\caption{Validation perplexity vs. perplexity of generated texts. We find that models of different scales do not have the same perplexity on the generated texts when decoded with $p_s - p_l$ or $p_l - p_s$ given the same validation perplexity, but they largely align when decoded with other configurations.}
\label{fig:genppl_ppl}
\end{figure*}

We plot the perpelxity trajectory of generated texts against validation perplexity in \autoref{fig:genppl_ppl}. The trajectories largely align well across model sizes for $p_s$, $p_s + p_l$ and $p_l$ but diverge in the case of $p_l - p_s$ and $p_s - p_l$. This indicates that the underlying distributions of different-sized models given the same perplexity are similar but not exactly identical.

\section{Downstream Tasks}
\subsection{Task Selection and Evaluation}
\label{app:task_selection}
Out of comuputational considerations, we only evaluate multiple-choice tasks that have fewer than 1000 evaluation examples. The list of selected tasks is shown in \autoref{tab:bigbench_tasks}. We report 2-shot in-context learning performance on the \texttt{default} set of each \bigbench{} dataset.

\begin{table*}
    \centering
    \resizebox{\linewidth}{!}{
    \begin{tabular}{lll} \toprule
        \href{https://github.com/google/BIG-bench/tree/main/bigbench/benchmark_tasks/anachronisms}{\texttt{anachronisms}} & \href{https://github.com/google/BIG-bench/tree/main/bigbench/benchmark_tasks/analogical\_similarity}{\texttt{analogical\_similarity}} & \href{https://github.com/google/BIG-bench/tree/main/bigbench/benchmark_tasks/analytic\_entailment}{\texttt{analytic\_entailment}} \\
\href{https://github.com/google/BIG-bench/tree/main/bigbench/benchmark_tasks/authorship\_verification}{\texttt{authorship\_verification}} & \href{https://github.com/google/BIG-bench/tree/main/bigbench/benchmark_tasks/causal\_judgment}{\texttt{causal\_judgment}} & \href{https://github.com/google/BIG-bench/tree/main/bigbench/benchmark_tasks/cause\_and\_effect}{\texttt{cause\_and\_effect}} \\
\href{https://github.com/google/BIG-bench/tree/main/bigbench/benchmark_tasks/code\_line\_description}{\texttt{code\_line\_description}} & \href{https://github.com/google/BIG-bench/tree/main/bigbench/benchmark_tasks/common\_morpheme}{\texttt{common\_morpheme}} & \href{https://github.com/google/BIG-bench/tree/main/bigbench/benchmark_tasks/conceptual\_combinations}{\texttt{conceptual\_combinations}} \\
\href{https://github.com/google/BIG-bench/tree/main/bigbench/benchmark_tasks/crash\_blossom}{\texttt{crash\_blossom}} & \href{https://github.com/google/BIG-bench/tree/main/bigbench/benchmark_tasks/crass\_ai}{\texttt{crass\_ai}} & \href{https://github.com/google/BIG-bench/tree/main/bigbench/benchmark_tasks/cryobiology\_spanish}{\texttt{cryobiology\_spanish}} \\
\href{https://github.com/google/BIG-bench/tree/main/bigbench/benchmark_tasks/dark\_humor\_detection}{\texttt{dark\_humor\_detection}} & \href{https://github.com/google/BIG-bench/tree/main/bigbench/benchmark_tasks/date\_understanding}{\texttt{date\_understanding}} & \href{https://github.com/google/BIG-bench/tree/main/bigbench/benchmark_tasks/disambiguation\_qa}{\texttt{disambiguation\_qa}} \\
\href{https://github.com/google/BIG-bench/tree/main/bigbench/benchmark_tasks/discourse\_marker\_prediction}{\texttt{discourse\_marker\_prediction}} & \href{https://github.com/google/BIG-bench/tree/main/bigbench/benchmark_tasks/emoji\_movie}{\texttt{emoji\_movie}} & \href{https://github.com/google/BIG-bench/tree/main/bigbench/benchmark_tasks/empirical\_judgments}{\texttt{empirical\_judgments}} \\
\href{https://github.com/google/BIG-bench/tree/main/bigbench/benchmark_tasks/english\_russian\_proverbs}{\texttt{english\_russian\_proverbs}} & \href{https://github.com/google/BIG-bench/tree/main/bigbench/benchmark_tasks/entailed\_polarity}{\texttt{entailed\_polarity}} & \href{https://github.com/google/BIG-bench/tree/main/bigbench/benchmark_tasks/entailed\_polarity\_hindi}{\texttt{entailed\_polarity\_hindi}} \\
\href{https://github.com/google/BIG-bench/tree/main/bigbench/benchmark_tasks/evaluating\_information\_essentiality}{\texttt{evaluating\_information\_essentiality}} & \href{https://github.com/google/BIG-bench/tree/main/bigbench/benchmark_tasks/fantasy\_reasoning}{\texttt{fantasy\_reasoning}} & \href{https://github.com/google/BIG-bench/tree/main/bigbench/benchmark_tasks/figure\_of\_speech\_detection}{\texttt{figure\_of\_speech\_detection}} \\
\href{https://github.com/google/BIG-bench/tree/main/bigbench/benchmark_tasks/hhh\_alignment}{\texttt{hhh\_alignment}} & \href{https://github.com/google/BIG-bench/tree/main/bigbench/benchmark_tasks/hinglish\_toxicity}{\texttt{hinglish\_toxicity}} & \href{https://github.com/google/BIG-bench/tree/main/bigbench/benchmark_tasks/human\_organs\_senses}{\texttt{human\_organs\_senses}} \\
\href{https://github.com/google/BIG-bench/tree/main/bigbench/benchmark_tasks/identify\_math\_theorems}{\texttt{identify\_math\_theorems}} & \href{https://github.com/google/BIG-bench/tree/main/bigbench/benchmark_tasks/identify\_odd\_metaphor}{\texttt{identify\_odd\_metaphor}} & \href{https://github.com/google/BIG-bench/tree/main/bigbench/benchmark_tasks/implicatures}{\texttt{implicatures}} \\
\href{https://github.com/google/BIG-bench/tree/main/bigbench/benchmark_tasks/implicit\_relations}{\texttt{implicit\_relations}} & \href{https://github.com/google/BIG-bench/tree/main/bigbench/benchmark_tasks/intent\_recognition}{\texttt{intent\_recognition}} & \href{https://github.com/google/BIG-bench/tree/main/bigbench/benchmark_tasks/international\_phonetic\_alphabet\_nli}{\texttt{international\_phonetic\_alphabet\_nli}} \\
\href{https://github.com/google/BIG-bench/tree/main/bigbench/benchmark_tasks/irony\_identification}{\texttt{irony\_identification}} & \href{https://github.com/google/BIG-bench/tree/main/bigbench/benchmark_tasks/kannada}{\texttt{kannada}} & \href{https://github.com/google/BIG-bench/tree/main/bigbench/benchmark_tasks/key\_value\_maps}{\texttt{key\_value\_maps}} \\
\href{https://github.com/google/BIG-bench/tree/main/bigbench/benchmark_tasks/known\_unknowns}{\texttt{known\_unknowns}} & \href{https://github.com/google/BIG-bench/tree/main/bigbench/benchmark_tasks/logical\_args}{\texttt{logical\_args}} & \href{https://github.com/google/BIG-bench/tree/main/bigbench/benchmark_tasks/logical\_sequence}{\texttt{logical\_sequence}} \\
\href{https://github.com/google/BIG-bench/tree/main/bigbench/benchmark_tasks/mathematical\_induction}{\texttt{mathematical\_induction}} & \href{https://github.com/google/BIG-bench/tree/main/bigbench/benchmark_tasks/metaphor\_boolean}{\texttt{metaphor\_boolean}} & \href{https://github.com/google/BIG-bench/tree/main/bigbench/benchmark_tasks/metaphor\_understanding}{\texttt{metaphor\_understanding}} \\
\href{https://github.com/google/BIG-bench/tree/main/bigbench/benchmark_tasks/misconceptions}{\texttt{misconceptions}} & \href{https://github.com/google/BIG-bench/tree/main/bigbench/benchmark_tasks/misconceptions\_russian}{\texttt{misconceptions\_russian}} & \href{https://github.com/google/BIG-bench/tree/main/bigbench/benchmark_tasks/moral\_permissibility}{\texttt{moral\_permissibility}} \\
\href{https://github.com/google/BIG-bench/tree/main/bigbench/benchmark_tasks/movie\_recommendation}{\texttt{movie\_recommendation}} & \href{https://github.com/google/BIG-bench/tree/main/bigbench/benchmark_tasks/nonsense\_words\_grammar}{\texttt{nonsense\_words\_grammar}} & \href{https://github.com/google/BIG-bench/tree/main/bigbench/benchmark_tasks/odd\_one\_out}{\texttt{odd\_one\_out}} \\
\href{https://github.com/google/BIG-bench/tree/main/bigbench/benchmark_tasks/penguins\_in\_a\_table}{\texttt{penguins\_in\_a\_table}} & \href{https://github.com/google/BIG-bench/tree/main/bigbench/benchmark_tasks/periodic\_elements}{\texttt{periodic\_elements}} & \href{https://github.com/google/BIG-bench/tree/main/bigbench/benchmark_tasks/persian\_idioms}{\texttt{persian\_idioms}} \\
\href{https://github.com/google/BIG-bench/tree/main/bigbench/benchmark_tasks/phrase\_relatedness}{\texttt{phrase\_relatedness}} & \href{https://github.com/google/BIG-bench/tree/main/bigbench/benchmark_tasks/physical\_intuition}{\texttt{physical\_intuition}} & \href{https://github.com/google/BIG-bench/tree/main/bigbench/benchmark_tasks/physics}{\texttt{physics}} \\
\href{https://github.com/google/BIG-bench/tree/main/bigbench/benchmark_tasks/presuppositions\_as\_nli}{\texttt{presuppositions\_as\_nli}} & \href{https://github.com/google/BIG-bench/tree/main/bigbench/benchmark_tasks/riddle\_sense}{\texttt{riddle\_sense}} & \href{https://github.com/google/BIG-bench/tree/main/bigbench/benchmark_tasks/ruin\_names}{\texttt{ruin\_names}} \\
\href{https://github.com/google/BIG-bench/tree/main/bigbench/benchmark_tasks/salient\_translation\_error\_detection}{\texttt{salient\_translation\_error\_detection}} & \href{https://github.com/google/BIG-bench/tree/main/bigbench/benchmark_tasks/sentence\_ambiguity}{\texttt{sentence\_ambiguity}} & \href{https://github.com/google/BIG-bench/tree/main/bigbench/benchmark_tasks/similarities\_abstraction}{\texttt{similarities\_abstraction}} \\
\href{https://github.com/google/BIG-bench/tree/main/bigbench/benchmark_tasks/simple\_arithmetic\_json\_multiple\_choice}{\texttt{simple\_arithmetic\_json\_multiple\_choice}} & \href{https://github.com/google/BIG-bench/tree/main/bigbench/benchmark_tasks/simple\_ethical\_questions}{\texttt{simple\_ethical\_questions}} & \href{https://github.com/google/BIG-bench/tree/main/bigbench/benchmark_tasks/snarks}{\texttt{snarks}} \\
\href{https://github.com/google/BIG-bench/tree/main/bigbench/benchmark_tasks/social\_support}{\texttt{social\_support}} & \href{https://github.com/google/BIG-bench/tree/main/bigbench/benchmark_tasks/sports\_understanding}{\texttt{sports\_understanding}} & \href{https://github.com/google/BIG-bench/tree/main/bigbench/benchmark_tasks/strange\_stories}{\texttt{strange\_stories}} \\
\href{https://github.com/google/BIG-bench/tree/main/bigbench/benchmark_tasks/suicide\_risk}{\texttt{suicide\_risk}} & \href{https://github.com/google/BIG-bench/tree/main/bigbench/benchmark_tasks/swahili\_english\_proverbs}{\texttt{swahili\_english\_proverbs}} & \href{https://github.com/google/BIG-bench/tree/main/bigbench/benchmark_tasks/symbol\_interpretation}{\texttt{symbol\_interpretation}} \\
\href{https://github.com/google/BIG-bench/tree/main/bigbench/benchmark_tasks/understanding\_fables}{\texttt{understanding\_fables}} & \href{https://github.com/google/BIG-bench/tree/main/bigbench/benchmark_tasks/undo\_permutation}{\texttt{undo\_permutation}} & \href{https://github.com/google/BIG-bench/tree/main/bigbench/benchmark_tasks/unit\_interpretation}{\texttt{unit\_interpretation}} \\
\href{https://github.com/google/BIG-bench/tree/main/bigbench/benchmark_tasks/what\_is\_the\_tao}{\texttt{what\_is\_the\_tao}} & \href{https://github.com/google/BIG-bench/tree/main/bigbench/benchmark_tasks/which\_wiki\_edit}{\texttt{which\_wiki\_edit}} \\

         \bottomrule   
    \end{tabular}}
    \caption{The list of multiple-choice tasks we use from BIG-Bench. Clicking the name of a task will direct you to the task's GitHub page.}
    \label{tab:bigbench_tasks}
\end{table*}

\subsection{Prompts}
\label{app:prompts}
We use fixed prompt formats from the \bigbench{} datasets. Optimizing the prompts might lead to extra margins in performance. Studying the relationship between prompt formats and downstream task performance along the trajectory is interesting, but we consider it out of the scope of this work. We present examples from four datasets in \autoref{tab:bigbench_examples}. 
\begin{table*}[t]
\centering
\begin{tabular}{p{\textwidth}}
\toprule
\textbf{date\_understanding}     \\
\vspace{0.05em}
    Q: Yesterday, Jan 21, 2011, Jane ate 2 pizzas and 5 wings. What is the date tomorrow in MM/DD/YYYY? \\
    A: 01/23/2011\\\vspace{0.05em}
    Q: It is 4/19/1969 today. What is the date yesterday in MM/DD/YYYY?\\A: 04/18/1969\\\vspace{0.05em}
    Q: Yesterday was April 30, 2021. What is the date today in MM/DD/YYYY?\\A: \\ \vspace{0.05em}
    Options: \texttt{05/01/2021,02/23/2021,03/11/2021,05/09/2021,06/12/2021}  \\ \midrule
    \textbf{nonsense\_words\_grammar} \\
    \vspace{0.05em}
    Q: How many things does the following sentence describe? The balforator, heddleilwilder and the sminniging crolostat operate superbly and without interrtulation.\\A: 3\\\vspace{0.05em}Q: How is the quijerinnedescribed in the next sentence? The umulophanitc quijerinne eriofrols the dusty grass.\\A: umulophanitc\\\vspace{0.05em}Q: Which word in the following sentence is a verb? The grilshaws bolheavened whincely.\\A:                                     \\
    \vspace{0.05em}
    Options: \texttt{The, grilshaws, bolheavened, whincely} \\ \midrule                                                         
    \textbf{entailed\_polarity}      \\ \vspace{0.05em} Given a fact, answer the following question with a yes or a no.\\Fact: Ed grew to like Mary. Q: Did Ed like Mary?\\A: yes\\\vspace{0.05em}Given a fact, answer the following question with a yes or a no.\\Fact: They did not condescend to go. Q: Did they go?\\A: no\\\vspace{0.05em}Given a fact, answer the following question with a yes or a no.\\Fact: The report was admitted to be incorrect. Q: Was the report incorrect?\\A: \\ \vspace{0.05em} Options: \texttt{yes, no}   \\ \midrule
    \textbf{sentence\_ambiguity} \\ \vspace{0.05em}
    Claim: Delhi is not the only Hindi-speakingstate in India.\\True or False? True\\\vspace{0.05em}Claim: The population of the second-largest country in the world in 2021 exceeds the population of the third, fourth, and fifth largest countries combined.\\True or False? True\\\vspace{0.05em}Claim: Pescatarians almost never consume vegetarian food.\\True or False?\\
    \vspace{0.05em} \\
    Options: \texttt{True, False} \\
    \bottomrule
\end{tabular}
\caption{Examples of prompts and answer options for four \bigbench{} multiple-choice tasks.}
\label{tab:bigbench_examples}
\end{table*}

\subsection{Linearity and Breakthroughness Tasks}
\label{app:linear}
\begin{table*}
\centering
\resizebox{\linewidth}{!}{
\begin{tabular}{lll} \toprule
        \multicolumn{3}{c}{\textbf{Linearity Tasks}} \\
        \href{https://github.com/google/BIG-bench/tree/main/bigbench/benchmark_tasks/date\_understanding}{\texttt{date\_understanding}} & \href{https://github.com/google/BIG-bench/tree/main/bigbench/benchmark_tasks/fantasy\_reasoning}{\texttt{fantasy\_reasoning}} & \href{https://github.com/google/BIG-bench/tree/main/bigbench/benchmark_tasks/figure\_of\_speech\_detection}{\texttt{figure\_of\_speech\_detection}} \\
\href{https://github.com/google/BIG-bench/tree/main/bigbench/benchmark_tasks/hhh\_alignment}{\texttt{hhh\_alignment}} & \href{https://github.com/google/BIG-bench/tree/main/bigbench/benchmark_tasks/implicit\_relations}{\texttt{implicit\_relations}} & \href{https://github.com/google/BIG-bench/tree/main/bigbench/benchmark_tasks/intent\_recognition}{\texttt{intent\_recognition}} \\
\href{https://github.com/google/BIG-bench/tree/main/bigbench/benchmark_tasks/misconceptions}{\texttt{misconceptions}} & \href{https://github.com/google/BIG-bench/tree/main/bigbench/benchmark_tasks/similarities\_abstraction}{\texttt{similarities\_abstraction}} & \href{https://github.com/google/BIG-bench/tree/main/bigbench/benchmark_tasks/simple\_ethical\_questions}{\texttt{simple\_ethical\_questions}} \\
\href{https://github.com/google/BIG-bench/tree/main/bigbench/benchmark_tasks/strange\_stories}{\texttt{strange\_stories}} & \href{https://github.com/google/BIG-bench/tree/main/bigbench/benchmark_tasks/undo\_permutation}{\texttt{undo\_permutation}} & \href{https://github.com/google/BIG-bench/tree/main/bigbench/benchmark_tasks/nonsense\_words\_grammar}{\texttt{nonsense\_words\_grammar}} \\ \midrule
\multicolumn{3}{c}{\textbf{Breakthroughness Tasks}} \\
\href{https://github.com/google/BIG-bench/tree/main/bigbench/benchmark_tasks/code\_line\_description}{\texttt{code\_line\_description}} & \href{https://github.com/google/BIG-bench/tree/main/bigbench/benchmark_tasks/human\_organs\_senses}{\texttt{human\_organs\_senses}} &
\href{https://github.com/google/BIG-bench/tree/main/bigbench/benchmark_tasks/phrase\_relatedness}{\texttt{phrase\_relatedness}} \\   \href{https://github.com/google/BIG-bench/tree/main/bigbench/benchmark_tasks/swahili\_english\_proverbs}{\texttt{swahili\_english\_proverbs}} & \href{https://github.com/google/BIG-bench/tree/main/bigbench/benchmark_tasks/what\_is\_the\_tao}{\texttt{what\_is\_the\_tao}} & \href{https://github.com/google/BIG-bench/tree/main/bigbench/benchmark_tasks/implicatures}{\texttt{implicatures}} \\ \bottomrule   
\end{tabular}}
\caption{The list of linearity and breakthroughness tasks.}
\label{tab:bigbench_LB_tasks}
\end{table*}

\citet{srivastava2022beyond} identify tasks showing a linearity or breakthroughness pattern and \citep{wei2022emergent} coin the term \textit{emergent ability} for models showing breakthroughness patterns on certain tasks. Previous works mainly study scaling patterns of downstream tasks with final model checkpoints, and we extend this to training trajectories of models across scales. We largely follow \citet{srivastava2022beyond} to identify tasks with linearity and breakthroughness patterns -- the former depicts the trend where the task performance scales with the model size reliably, and for the latter, the performance remains low until a critical model size. 

We select 12 tasks that show a linearity pattern and 6 tasks that show a breakthroughness pattern based on the metrics proposed in \citep{srivastava2022beyond}. For each model size $x_i$ and the corresponding performance $y_i$, the metrics are defined as
\begin{align}
    L = \frac{I(y)}{\sqrt{\frac{1}{n} \sum_i z_i^2}}; B = \frac{I(y)}{\sqrt{\text{Median}(\{z_i^2\})}};
\end{align}
where $I(y) = \text{sign}(\arg\max_i y_i - \arg\max_i y_i)$ \\ $\cdot (\max_i y_i - \min_i y_i)$ is a measure to capture the overall improvement of performance when scaling up. We find that these two measures are not sufficient for identifying the scaling trends for linearity and breakthroughness, thus we also manually check the scaling pattern to verify. The linearity and breakthroughness tasks are lists in \autoref{tab:bigbench_LB_tasks}.

\subsection{Trajectory of Each Task}
\label{app:single_task}
\begin{figure*}
    \centering
    \includegraphics[width=0.7\linewidth]{images/legend-allmodels-dash.pdf} \\
    \includegraphics[width=0.2\linewidth]{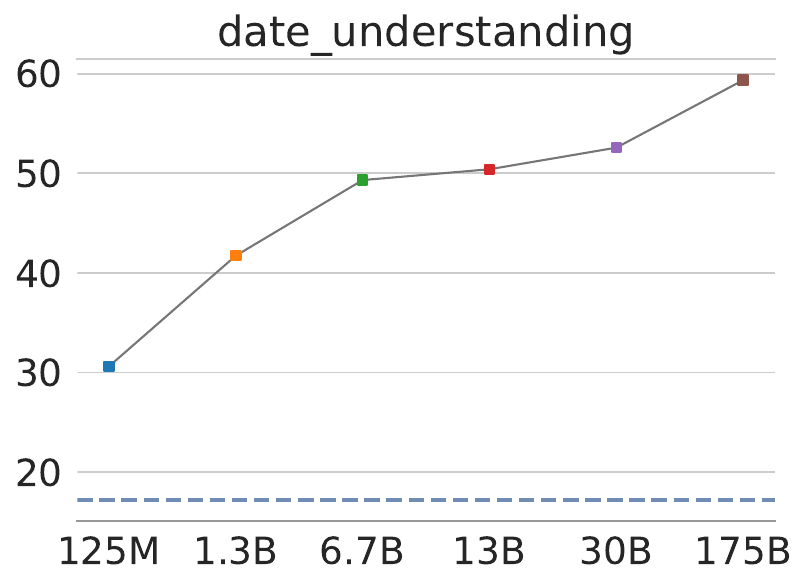}
    \includegraphics[width=0.2\linewidth]{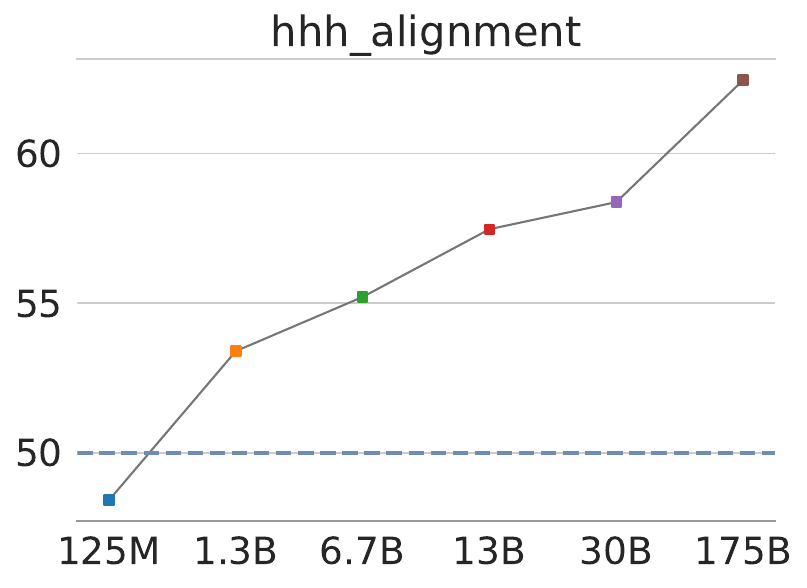}
    \includegraphics[width=0.2\linewidth]{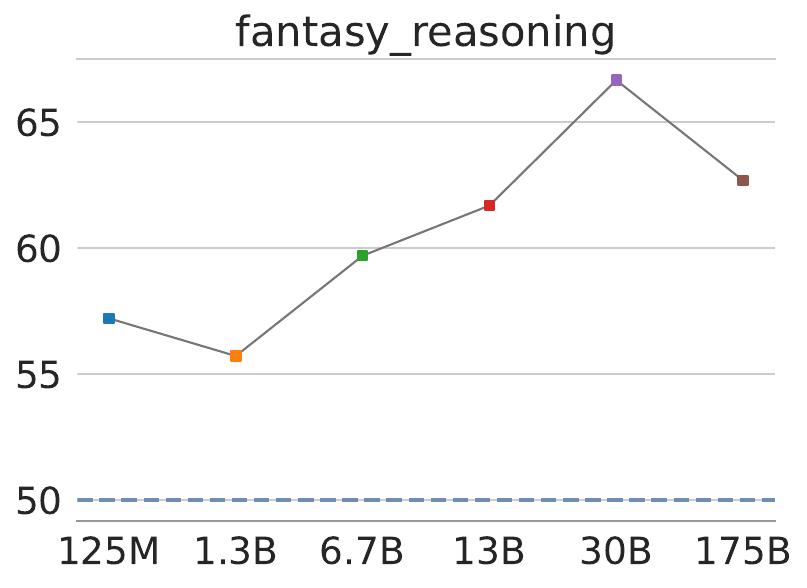}
    \includegraphics[width=0.2\linewidth]{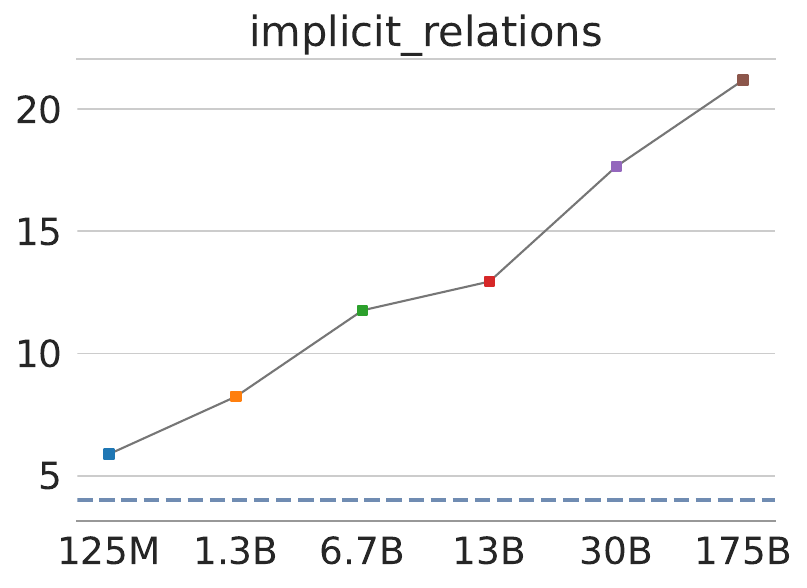}
    \includegraphics[width=0.2\linewidth]{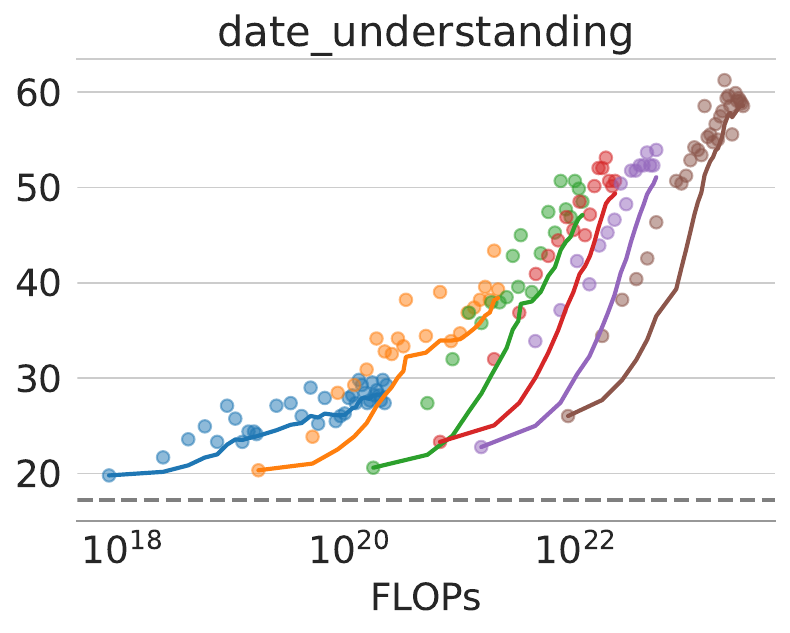}
    \includegraphics[width=0.2\linewidth]{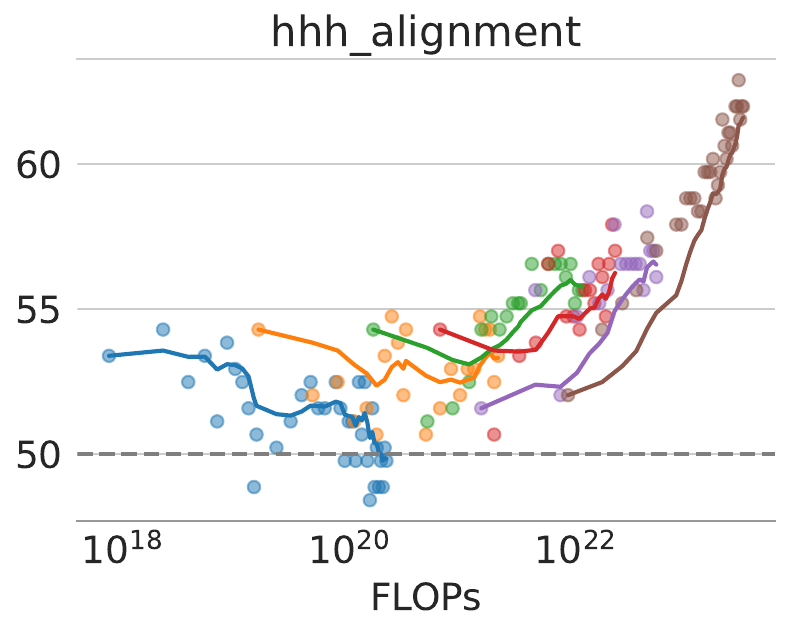}
    \includegraphics[width=0.2\linewidth]{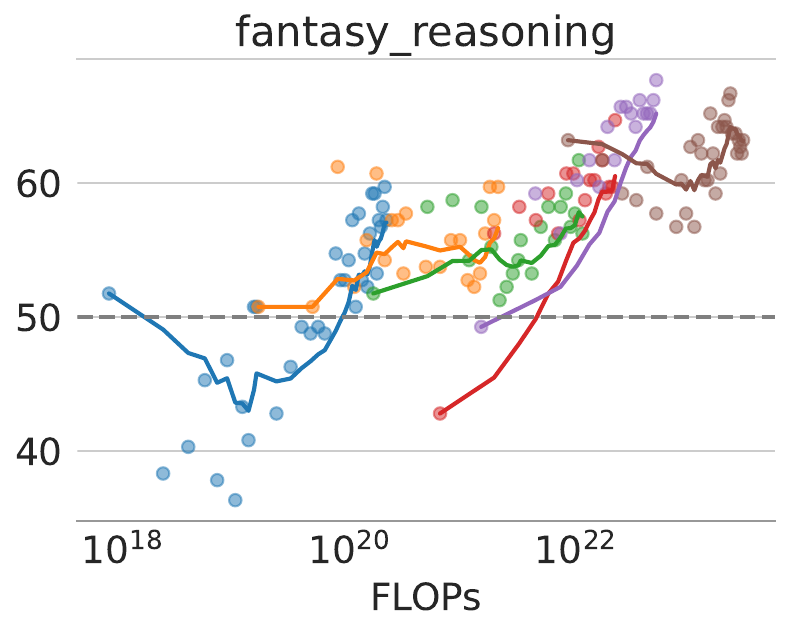}
    \includegraphics[width=0.2\linewidth]{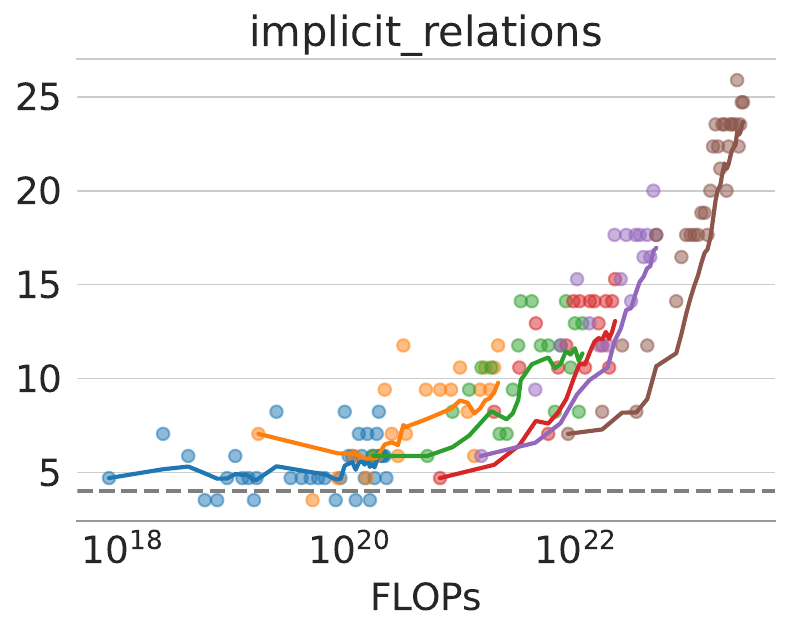} \\
    \includegraphics[width=0.2\linewidth]{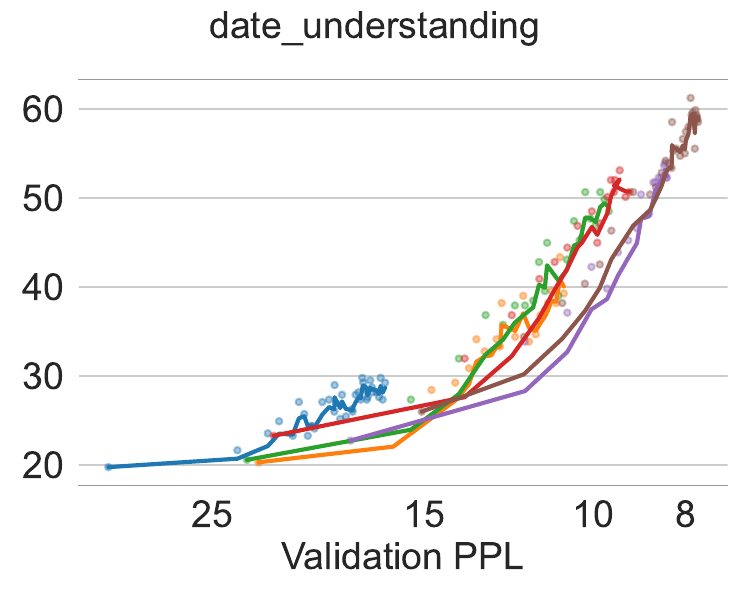}
    \includegraphics[width=0.2\linewidth]{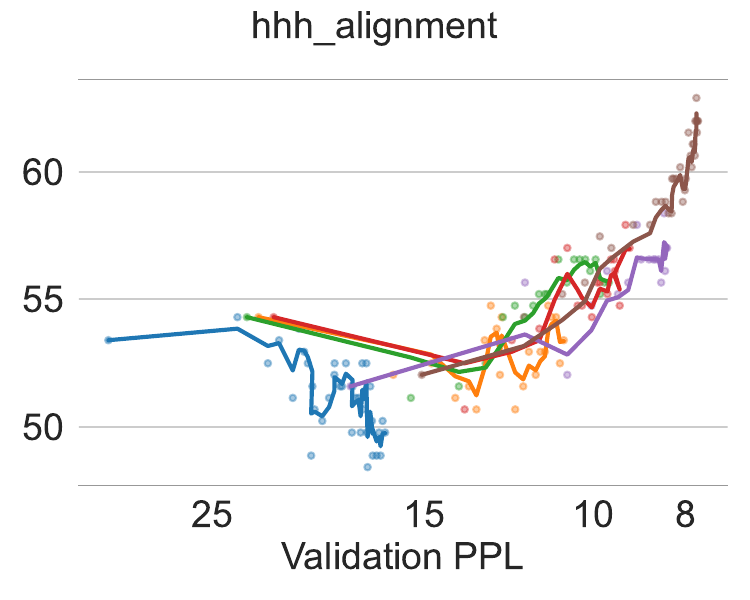}
    \includegraphics[width=0.2\linewidth]{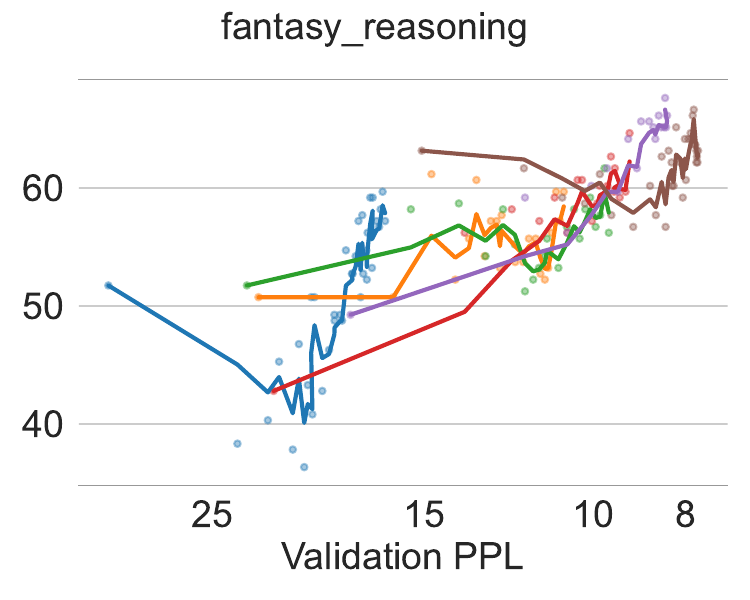}
    \includegraphics[width=0.2\linewidth]{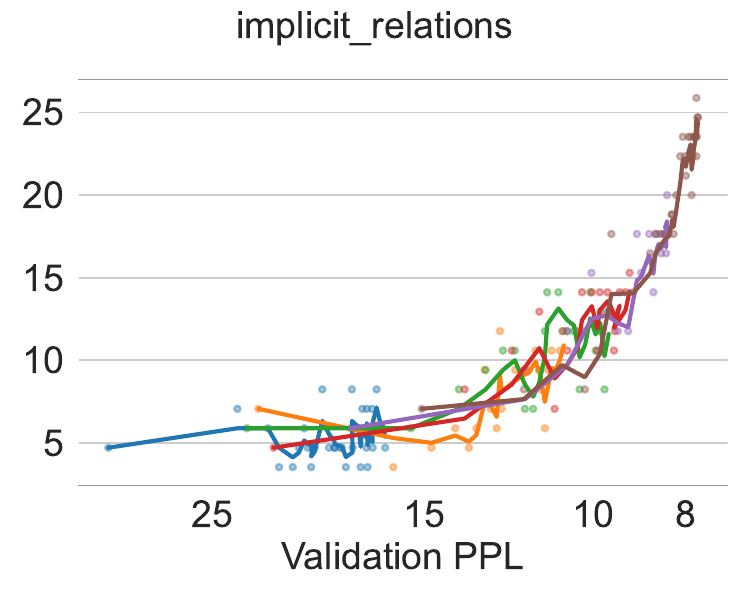} \\
    \includegraphics[width=0.2\linewidth]{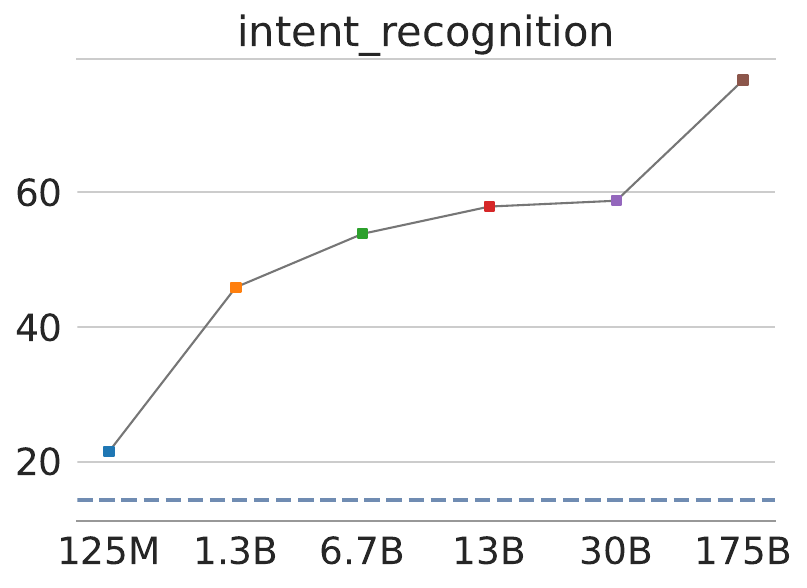}
    \includegraphics[width=0.2\linewidth]{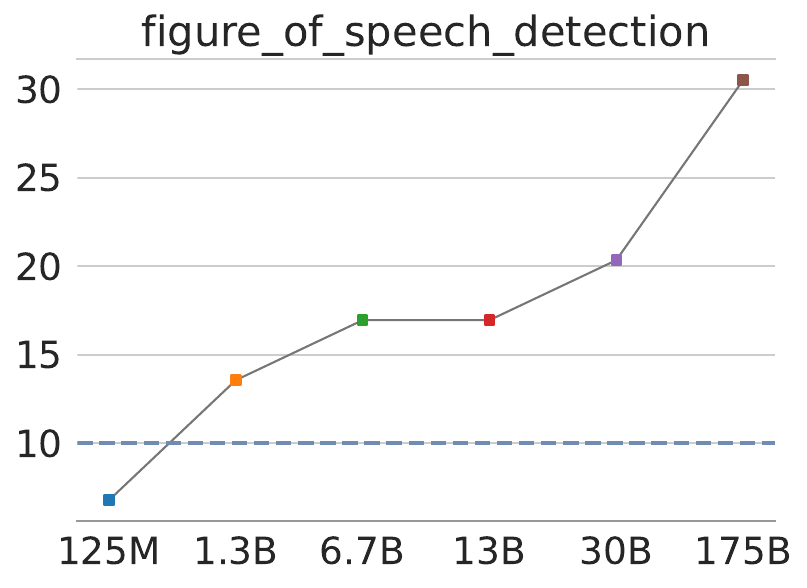}
    \includegraphics[width=0.2\linewidth]{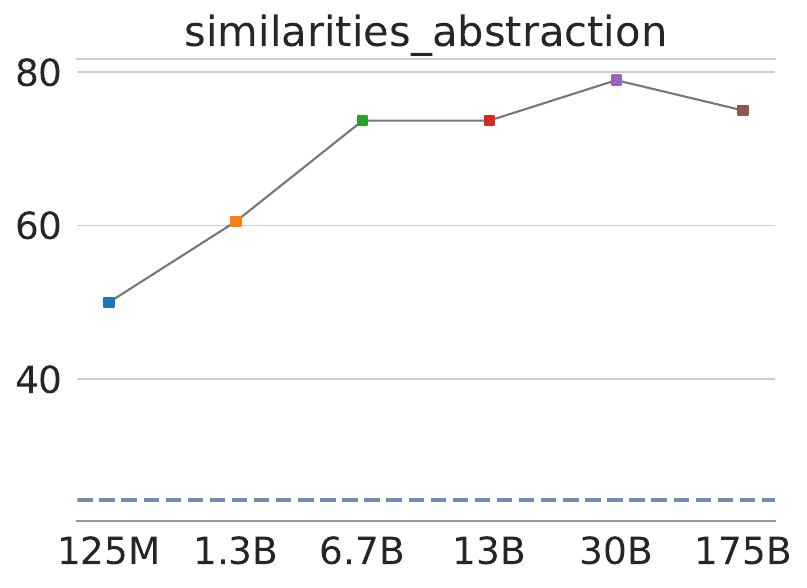}
    \includegraphics[width=0.2\linewidth]{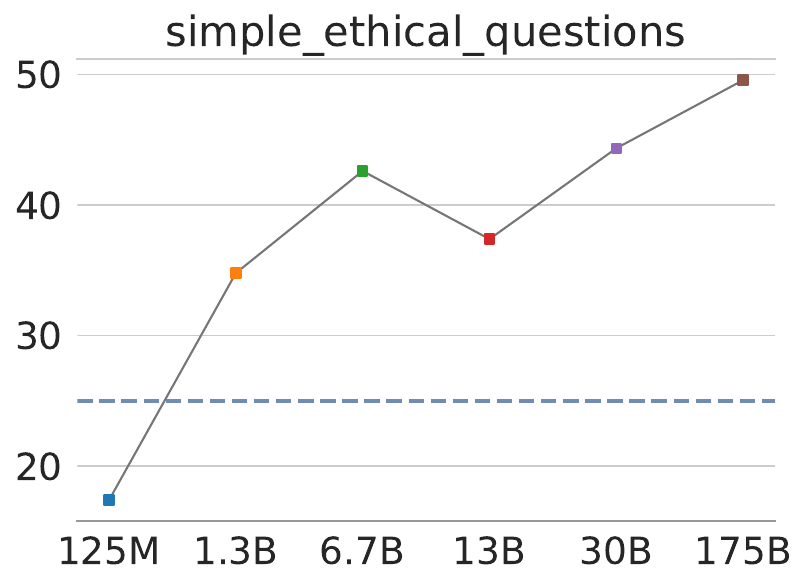}  \\
     
    \includegraphics[width=0.2\linewidth]{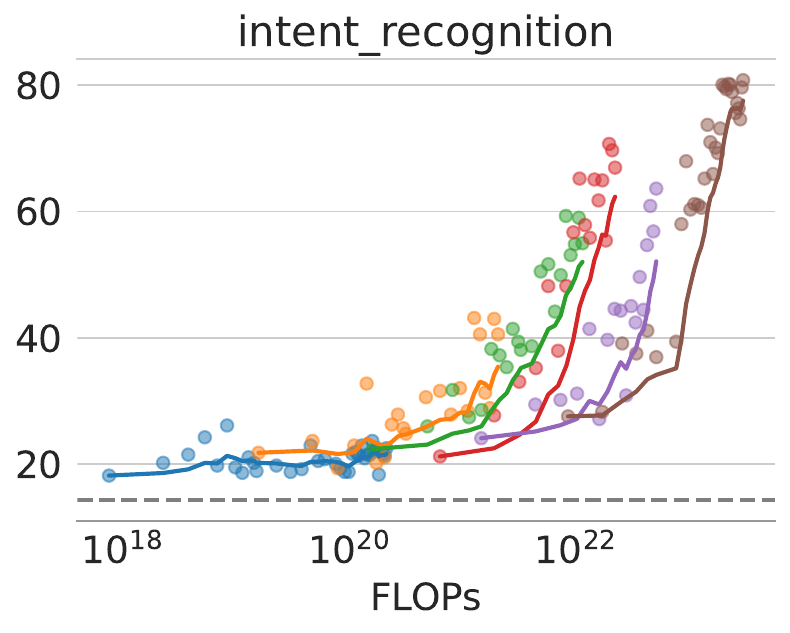}
    \includegraphics[width=0.2\linewidth]{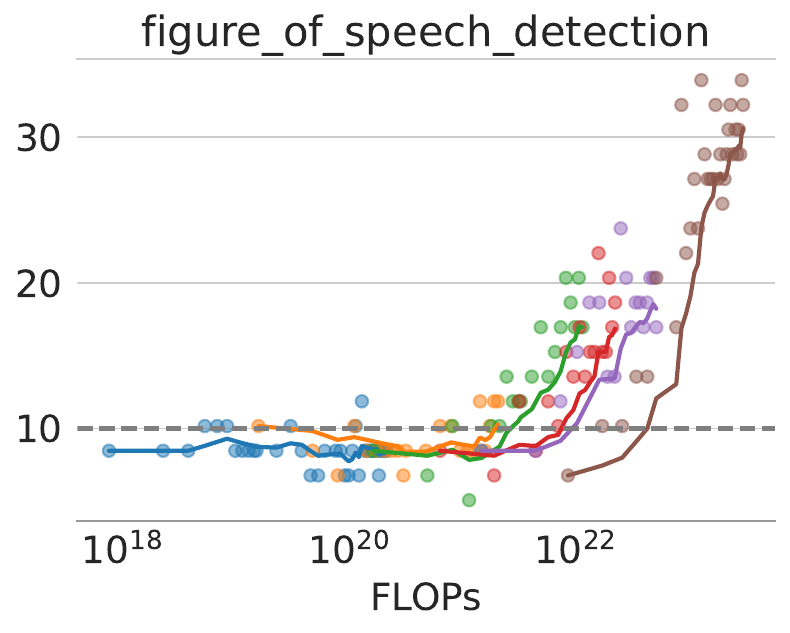} 
    \includegraphics[width=0.2\linewidth]{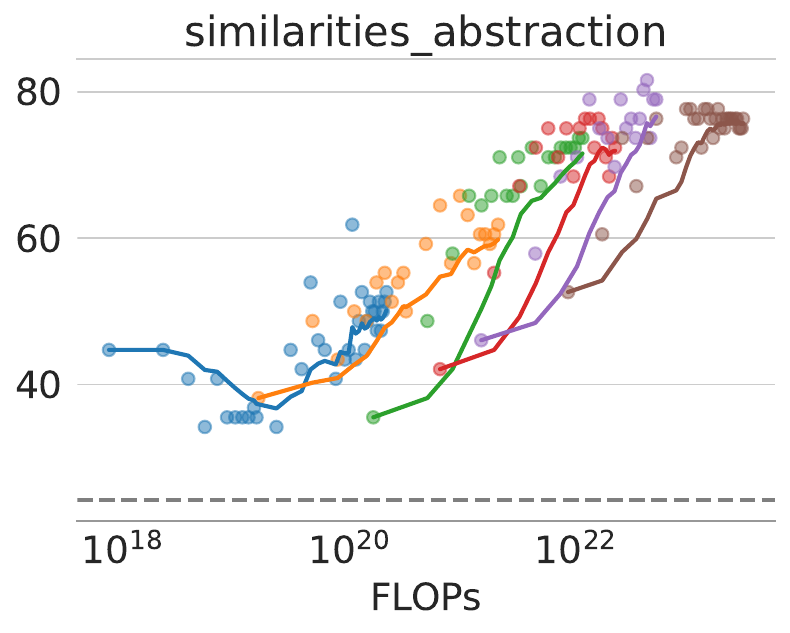}
    \includegraphics[width=0.2\linewidth]{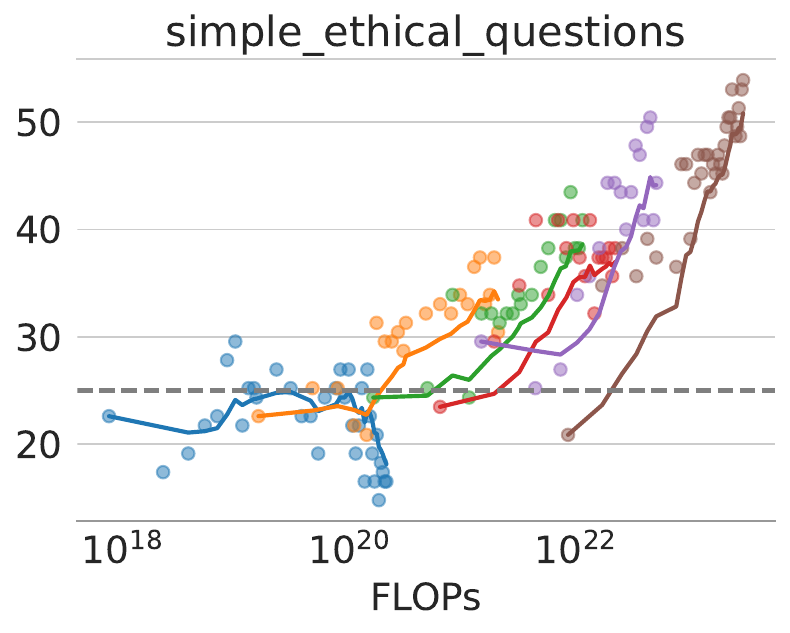} \\
    \includegraphics[width=0.2\linewidth]{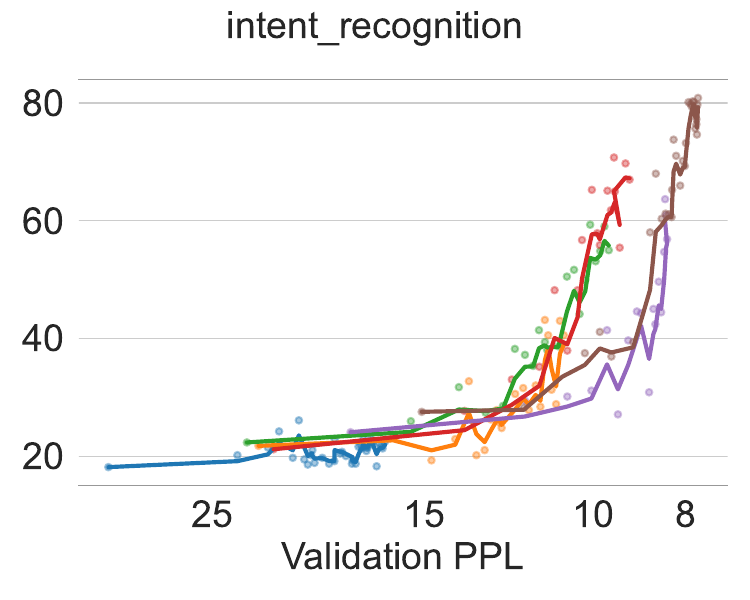}
    \includegraphics[width=0.2\linewidth]{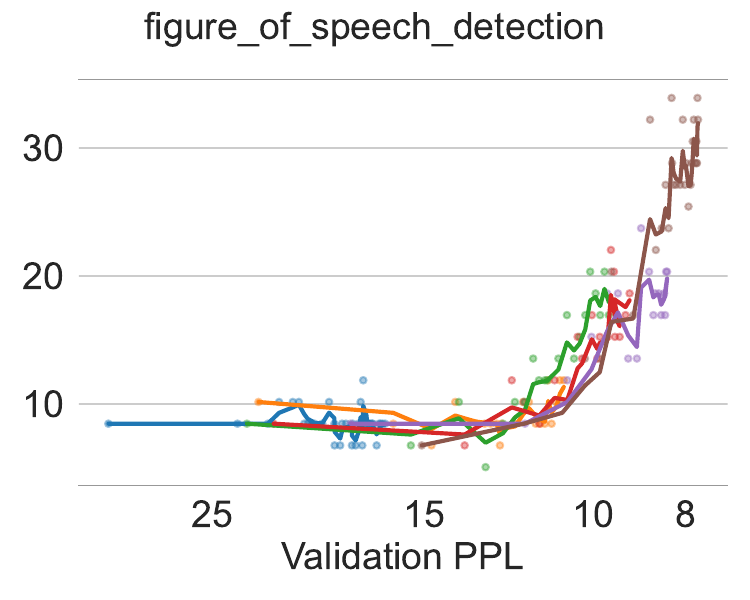}
    \includegraphics[width=0.2\linewidth]{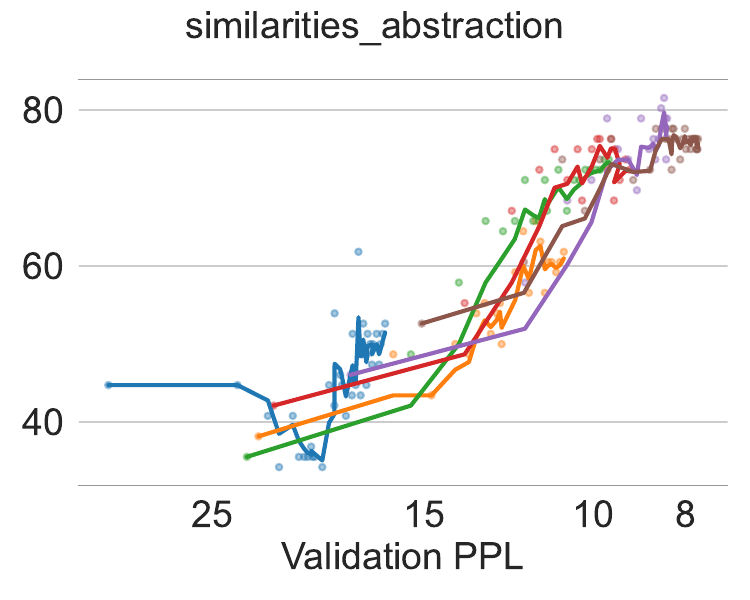}
    \includegraphics[width=0.2\linewidth]{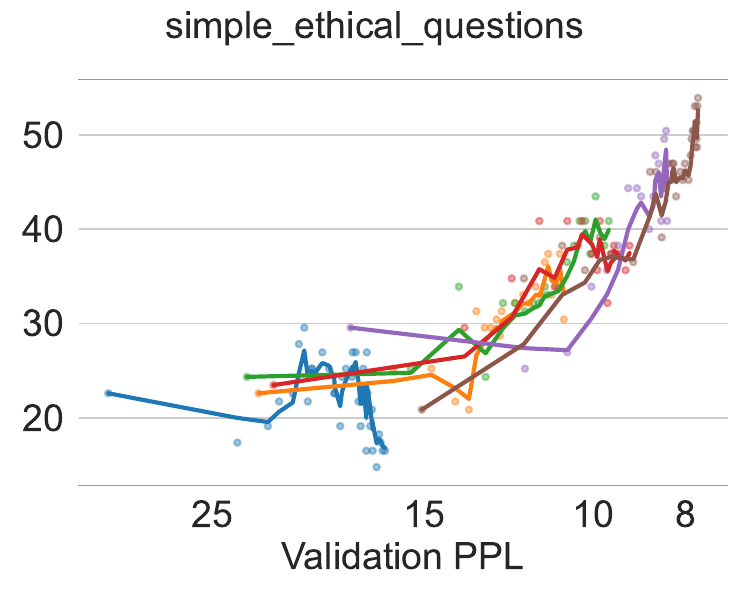}  \\
    \includegraphics[width=0.2\linewidth]{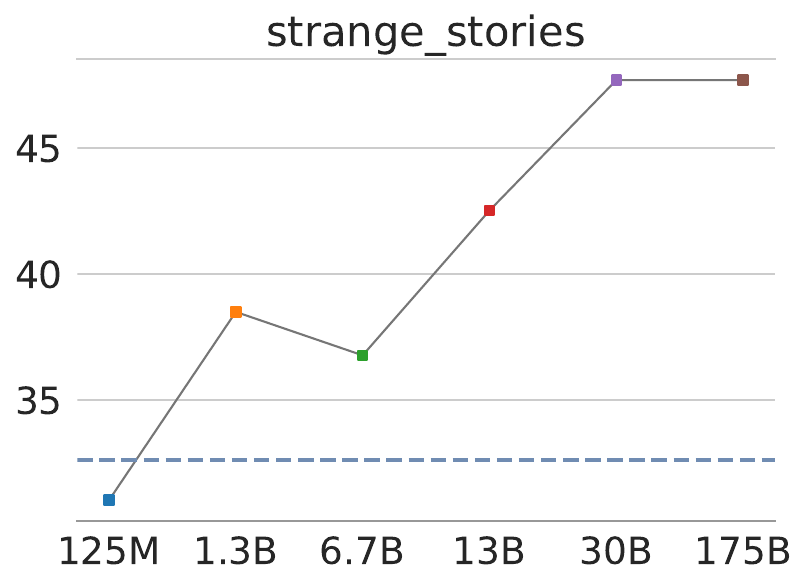}
    \includegraphics[width=0.2\linewidth]{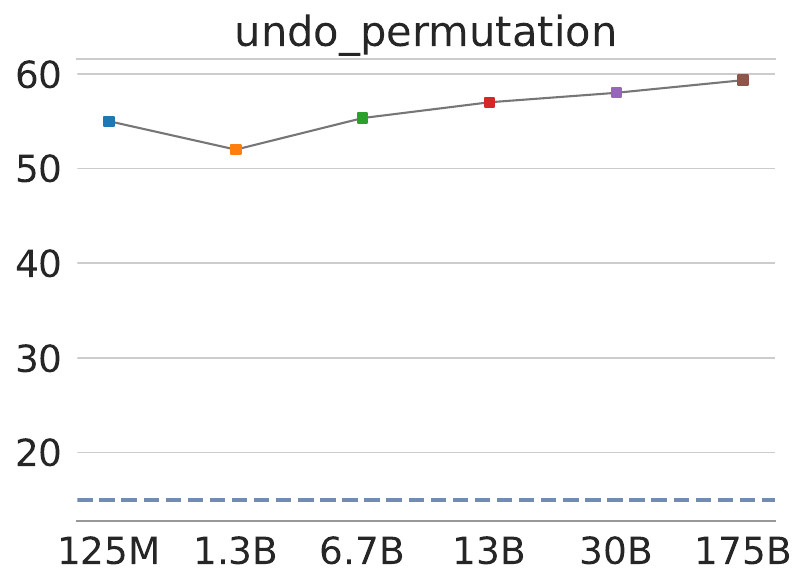}
    \includegraphics[width=0.2\linewidth]{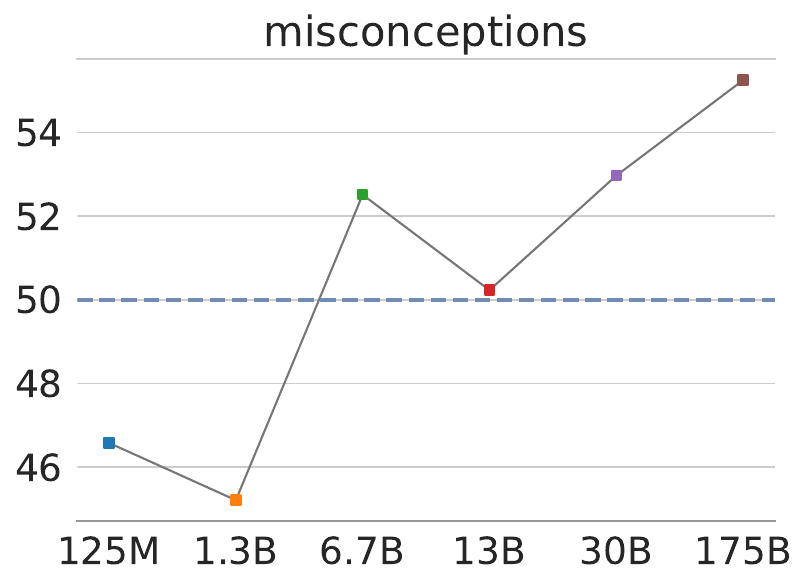}
    \includegraphics[width=0.2\linewidth]{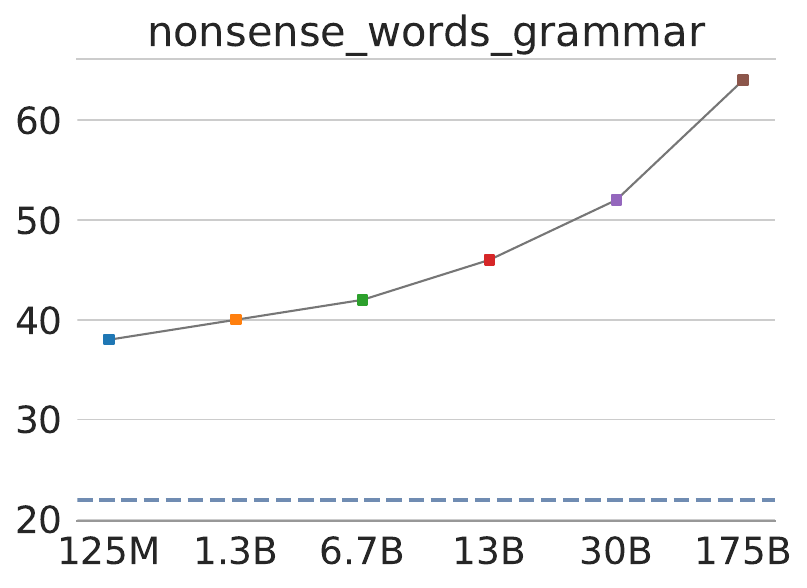}  \\
    \includegraphics[width=0.2\linewidth]{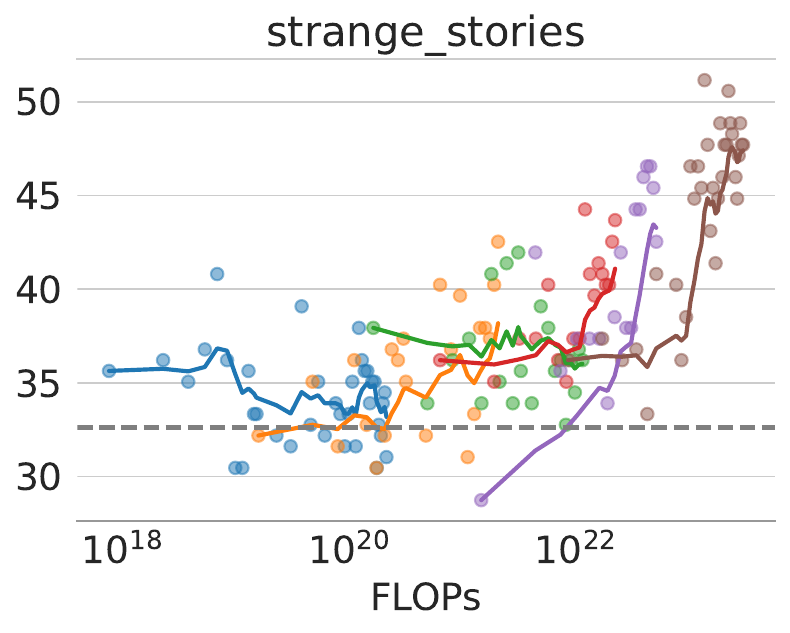}
    \includegraphics[width=0.2\linewidth]{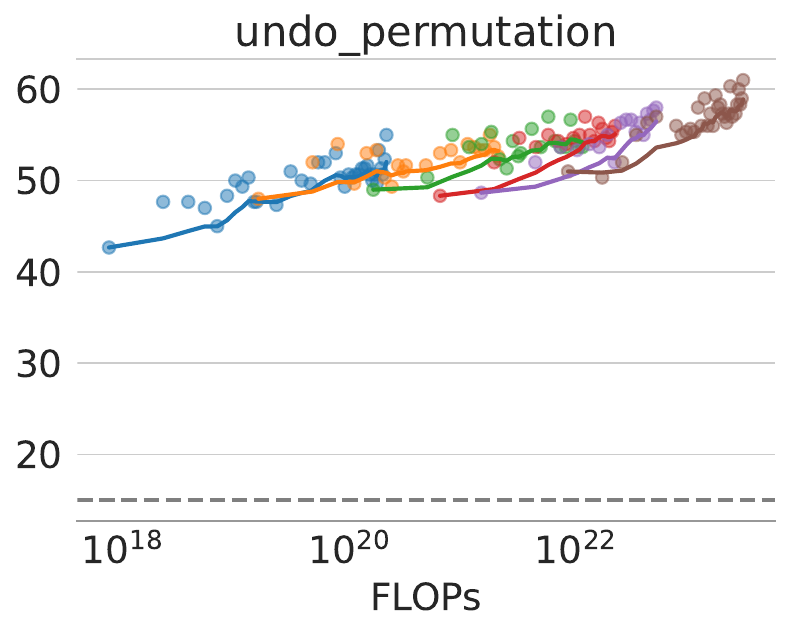} 
    \includegraphics[width=0.2\linewidth]{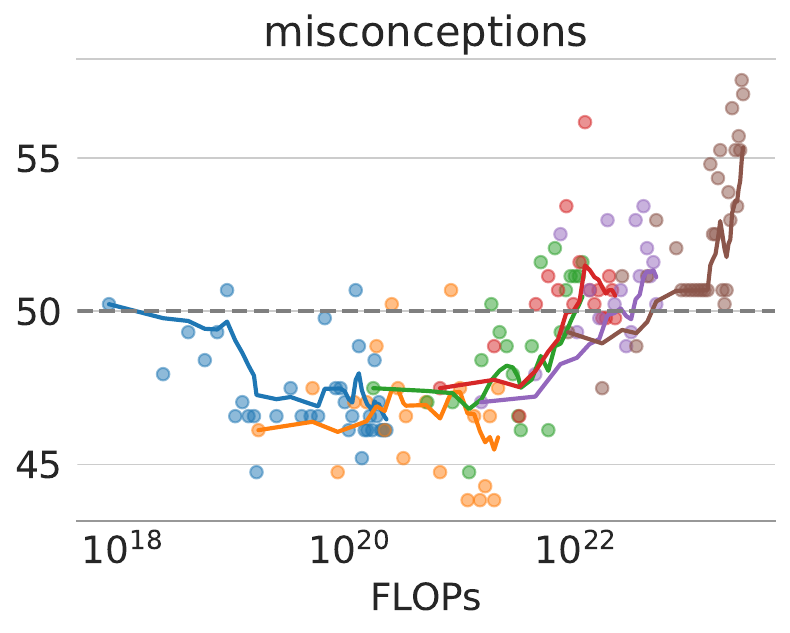}
    \includegraphics[width=0.2\linewidth]{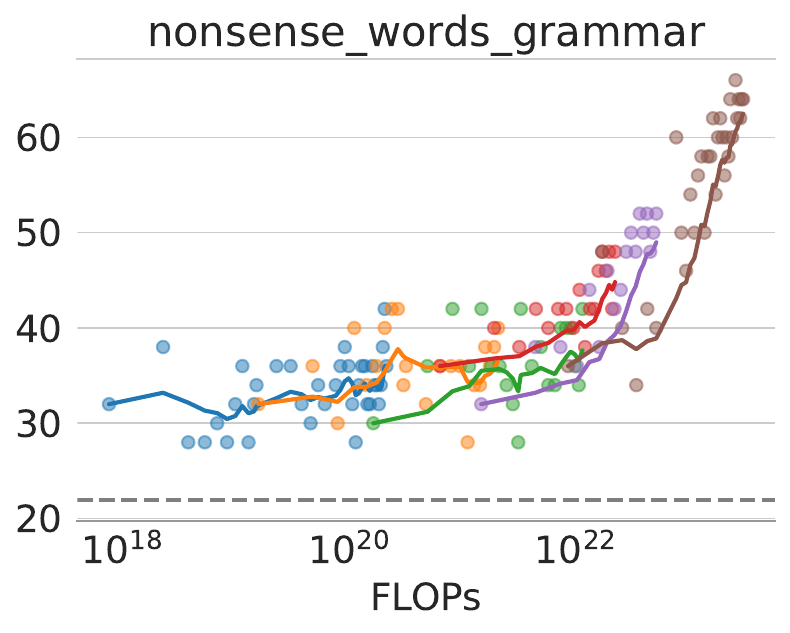} \\
    \includegraphics[width=0.2\linewidth]{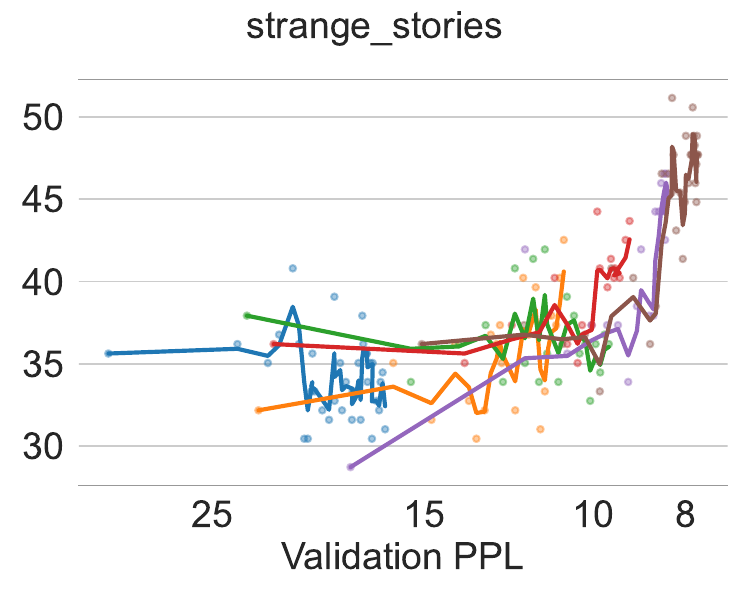}
    \includegraphics[width=0.2\linewidth]{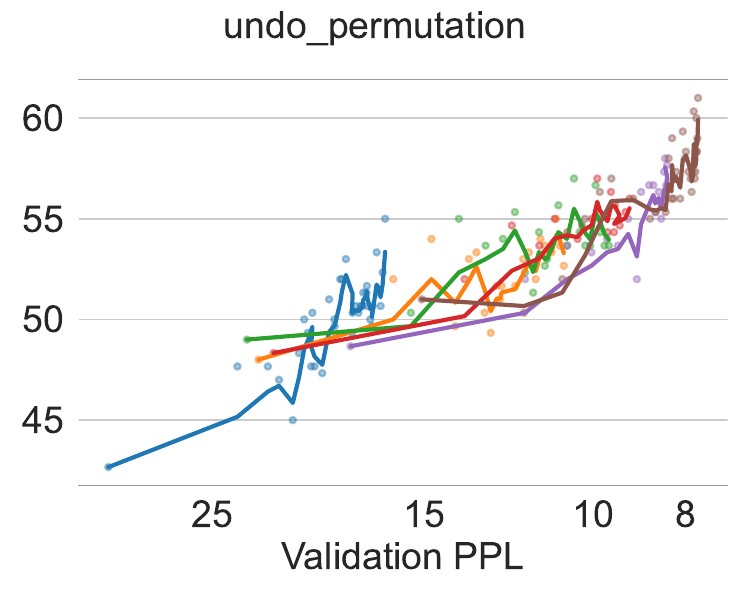}
    \includegraphics[width=0.2\linewidth]{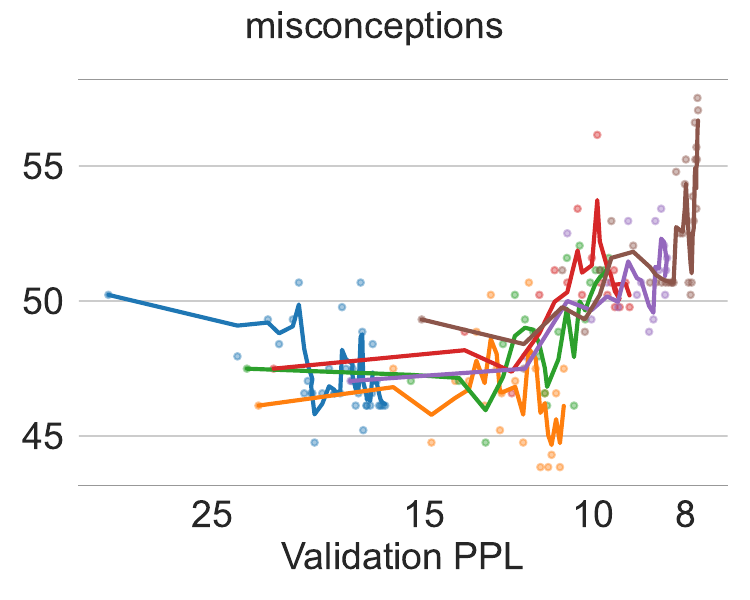}
    \includegraphics[width=0.2\linewidth]{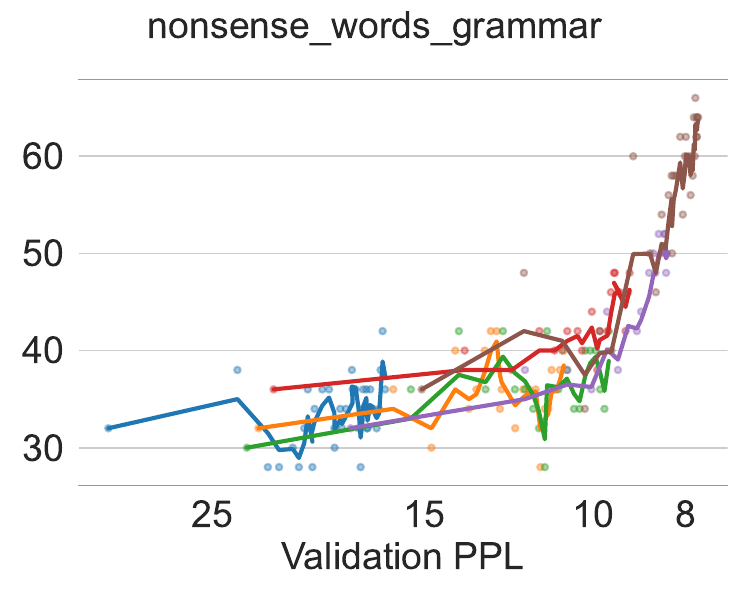}  \\
    \caption{Scaling curves and trajectories of linearity tasks. }
    \label{fig:single_linearity}
\end{figure*}

\begin{figure*}
    \centering
    \includegraphics[width=0.7\linewidth]{images/legend-allmodels-dash.pdf} \\
    \includegraphics[width=0.2\linewidth]{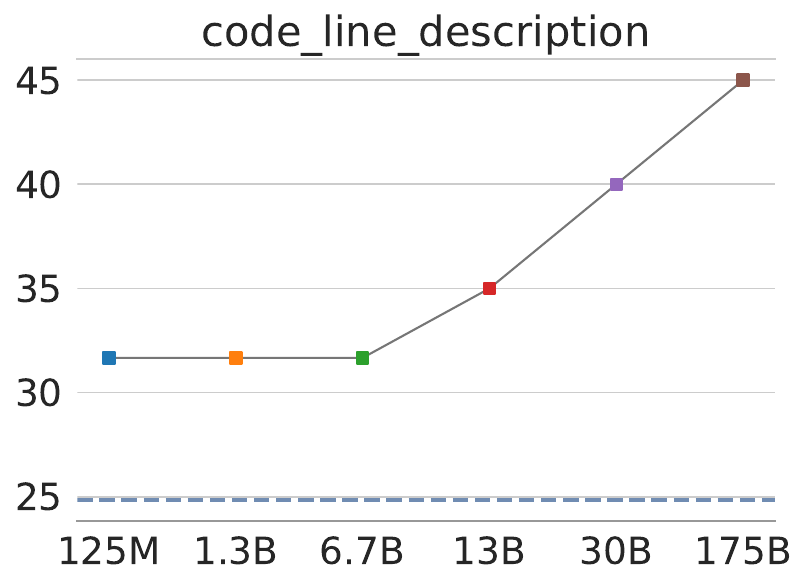}
    \includegraphics[width=0.2\linewidth]{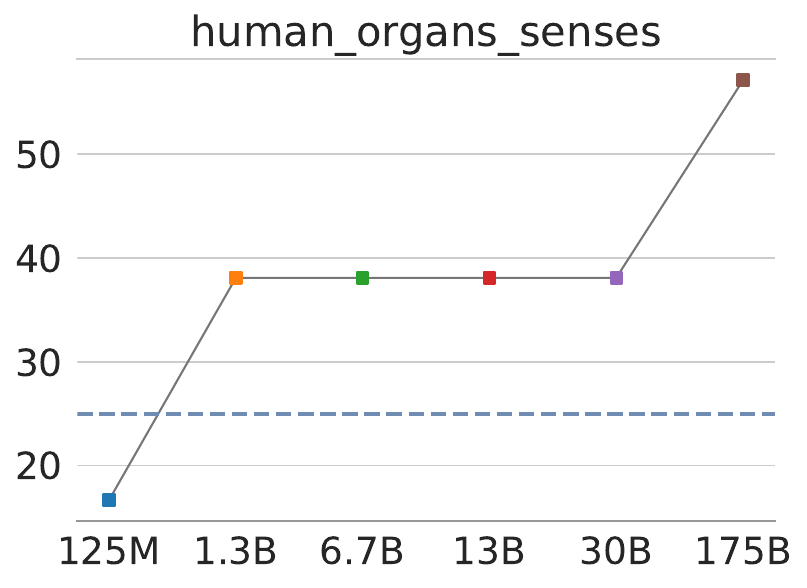}
    \includegraphics[width=0.2\linewidth]{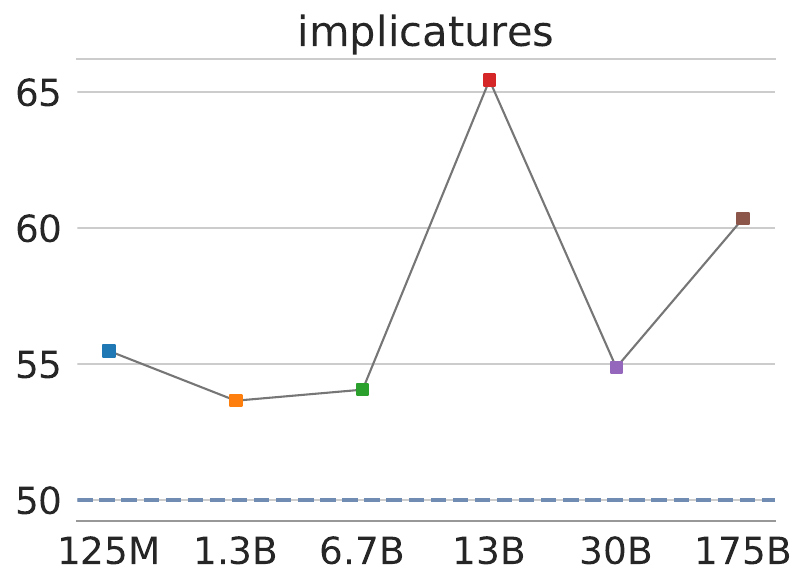}
    \includegraphics[width=0.2\linewidth]{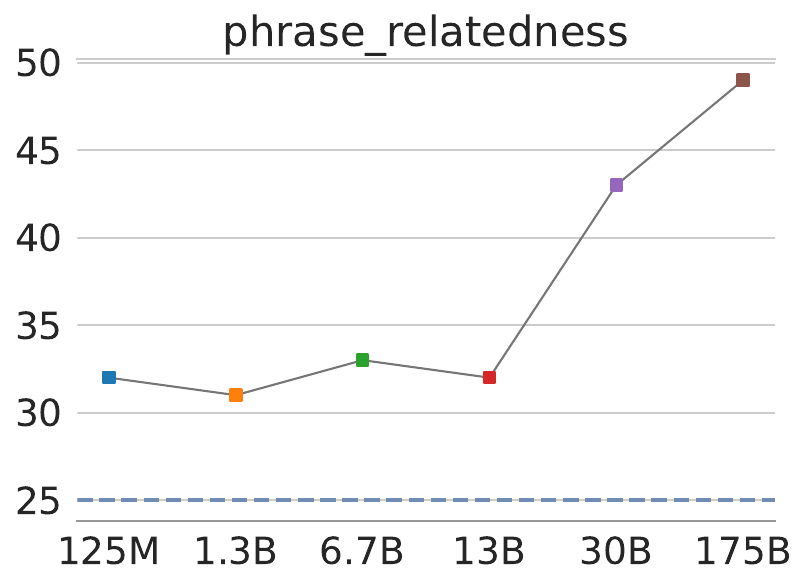}
    \includegraphics[width=0.2\linewidth]{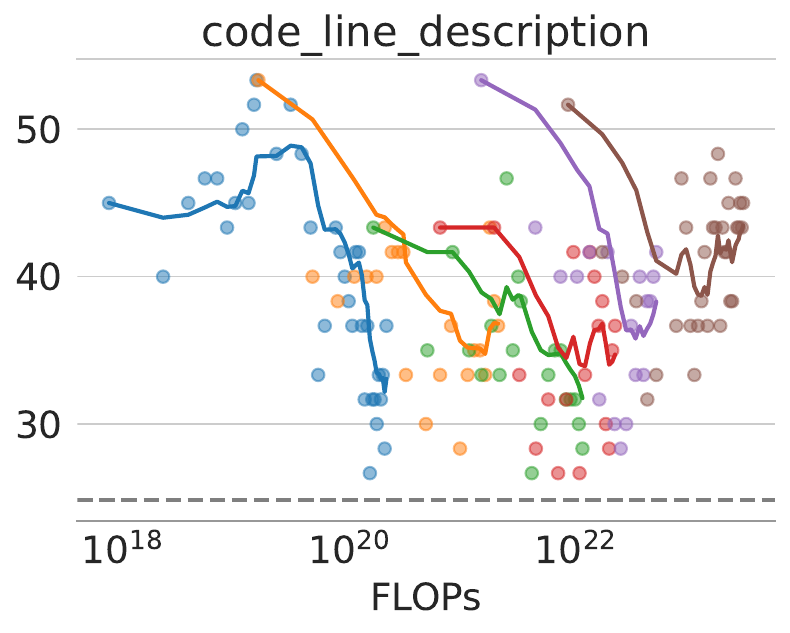}
    \includegraphics[width=0.2\linewidth]{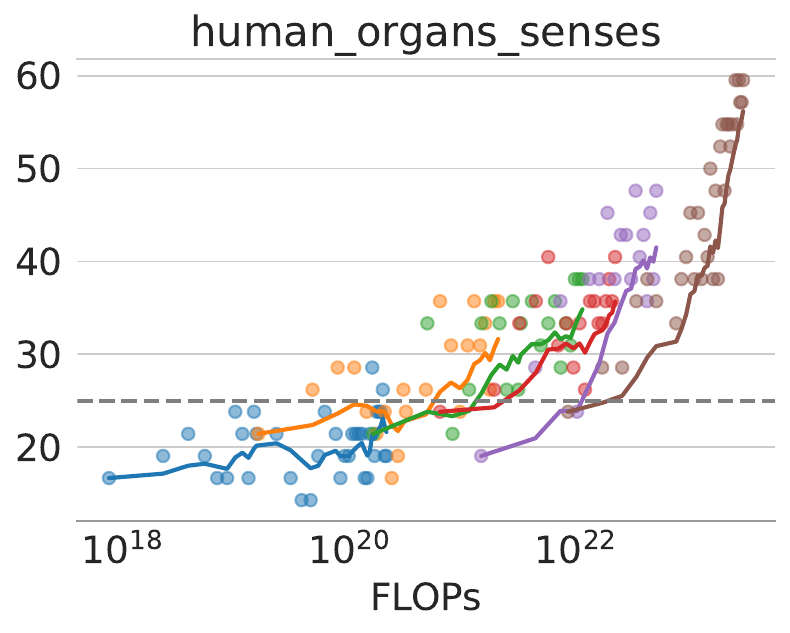}
    \includegraphics[width=0.2\linewidth]{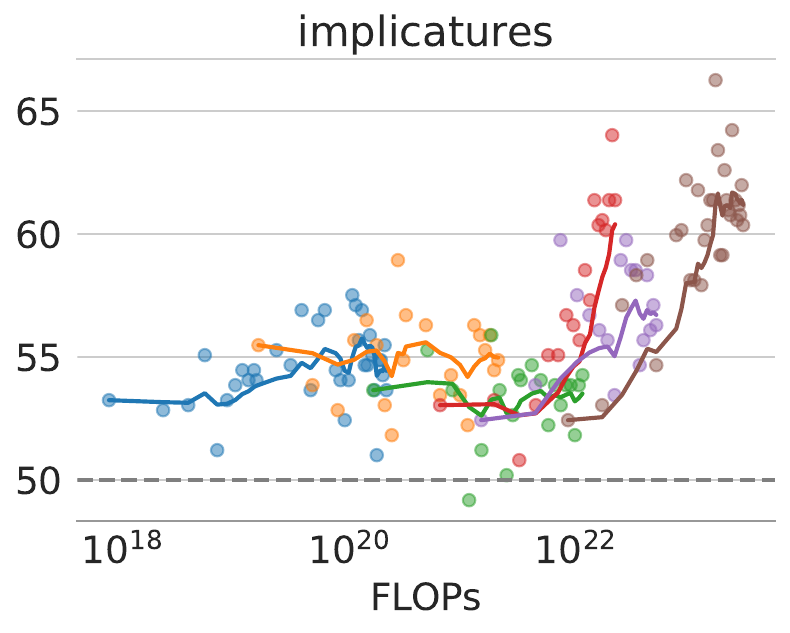}
    \includegraphics[width=0.2\linewidth]{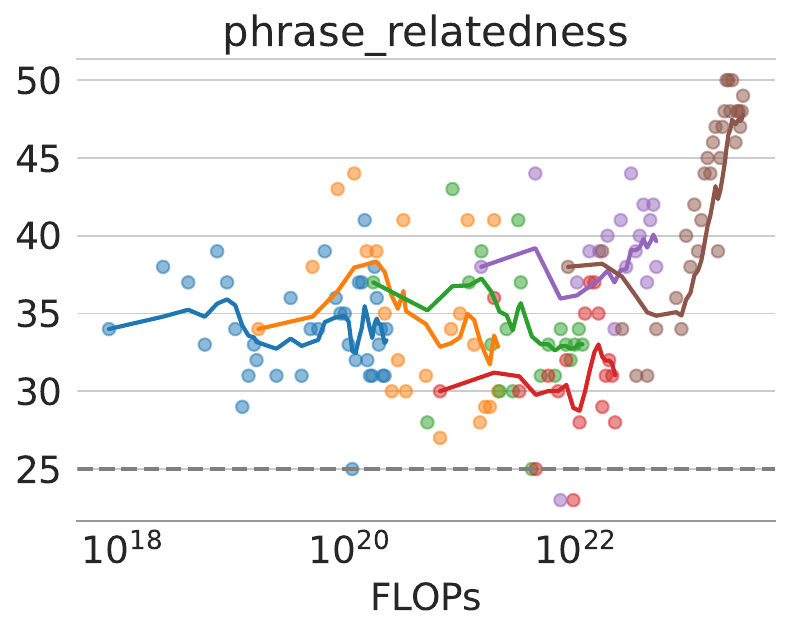}
    \includegraphics[width=0.2\linewidth]{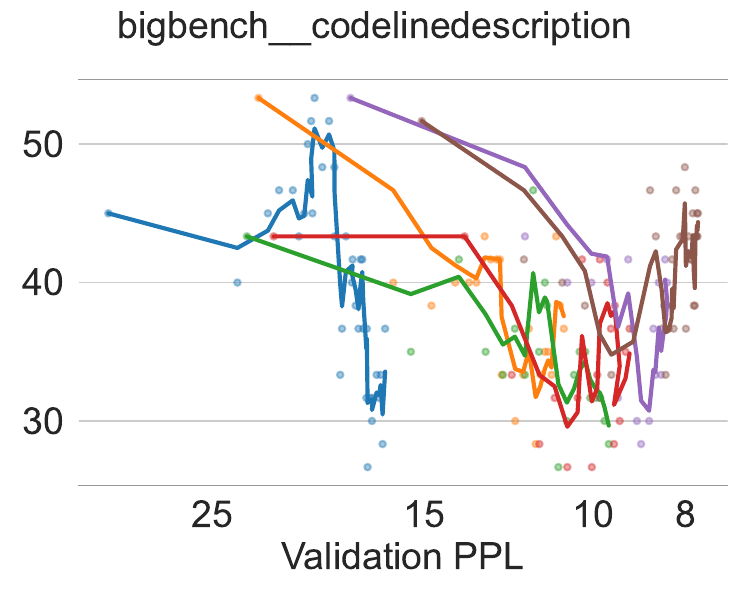}
    \includegraphics[width=0.2\linewidth]{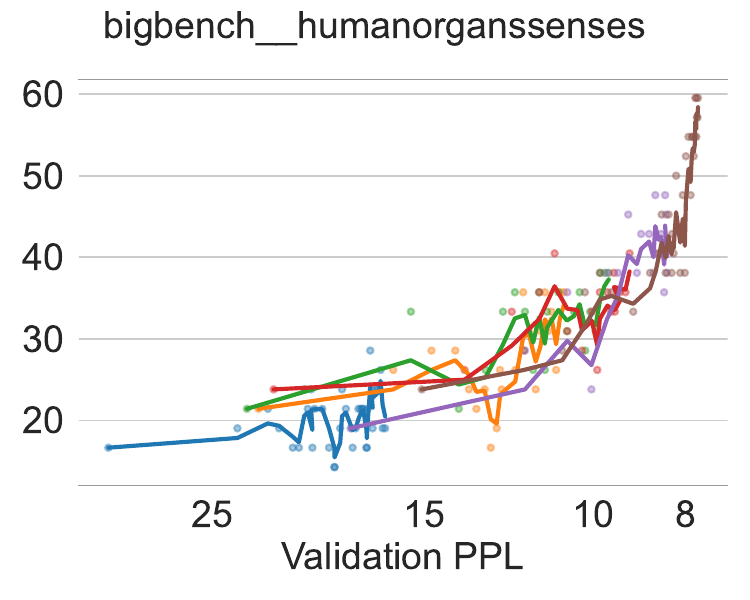}
    \includegraphics[width=0.2\linewidth]{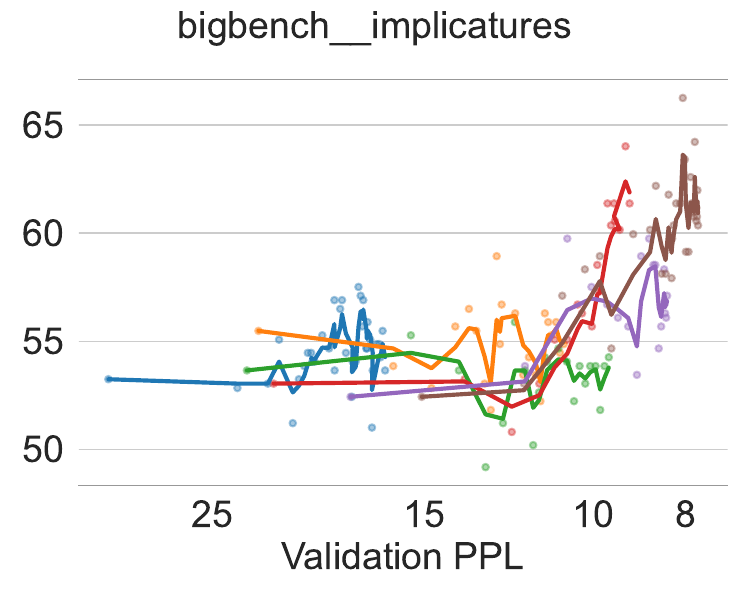}
    \includegraphics[width=0.2\linewidth]{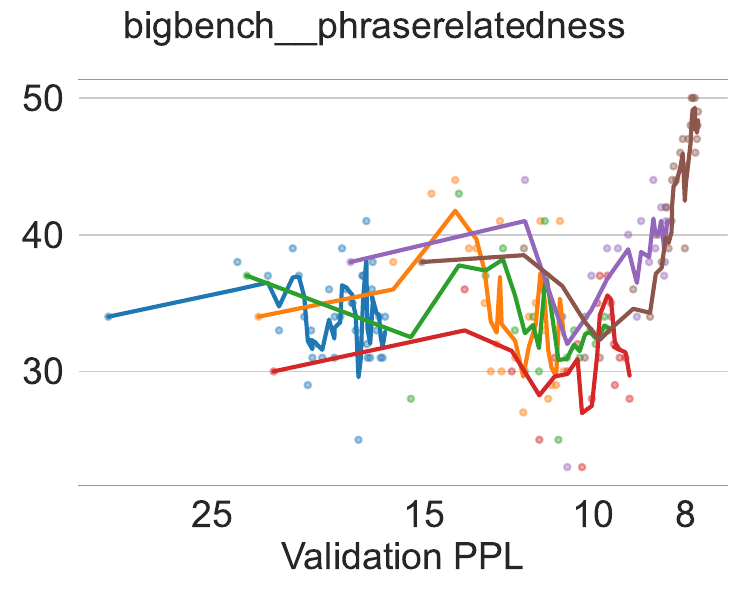}
    \includegraphics[width=0.2\linewidth]{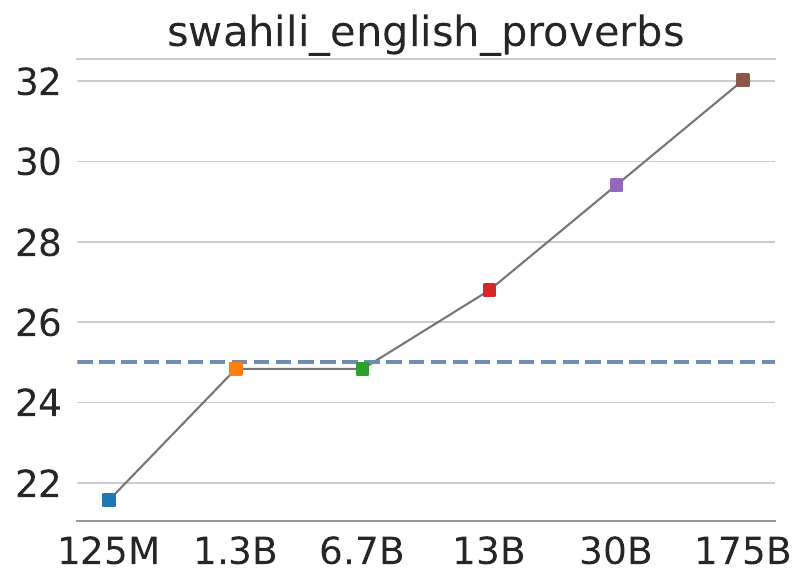}
    \includegraphics[width=0.2\linewidth]{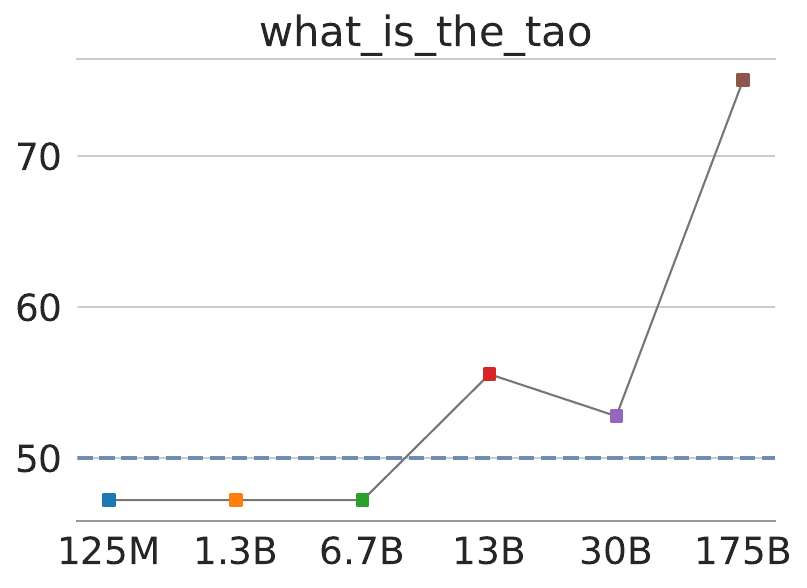} \\
    \includegraphics[width=0.2\linewidth]{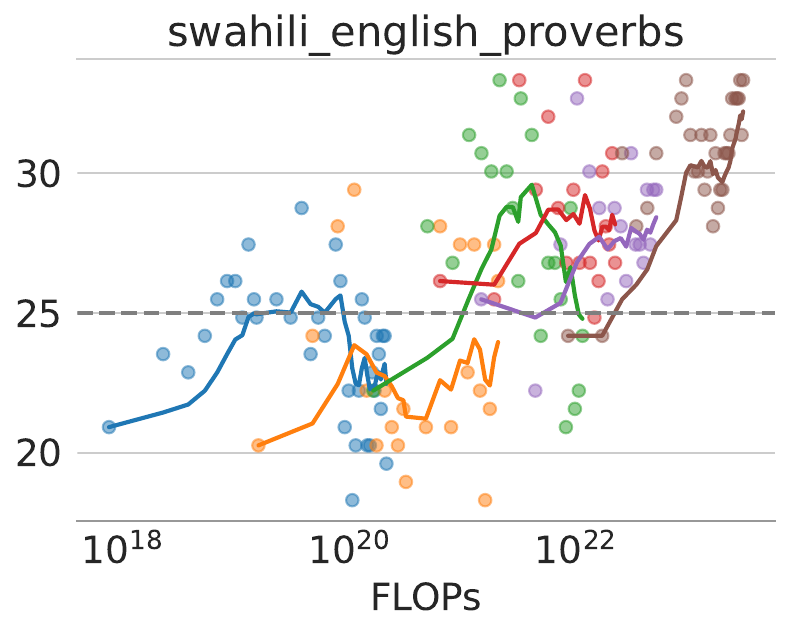}
    \includegraphics[width=0.2\linewidth]{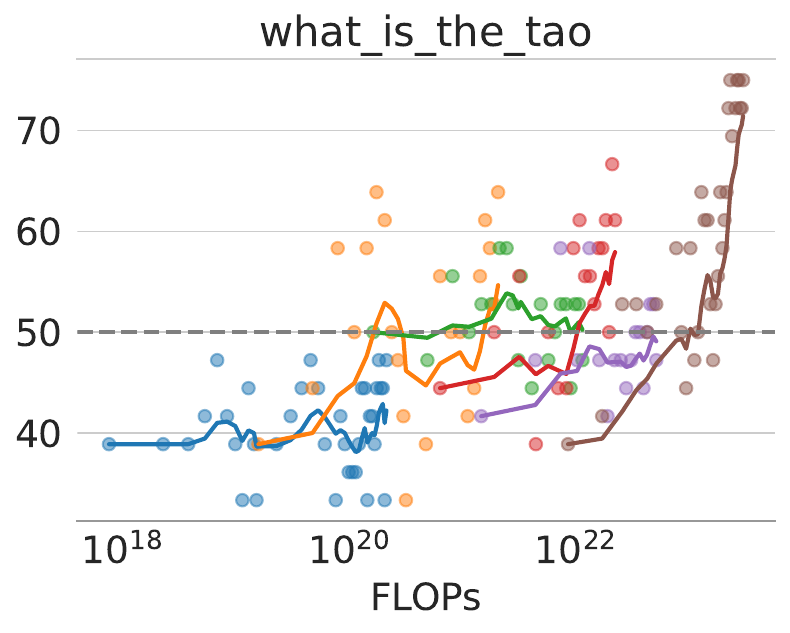} \\
    \includegraphics[width=0.2\linewidth]{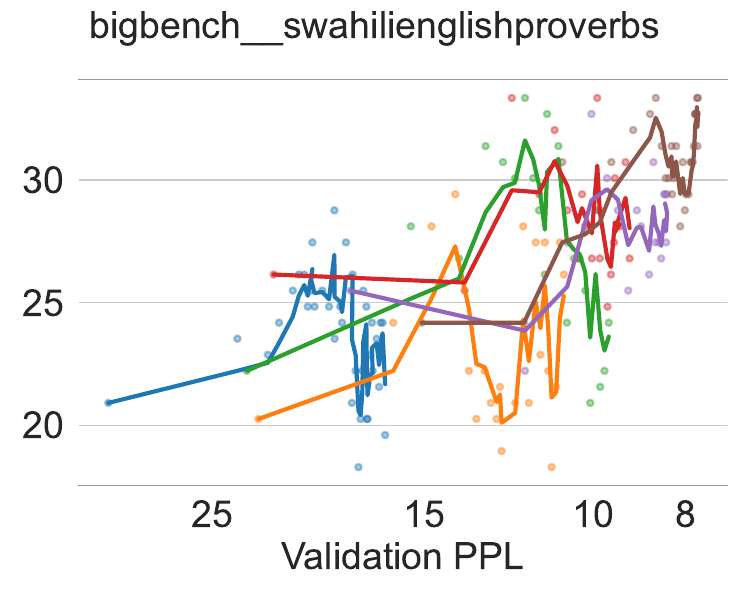}
    \includegraphics[width=0.2\linewidth]{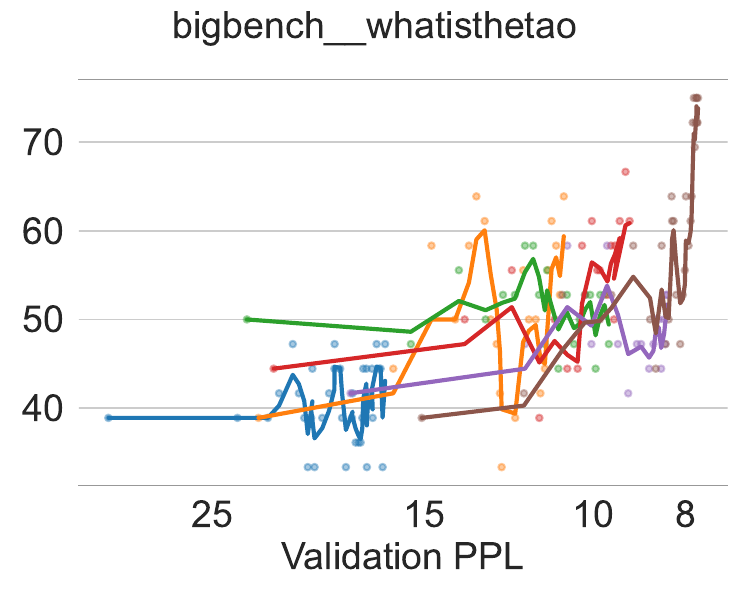} \\
    \caption{Scaling curves and trajectories of breakthroughness tasks. }
    \label{fig:single_breakthrough}
\end{figure*}

We present the scaling curves (on the final model checkpoints) and training trajectories of each linearity and breakthroughness task in \autoref{fig:single_linearity} and \autoref{fig:single_breakthrough}. The evaluation of each task presents a large variance across the training steps. Though the tasks might present a breakthroughness pattern on the scaling curves, their trajectory curves show that language models pick up the task gradually. 

\end{document}